%% file: main.tex
\documentclass{article}

\PassOptionsToPackage{table}{xcolor}
\usepackage{iclr2027_conference,times}
\usepackage{hyperref}
\usepackage{url}

\usepackage[T1]{fontenc}
\usepackage[utf8]{inputenc}
\usepackage{microtype}
\usepackage{inconsolata}
\usepackage[scaled=0.92]{helvet}
\usepackage{graphicx}
\usepackage{xcolor}

\usepackage{amsmath,amssymb,amsfonts}
\usepackage{booktabs}
\usepackage{array}
\usepackage{multirow}
\usepackage[flushleft]{threeparttable}   
\usepackage[most]{tcolorbox}
\usepackage{enumitem}
\usepackage{siunitx}
\usepackage{placeins}   
\usepackage{ifthen}     

\usepackage{pgfplots}
\pgfplotsset{compat=1.18}
\usepgfplotslibrary{groupplots}
\usetikzlibrary{positioning, arrows.meta, shapes.geometric, shapes.misc, fit,
                backgrounds, calc, decorations.pathreplacing, matrix, patterns}
\definecolor{otblue}{HTML}{0072B2}
\definecolor{otorange}{HTML}{E69F00}
\definecolor{otteal}{HTML}{009E73}
\definecolor{otpurple}{HTML}{CC79A7}
\definecolor{otgray}{HTML}{5A5A5A}
\definecolor{lcNavy}{HTML}{1E3A8A}
\definecolor{lcSlate}{HTML}{64748B}
\definecolor{lcCyan}{HTML}{38BDF8}
\definecolor{lcCrimson}{HTML}{E11D48}
\definecolor{lcPink}{HTML}{E11D48}
\definecolor{lcTeal}{HTML}{0F766E}
\definecolor{lcAmber}{HTML}{F59E0B}
\definecolor{lcText}{HTML}{1F2937}
\definecolor{lcMuted}{HTML}{475569}
\newcommand{\pill}[3]{\tikz[baseline=(p.base)]{\node[rounded corners=2.2pt,inner xsep=3pt,inner ysep=1.1pt,fill=#1,text=#2,font=\footnotesize](p){#3};}}
\newcommand{\vConsulted}{\pill{lcTeal!14}{lcTeal!85!black}{consulted}}
\newcommand{\vInert}{\pill{lcCyan!22}{lcNavy}{inert}}
\newcommand{\vUnresolved}{\pill{lcAmber!28}{lcAmber!55!black}{unresolved}}
\newcommand{\vPlaceholder}{\pill{lcSlate!18}{lcSlate!55!black}{placeholder}}
\newcommand{\vContent}{\pill{lcCrimson!13}{lcCrimson!80!black}{content used}}
\definecolor{brick}{HTML}{A55A5A}

\input{macros}
\usepackage{xcolor}
\usepackage[normalem]{ulem}

\title{Does Latent Chain-of-Thought Actually Reason? A Single, Unifying Taxonomy for Latent Reasoning}

\author{Jingjie Qiu$^{1}$, Sheldon Yu$^{1}$, Xunyi Jiang$^{2}$, Jiajin Liu$^{2}$, Zihan Huang$^{1}$,\\
\textbf{Zhendong Chu$^{3}$, Tong Yu$^{4}$, Julian McAuley$^{1}$, Junda Wu$^{4}$}\\
{\normalfont $^{1}$University of California, San Diego\quad
$^{2}$New York University\quad
$^{3}$University of Virginia}\\
{\normalfont $^{4}$Adobe Research}\\
{\normalfont\small\texttt{\{jiq013,ziy040,zih043,jmcauley\}@ucsd.edu}\quad
\texttt{\{tyu,jundaw\}@adobe.com}}\\
{\normalfont\small\texttt{\{xj2511,jiajinliu\}@nyu.edu}\quad
\texttt{zc9uy@virginia.edu}}}

\iclrfinalcopy

\begin{document}
\maketitle
\lhead{Preprint}

\input{sections/0_abstract}

\input{sections/1_intro_cur}
\input{sections/2_related}
\input{sections/3_setup}
\input{sections/4_rq1_determination}   
\input{sections/5_rq2_static}          
\input{sections/6_rq3_dynamics}        
\input{sections/7_discussion}

\input{sections/9_statements}
\bibliography{main}
\bibliographystyle{iclr2027_conference}

\appendix
\input{appendix/a_details}       
\input{appendix/b_rq1}           
\input{appendix/c_rq2}           
\input{appendix/d_rq3}           
\input{appendix/e_checkpoints}   
\input{appendix/f_census}        
\input{appendix/g_limitations}   

\input{appendix/h_corrections}   

\end{document}

%% file: macros.tex
\newif\ifdraft\draftfalse   

\newcommand{\dash}{\textcolor{black!45}{--}}

\newcommand{\coconut}{\textsc{Coconut}}
\newcommand{\codi}{\textsc{CODI}}
\newcommand{\simcot}{\textsc{SIM-CoT}}
\newcommand{\simcoco}{\textsc{SIM-CoT-Coconut}}
\newcommand{\prosqa}{ProsQA}
\newcommand{\prontoqa}{PrOntoQA}
\newcommand{\gsm}{GSM8k}

\newtcolorbox{prereg}[1][Pre-registered reading (not yet run)]{%
  enhanced,
  colback=black!3, colframe=black!62, coltitle=white,
  fonttitle=\bfseries\small, title={#1},
  boxrule=0.5pt, arc=1.6pt, left=6pt, right=6pt, top=4pt, bottom=4pt,
  before skip=8pt, after skip=8pt}

\newcommand{\Neval}{300}                 
\newcommand{\seedval}{0}
\newcommand{\nperm}{200}                 
\newcommand{\probepca}{30}
\newcommand{\inertBound}{3}               

\newcommand{\cocoprosqaCzero}{0.973}
\newcommand{\cocoprosqaCone}{0.977}
\newcommand{\cocoprosqaHelped}{0}
\newcommand{\cocoprosqaHurt}{1}

\newcommand{\cocoprosqadwCzero}{0.973}
\newcommand{\cocoprosqadwCone}{0.973}
\newcommand{\cocoprosqadwHelped}{0}
\newcommand{\cocoprosqadwHurt}{0}
\newcommand{\replmaxdelta}{0.010}        

\newcommand{\cocoprontoCzero}{0.990}
\newcommand{\cocoprontoCone}{0.993}
\newcommand{\cocoprontoHelped}{0}
\newcommand{\cocoprontoHurt}{1}

\newcommand{\codiprosqaCzero}{0.853}
\newcommand{\codiprosqaCone}{0.850}
\newcommand{\codiprosqaHelped}{10}
\newcommand{\codiprosqaHurt}{9}

\newcommand{\codiprontoCzero}{0.933}
\newcommand{\codiprontoCone}{0.940}
\newcommand{\codiprontoHelped}{0}
\newcommand{\codiprontoHurt}{2}

\newcommand{\cocogsmCzero}{0.333}
\newcommand{\cocogsmCone}{0.103}
\newcommand{\cocogsmHelped}{74}
\newcommand{\cocogsmHurt}{5}

\newcommand{\codigsmCzero}{0.430}
\newcommand{\codigsmCone}{0.263}
\newcommand{\codigsmHelped}{69}
\newcommand{\codigsmHurt}{19}

\newcommand{\simgsmCzero}{0.450}
\newcommand{\simgsmCone}{0.047}
\newcommand{\simgsmHelped}{122}

\newcommand{\cocogsmKtwo}{0.197}
\newcommand{\cocogsmKfour}{0.260}
\newcommand{\codigsmKtwo}{0.287}
\newcommand{\codigsmKfour}{0.387}

\newcommand{\cocoprosqadwKtwo}{0.970}
\newcommand{\cocoprosqadwKfour}{0.973}
\newcommand{\tailshareLo}{24\%}
\newcommand{\tailshareHi}{32\%}

\newcommand{\cocogsmDnoise}{-0.243}
\newcommand{\cocogsmDmean}{-0.220}
\newcommand{\cocogsmDdonor}{-0.307}
\newcommand{\codigsmDnoise}{-0.370}
\newcommand{\codigsmDmean}{-0.283}
\newcommand{\codigsmDdonor}{-0.380}

\newcommand{\codiprosqaDnoise}{-0.787}   
\newcommand{\codiprosqaDdonor}{0.000}

\newcommand{\simgsmAeight}{0.123}

\newcommand{\twinretention}{0.810}       

\newcommand{\parityCodiProsqaRef}{0.816}
\newcommand{\parityCodiProsqaFull}{0.818}  

\newcommand{\slotsCoconut}{6}            
\newcommand{\slotsCodi}{6}               
\newcommand{\slotsSimReleased}{10}       
\newcommand{\slotsSimOurs}{6}            

\newcommand{\superposthresh}{30\%}       

\newcommand{\twinretentionprosqa}{0.640} 

\newcommand{\probeauditcorr}{-0.94}

\newcommand{\llamaB}{Llama-3.2-1B-Instruct}
\newcommand{\llamaBshort}{Llama-1B-Inst.}   

\newcommand{\simSixGsmCzero}{0.360}
\newcommand{\simSixGsmCone}{0.080}

\newcommand{\simSixGsmHelped}{85}
\newcommand{\simSixGsmHurt}{1}
\newcommand{\simProsqaCzero}{0.993}
\newcommand{\simProsqaCone}{0.800}

\newcommand{\simProsqaHelped}{58}
\newcommand{\simProsqaHurt}{0}
\newcommand{\simProntoCzero}{1.000}
\newcommand{\simProntoCone}{0.827}

\newcommand{\simProntoHelped}{52}
\newcommand{\simProntoHurt}{0}

\newcommand{\cocogsmGap}{0.230}

\newcommand{\cocogsmLlCzero}{0.370}
\newcommand{\cocogsmLlCone}{0.253}

\newcommand{\cocogsmLlHelped}{59}
\newcommand{\cocogsmLlHurt}{24}
\newcommand{\cocoprosqaLlCzero}{0.983}
\newcommand{\cocoprosqaLlCone}{0.987}
\newcommand{\cocoprosqaLlHelped}{0}
\newcommand{\cocoprosqaLlHurt}{1}
\newcommand{\cocoprontoLlCzero}{0.990}
\newcommand{\cocoprontoLlCone}{0.987}

\newcommand{\cocoprontoLlHelped}{1}
\newcommand{\cocoprontoLlHurt}{0}
\newcommand{\codigsmLlCzero}{0.553}
\newcommand{\codigsmLlCone}{0.360}

\newcommand{\codigsmLlHelped}{67}
\newcommand{\codigsmLlHurt}{9}
\newcommand{\codiprosqaLlCzero}{0.983}
\newcommand{\codiprosqaLlCone}{0.940}

\newcommand{\codiprosqaLlHelped}{13}
\newcommand{\codiprosqaLlHurt}{0}
\newcommand{\codiprontoLlCzero}{0.940}
\newcommand{\codiprontoLlCone}{0.937}

\newcommand{\codiprontoLlHelped}{1}
\newcommand{\codiprontoLlHurt}{0}

\newcommand{\simSixGsmKtwo}{0.203}
\newcommand{\simSixGsmKfour}{0.270}
\newcommand{\simProsqaKtwo}{0.960}
\newcommand{\simProsqaKfour}{0.993}
\newcommand{\simProntoKtwo}{0.857}
\newcommand{\simProntoKfour}{0.913}
\newcommand{\cocoprontoKtwo}{0.993}
\newcommand{\cocoprontoKfour}{0.993}
\newcommand{\codiprosqaKtwo}{0.867}
\newcommand{\codiprosqaKfour}{0.853}
\newcommand{\codiprontoKtwo}{0.937}
\newcommand{\codiprontoKfour}{0.940}
\newcommand{\cocogsmLlKtwo}{0.367}
\newcommand{\cocogsmLlKfour}{0.373}

\newcommand{\simSixGsmDnoise}{-0.290}
\newcommand{\simSixGsmDmean}{-0.263}
\newcommand{\simSixGsmDdonor}{-0.333}
\newcommand{\simProsqaDnoise}{-0.047}
\newcommand{\simProsqaDmean}{-0.073}
\newcommand{\simProsqaDdonor}{-0.077}
\newcommand{\simProntoDnoise}{-0.157}
\newcommand{\simProntoDmean}{-0.440}
\newcommand{\simProntoDdonor}{-0.507}
\newcommand{\cocogsmLlDnoise}{-0.060}
\newcommand{\cocogsmLlDmean}{-0.047}
\newcommand{\cocogsmLlDdonor}{-0.080}
\newcommand{\codigsmLlDnoise}{-0.423}
\newcommand{\codigsmLlDmean}{-0.247}
\newcommand{\codigsmLlDdonor}{-0.373}
\newcommand{\codiprosqaLlDnoise}{0.000}
\newcommand{\codiprosqaLlDmean}{0.000}
\newcommand{\codiprosqaLlDdonor}{0.000}
\newcommand{\codiprosqaDmean}{0.000}   

\newcommand{\cocogsmPoisonPair}{27/4}
\newcommand{\cocogsmPoisonP}{0.000}
\newcommand{\simSixGsmPoisonPair}{20/4}

\newcommand{\simProntoPoisonPair}{124/24}

\newcommand{\codigsmPoisonPair}{73/9}
\newcommand{\codigsmPoisonP}{0.000}
\newcommand{\codigsmLlPoisonPair}{66/12}
\newcommand{\codigsmLlPoisonP}{0.000}
\newcommand{\simgsmPoisonN}{900}       

\newcommand{\simgsmPoisonP}{0.00006}

\newcommand{\simProsqaInvPair}{14/49}  

\newcommand{\codiprosqaBoundPair}{9/10}
\newcommand{\codiprosqaBoundN}{30000}  
\newcommand{\cocogsmLlBoundN}{1800}    
\newcommand{\cocogsmLlPoolN}{1319}     
\newcommand{\codiprosqaLlInvPair}{0/13}
\newcommand{\codiprosqaLlInvP}{0.000}

\newcommand{\poisonCount}{5}
\newcommand{\naCount}{7}          
\newcommand{\boundedCount}{3}     
\newcommand{\invertedCount}{3}    

\newcommand{\codiprosqaRandNorm}{-0.790}
\newcommand{\codiprosqaMeanEps}{0.007}
\newcommand{\codigsmRandNorm}{-0.320}
\newcommand{\codigsmMeanEps}{-0.297}
\newcommand{\codigsmLlRandNorm}{-0.420}
\newcommand{\codigsmLlMeanEps}{-0.247}
\newcommand{\cocogsmRandNorm}{-0.257}
\newcommand{\cocogsmMeanEps}{-0.223}
\newcommand{\simProntoRandNorm}{-0.150}
\newcommand{\simProntoMeanEps}{-0.440}
\newcommand{\cocogsmTwoByTwoSpread}{0.037}  

\newcommand{\twinNenlarged}{480}
\newcommand{\cocogsmEnlUsable}{111}
\newcommand{\cocogsmEnlAtwo}{0.405}
\newcommand{\cocogsmEnlAfour}{0.225}
\newcommand{\cocogsmEnlRtwo}{0.459}
\newcommand{\cocogsmEnlRfour}{0.532}
\newcommand{\codigsmEnlUsable}{153}
\newcommand{\codigsmEnlAtwo}{0.582}
\newcommand{\codigsmEnlAfour}{0.327}
\newcommand{\codigsmEnlRtwo}{0.072}
\newcommand{\codigsmEnlRfour}{0.268}
\newcommand{\simSixGsmUsable}{57}
\newcommand{\simSixGsmAtwo}{0.421}
\newcommand{\simSixGsmAfour}{0.211}
\newcommand{\simSixGsmRtwo}{0.386}
\newcommand{\simSixGsmRfour}{0.526}
\newcommand{\simProsqaUsable}{64}
\newcommand{\simProsqaAtwo}{0.750}
\newcommand{\simProsqaAfour}{0.250}
\newcommand{\simProsqaRtwo}{0.250}
\newcommand{\simProsqaRfour}{0.750}

\newcommand{\pairCocoN}{88}
\newcommand{\pairCocoTwoGpt}{0.386}
\newcommand{\pairCocoTwoLl}{0.955}
\newcommand{\pairCocoTwoDisc}{1/51}
\newcommand{\pairCocoTwoP}{0.0000}
\newcommand{\pairCocoTwoGapv}{0.568}
\newcommand{\pairCocoTwoMDE}{0.234}
\newcommand{\pairCocoFourGpt}{0.239}
\newcommand{\pairCocoFourLl}{0.818}
\newcommand{\pairCocoFourDisc}{4/55}
\newcommand{\pairCocoFourP}{0.0000}
\newcommand{\pairCocoFourGapv}{0.580}
\newcommand{\pairCocoFourMDE}{0.251}
\newcommand{\pairCodiN}{140}
\newcommand{\pairCodiTwoGpt}{0.600}
\newcommand{\pairCodiTwoLl}{0.807}
\newcommand{\pairCodiTwoDisc}{5/34}
\newcommand{\pairCodiTwoP}{0.0000}
\newcommand{\pairCodiTwoGapv}{0.207}
\newcommand{\pairCodiTwoMDE}{0.133}
\newcommand{\pairCodiFourGpt}{0.336}
\newcommand{\pairCodiFourLl}{0.479}
\newcommand{\pairCodiFourDisc}{9/29}
\newcommand{\pairCodiFourP}{0.0017}
\newcommand{\pairCodiFourGapv}{0.143}
\newcommand{\pairCodiFourMDE}{0.125}

\newcommand{\enlGapShareLo}{83\%}
\newcommand{\enlGapShareHi}{86\%}
\newcommand{\specFlipLo}{0.25}
\newcommand{\specFlipHi}{0.40}
\newcommand{\specNullLo}{0.04}
\newcommand{\specNullHi}{0.07}

\newcommand{\simSixGsmPoisonGap}{0.053}
\newcommand{\simSixGsmPoisonMDE}{0.048}
\newcommand{\simProsqaInvGap}{0.117}
\newcommand{\simProsqaInvMDE}{0.074}
\newcommand{\simProntoPoisonGap}{0.333}
\newcommand{\simProntoPoisonMDE}{0.116}

\newcommand{\superposThreshCount}{90}       
\newcommand{\superposNull}{0.0025}       
\newcommand{\superposMDE}{0.0072}
\newcommand{\ceilGpt}{0.817}             
\newcommand{\ceilLlama}{0.627}           

\newcommand{\cocoprosqaBoth}{0.100}

\newcommand{\cocoprosqadwBoth}{0.080}
\newcommand{\cocoprosqadwBothSlot}{p0}
\newcommand{\cocoprosqaLlBoth}{0.017}
\newcommand{\cocoprosqaLlBothSlot}{p4}
\newcommand{\simProsqaBoth}{0.067}
\newcommand{\simProsqaBothSlot}{p4}
\newcommand{\simProsqaElig}{5/6}

\newcommand{\chanceMag}{0.200}           
\newcommand{\mdeMag}{0.238}

\newcommand{\chanceProsqaConcept}{0.026}
\newcommand{\mdeProsqaConcept}{0.042}
\newcommand{\cocoprosqaPeak}{0.823}
\newcommand{\cocoprosqaPeakCell}{p5 L1}
\newcommand{\cocoprosqaCleared}{78/78}
\newcommand{\cocoprosqadwPeak}{0.820}
\newcommand{\cocoprosqadwPeakCell}{p2 L0}
\newcommand{\cocoprosqadwCleared}{78/78}
\newcommand{\cocoprosqaLlPeak}{0.217}
\newcommand{\cocoprosqaLlPeakCell}{p4 L7}
\newcommand{\cocoprosqaLlCleared}{102/102}
\newcommand{\simProsqaPeak}{0.967}
\newcommand{\simProsqaPeakCell}{p5 L7}
\newcommand{\simProsqaCleared}{56/78}
\newcommand{\codiprosqaPeak}{0.100}
\newcommand{\codiprosqaPeakCell}{p0 L1}
\newcommand{\codiprosqaCleared}{27/78}
\newcommand{\codiprosqaLlPeak}{0.220}
\newcommand{\codiprosqaLlPeakCell}{p1 L7}
\newcommand{\codiprosqaLlCleared}{63/102}
\newcommand{\codigsmCleared}{78/78}
\newcommand{\codigsmPeak}{0.417}
\newcommand{\codigsmPeakCell}{p4 L12}
\newcommand{\simgsmCleared}{130/130}
\newcommand{\simgsmPeak}{0.650}
\newcommand{\simgsmPeakCell}{p9 L5}

\newcommand{\portraitUlo}{0.063}         
\newcommand{\portraitUemb}{0.760}        
\newcommand{\portraitUtop}{0.753}        

\newcommand{\cocogsmAccZero}{0.310}
\newcommand{\cocogsmAccLast}{0.587}
\newcommand{\cocogsmAccDelta}{+0.277}
\newcommand{\simSixGsmAccZero}{0.320}
\newcommand{\simSixGsmAccLast}{0.553}
\newcommand{\simSixGsmAccDelta}{+0.233}
\newcommand{\simgsmAccZero}{0.320}
\newcommand{\simgsmAccLast}{0.607}
\newcommand{\simgsmAccDelta}{+0.287}
\newcommand{\codigsmAccZero}{0.407}
\newcommand{\codigsmAccLast}{0.377}
\newcommand{\codigsmAccDelta}{-0.030}
\newcommand{\codigsmAccP}{0.452}
\newcommand{\cocogsmLlAccZero}{0.370}
\newcommand{\cocogsmLlAccLast}{0.383}
\newcommand{\cocogsmLlAccDelta}{+0.013}
\newcommand{\cocogsmLlAccP}{0.736}
\newcommand{\codigsmLlAccZero}{0.393}
\newcommand{\codigsmLlAccLast}{0.380}
\newcommand{\codigsmLlAccDelta}{-0.013}
\newcommand{\codigsmLlAccP}{0.737}
\newcommand{\accBuildP}{0.0001}          
\newcommand{\accDeltaLo}{+0.23}
\newcommand{\accDeltaHi}{+0.29}

\newcommand{\offpathClean}{0/24}         
\newcommand{\offpathCodi}{4/24}          

\newcommand{\diagKzeroPzero}{0.389}

\newcommand{\diagKonePzero}{0.403}\newcommand{\diagKonePone}{0.369}

\newcommand{\diagKtwoPone}{0.533}\newcommand{\diagKtwoPtwo}{0.629}

\newcommand{\diagKthreePtwo}{0.838}\newcommand{\diagKthreePthree}{0.846}
\newcommand{\diagKzeroPone}{0.035}\newcommand{\diagKzeroPtwo}{0.014}\newcommand{\diagKzeroPthree}{0.045}\newcommand{\diagKzeroPfour}{0.031}\newcommand{\diagKzeroPfive}{0.045}
\newcommand{\diagKonePtwo}{0.055}\newcommand{\diagKonePthree}{0.066}\newcommand{\diagKonePfour}{0.062}\newcommand{\diagKonePfive}{0.066}
\newcommand{\diagKtwoPzero}{0.052}\newcommand{\diagKtwoPthree}{0.320}\newcommand{\diagKtwoPfour}{0.351}\newcommand{\diagKtwoPfive}{0.309}
\newcommand{\diagKthreePzero}{0.060}\newcommand{\diagKthreePone}{0.137}\newcommand{\diagKthreePfour}{0.761}\newcommand{\diagKthreePfive}{0.752}
\newcommand{\fitCocoPzero}{0.597}\newcommand{\fitCocoPfive}{0.607}   
\newcommand{\fitSimPfive}{0.963}   

\newcommand{\dynProjDim}{128}
\newcommand{\dynKset}{5, 8, 12}
\newcommand{\dynNperm}{200}
\newcommand{\dynSeedStable}{146}
\newcommand{\dynSeedTotal}{153}
\newcommand{\dynSeeds}{5}
\newcommand{\dynFloorSlots}{0.5}

\newcommand{\tTwoCocoProsqaSingle}{14}
\newcommand{\tTwoCocoProsqaLast}{2.07}
\newcommand{\tTwoCocoProsqaNull}{4.21}
\newcommand{\tTwoCocoProsqaDelta}{2.14}
\newcommand{\tTwoCocoProsqaBSingle}{113}
\newcommand{\tTwoCocoProsqaBLast}{1.51}
\newcommand{\tTwoCocoProsqaBNull}{4.46}
\newcommand{\tTwoCocoProsqaBDelta}{2.94}
\newcommand{\tTwoCocoProsqaLlSingle}{0}
\newcommand{\tTwoCocoProsqaLlLast}{1.65}
\newcommand{\tTwoCocoProsqaLlNull}{4.26}
\newcommand{\tTwoCocoProsqaLlDelta}{2.61}
\newcommand{\tTwoCodiProsqaSingle}{53}
\newcommand{\tTwoCodiProsqaLast}{1.93}
\newcommand{\tTwoCodiProsqaNull}{4.17}
\newcommand{\tTwoCodiProsqaDelta}{2.24}
\newcommand{\tTwoCodiProsqaLlSingle}{0}
\newcommand{\tTwoCodiProsqaLlLast}{2.03}
\newcommand{\tTwoCodiProsqaLlNull}{3.47}
\newcommand{\tTwoCodiProsqaLlDelta}{1.44}
\newcommand{\tTwoSimProsqaSingle}{0}
\newcommand{\tTwoSimProsqaLast}{3.78}
\newcommand{\tTwoSimProsqaNull}{3.85}
\newcommand{\tTwoSimProsqaDelta}{0.07}
\newcommand{\tTwoCocoProntoSingle}{6}
\newcommand{\tTwoCocoProntoLast}{2.77}
\newcommand{\tTwoCocoProntoNull}{3.29}
\newcommand{\tTwoCocoProntoDelta}{0.52}
\newcommand{\tTwoCocoProntoLlSingle}{35}
\newcommand{\tTwoCocoProntoLlLast}{1.50}
\newcommand{\tTwoCocoProntoLlNull}{4.61}
\newcommand{\tTwoCocoProntoLlDelta}{3.11}
\newcommand{\tTwoCodiProntoSingle}{0}
\newcommand{\tTwoCodiProntoLast}{2.56}
\newcommand{\tTwoCodiProntoNull}{4.09}
\newcommand{\tTwoCodiProntoDelta}{1.53}
\newcommand{\tTwoCodiProntoLlSingle}{0}
\newcommand{\tTwoCodiProntoLlLast}{1.00}
\newcommand{\tTwoCodiProntoLlNull}{1.00}
\newcommand{\tTwoCodiProntoLlDelta}{0.00}
\newcommand{\tTwoSimProntoSingle}{0}
\newcommand{\tTwoSimProntoLast}{4.00}
\newcommand{\tTwoSimProntoNull}{4.49}
\newcommand{\tTwoSimProntoDelta}{0.49}
\newcommand{\tTwoCocoGsmSingle}{0}
\newcommand{\tTwoCocoGsmLast}{4.96}
\newcommand{\tTwoCocoGsmNull}{4.96}
\newcommand{\tTwoCocoGsmDelta}{0.00}
\newcommand{\tTwoSimSixGsmSingle}{0}
\newcommand{\tTwoSimSixGsmLast}{4.98}
\newcommand{\tTwoSimSixGsmNull}{4.99}
\newcommand{\tTwoSimSixGsmDelta}{0.01}

\newcommand{\tTwoCodiGsmSingle}{0}
\newcommand{\tTwoCodiGsmLast}{4.98}
\newcommand{\tTwoCodiGsmNull}{4.99}
\newcommand{\tTwoCodiGsmDelta}{0.01}
\newcommand{\tTwoCocoGsmLlSingle}{254}
\newcommand{\tTwoCocoGsmLlLast}{1.65}
\newcommand{\tTwoCocoGsmLlNull}{4.74}
\newcommand{\tTwoCocoGsmLlDelta}{3.09}
\newcommand{\tTwoCodiGsmLlSingle}{0}
\newcommand{\tTwoCodiGsmLlLast}{4.80}
\newcommand{\tTwoCodiGsmLlNull}{4.89}
\newcommand{\tTwoCodiGsmLlDelta}{0.09}

\newcommand{\tFourSimProsqaAcc}{0.030}

\newcommand{\ctrlProsqaCzero}{0.970}   
\newcommand{\ctrlProsqaCone}{0.973}   
\newcommand{\ctrlProsqaHelped}{0}   
\newcommand{\ctrlProsqaHurt}{1}   

\newcommand{\codigsmLlKtwo}{0.487}
\newcommand{\codigsmLlKfour}{0.453}

\newcommand{\ceilGptR}{1.000}            
\newcommand{\ceilLlamaR}{0.960}          
\newcommand{\cocoprosqaBothR}{0.067}     
\newcommand{\cocoprosqadwBothR}{0.090}   
\newcommand{\cocoprosqaLlBothR}{0.060}   

\newcommand{\simProsqaBothR}{0.083}      

\newcommand{\codiprosqaLlBothR}{0.023}   
\newcommand{\codiprosqaLlEligR}{6/6}
\newcommand{\codiprosqaLlBothSlotR}{p0}
\newcommand{\breadthMaxPeakR}{0.113}     
\newcommand{\codiprosqaFiveX}{0}         
\newcommand{\codiprosqaFiveXR}{7}        
\newcommand{\codiprosqaClearedR}{20/78}
\newcommand{\codiprosqaPeakR}{0.177}
\newcommand{\codiprosqaPeakCellR}{p2 L1}
\newcommand{\codiprosqaLlFiveX}{6}       
\newcommand{\codiprosqaLlFiveXR}{32}     
\newcommand{\codiprosqaLlClearedR}{70/102}
\newcommand{\codiprosqaLlPeakR}{0.410}
\newcommand{\codiprosqaLlPeakCellR}{p1 L6}
\newcommand{\cocoprosqaPeakR}{0.933}     
\newcommand{\cocoprosqadwPeakR}{0.927}   
\newcommand{\cocoprosqaLlPeakR}{0.797}   
\newcommand{\cocoprosqaLlPeakCellR}{p2 L2}
\newcommand{\simProsqaPeakR}{0.987}      
\newcommand{\simProsqaPeakCellR}{p5 L10}
\newcommand{\portraitUembR}{0.933}       
\newcommand{\portraitUloR}{0.497}        
\newcommand{\portraitUtopR}{0.907}       
\newcommand{\portraitBUembR}{0.927}      
\newcommand{\portraitBUloR}{0.550}       
\newcommand{\portraitBUtopR}{0.893}      
\newcommand{\mlpMaxGain}{0.0067}         
\newcommand{\cocoprosqaFiveX}{66}\newcommand{\cocoprosqaFiveXR}{78}\newcommand{\cocoprosqaClearedR}{78/78}\newcommand{\cocoprosqaPeakCellR}{p0 L0}
\newcommand{\cocoprosqadwFiveX}{71}\newcommand{\cocoprosqadwFiveXR}{78}\newcommand{\cocoprosqadwClearedR}{78/78}\newcommand{\cocoprosqadwPeakCellR}{p0 L0}
\newcommand{\simProsqaFiveX}{54}\newcommand{\simProsqaFiveXR}{54}\newcommand{\simProsqaClearedR}{55/78}
\newcommand{\cocoprosqaLlFiveX}{86}\newcommand{\cocoprosqaLlFiveXR}{102}\newcommand{\cocoprosqaLlClearedR}{102/102}

\newcommand{\presentCocoR}{24/24}\newcommand{\presentCocoBR}{23/24}\newcommand{\presentCocoLlR}{12/24}
\newcommand{\presentSimR}{20/24}\newcommand{\presentCodiR}{18/24}\newcommand{\presentCodiLlR}{17/24}
\newcommand{\presentCoco}{18/24}\newcommand{\presentCocoB}{15/24}\newcommand{\presentCocoLl}{11/24}
\newcommand{\presentSim}{18/24}\newcommand{\presentCodi}{15/24}\newcommand{\presentCodiLl}{13/24}
\newcommand{\offpathCodiR}{0/24}         
\newcommand{\offpathCodiLl}{0/24}        
\newcommand{\offpathCodiLlR}{2/24}       
\newcommand{\codiOnpathLo}{0.026}        
\newcommand{\codiOnpathHi}{0.107}
\newcommand{\simOnpathLo}{0.034}         
\newcommand{\simOnpathHi}{0.872}

\newcommand{\tFourFitSlot}{last latent slot}
\newcommand{\tFourCellsPCA}{17}\newcommand{\tFourRows}{69}
\newcommand{\tFourNullDependent}{32}
\newcommand{\tFourMovedTotal}{18}\newcommand{\tFourMovedGain}{9}\newcommand{\tFourMovedLoss}{6}
\newcommand{\dynCellsNotAlignedRefit}{11}   
\newcommand{\dynCellsUsablePCA}{15}
\newcommand{\tFourLiftMDEmin}{0.0085}

\newcommand{\tFourLCocoProsqaAcc}{0.607}\newcommand{\tFourLCocoProsqaAlong}{0.064}\newcommand{\tFourLCocoProsqaRatio}{1.00}
\newcommand{\tFourLCocoProsqaBAcc}{0.513}\newcommand{\tFourLCocoProsqaBAlong}{0.069}\newcommand{\tFourLCocoProsqaBRatio}{1.16}
\newcommand{\tFourLCocoProsqaLlAcc}{0.210}\newcommand{\tFourLCocoProsqaLlAlong}{0.052}\newcommand{\tFourLCocoProsqaLlRatio}{1.21}
\newcommand{\tFourLSimProsqaAcc}{0.963}\newcommand{\tFourLSimProsqaAlong}{0.170}\newcommand{\tFourLSimProsqaRatio}{1.94}
\newcommand{\tFourLSimProsqaLift}{-0.014}   
\newcommand{\tFourLCodiProsqaAcc}{0.043}\newcommand{\tFourLCodiProsqaAlong}{0.041}\newcommand{\tFourLCodiProsqaRatio}{1.08}
\newcommand{\tFourLCodiProsqaLlAcc}{0.037}  
\newcommand{\tFourLRCocoProsqaAcc}{0.813}\newcommand{\tFourLRCocoProsqaAlong}{0.033}\newcommand{\tFourLRCocoProsqaRatio}{0.49}
\newcommand{\tFourLRCocoProsqaBAcc}{0.850}\newcommand{\tFourLRCocoProsqaBAlong}{0.036}\newcommand{\tFourLRCocoProsqaBRatio}{0.62}
\newcommand{\tFourLRCocoProsqaLlAcc}{0.720}\newcommand{\tFourLRCocoProsqaLlAlong}{0.041}\newcommand{\tFourLRCocoProsqaLlRatio}{0.82}
\newcommand{\tFourLRSimProsqaAcc}{0.973}\newcommand{\tFourLRSimProsqaAlong}{0.111}\newcommand{\tFourLRSimProsqaRatio}{1.23}
\newcommand{\tFourLRCodiProsqaAcc}{0.030}\newcommand{\tFourLRCodiProsqaLlAcc}{0.027}   
\newcommand{\tFourLCocoGsmAcc}{0.603}\newcommand{\tFourLCocoGsmAlong}{0.051}\newcommand{\tFourLCocoGsmRatio}{0.81}
\newcommand{\tFourLSimSixGsmAcc}{0.613}\newcommand{\tFourLSimSixGsmAlong}{0.039}\newcommand{\tFourLSimSixGsmRatio}{0.72}
\newcommand{\tFourLSimTenGsmAcc}{0.660}\newcommand{\tFourLSimTenGsmAlong}{0.095}\newcommand{\tFourLSimTenGsmRatio}{1.09}
\newcommand{\tFourLCodiGsmAcc}{0.547}\newcommand{\tFourLCodiGsmAlong}{0.043}\newcommand{\tFourLCodiGsmRatio}{0.80}
\newcommand{\tFourLCocoGsmLlAcc}{0.557}\newcommand{\tFourLCocoGsmLlAlong}{0.036}\newcommand{\tFourLCocoGsmLlRatio}{0.78}
\newcommand{\tFourLCodiGsmLlAcc}{0.567}\newcommand{\tFourLCodiGsmLlAlong}{0.075}\newcommand{\tFourLCodiGsmLlRatio}{0.90}
\newcommand{\tFourLCocoProntoAcc}{0.587}\newcommand{\tFourLCocoProntoAlong}{0.032}\newcommand{\tFourLCocoProntoRatio}{0.67}
\newcommand{\tFourLCodiProntoAcc}{0.530}    
\newcommand{\tFourLCocoProntoLlAcc}{0.983}\newcommand{\tFourLCocoProntoLlAlong}{0.123}\newcommand{\tFourLCocoProntoLlRatio}{2.25}
\newcommand{\tFourLCodiProntoLlAcc}{0.913}\newcommand{\tFourLCodiProntoLlAlong}{0.020}\newcommand{\tFourLCodiProntoLlRatio}{0.33}
\newcommand{\tFourLSimProntoAcc}{1.000}\newcommand{\tFourLSimProntoAlong}{0.046}\newcommand{\tFourLSimProntoRatio}{0.69}

\newcommand{\simProsqaLlCzero}{0.997}      
\newcommand{\simProsqaLlCone}{0.077}       
\newcommand{\simProsqaLlHelped}{276}       
\newcommand{\simProsqaLlHurt}{0}           
\newcommand{\simProsqaLlHolmP}{0.0006}     
\newcommand{\simProsqaLlMoved}{0}          
\newcommand{\simProsqaLlDdonor}{0.000}     
\newcommand{\simProsqaLlDnoise}{0.000}     
\newcommand{\simProsqaLlDmean}{0.000}      

\newcommand{\cocoprosqadwDnoise}{0.000}
\newcommand{\cocoprosqadwDmean}{0.000}
\newcommand{\cocoprosqadwDdonor}{0.000}
\newcommand{\cocoprontoDnoise}{+0.003}
\newcommand{\cocoprontoDmean}{+0.003}
\newcommand{\cocoprontoDdonor}{+0.003}
\newcommand{\codiprontoDnoise}{-0.020}     
\newcommand{\codiprontoDmean}{+0.003}
\newcommand{\codiprontoDdonor}{+0.003}
\newcommand{\cocoprosqaLlDnoise}{0.000}
\newcommand{\cocoprosqaLlDmean}{0.000}
\newcommand{\cocoprosqaLlDdonor}{0.000}
\newcommand{\cocoprontoLlDnoise}{+0.003}
\newcommand{\cocoprontoLlDmean}{0.000}
\newcommand{\cocoprontoLlDdonor}{0.000}
\newcommand{\codiprontoLlDnoise}{0.000}
\newcommand{\codiprontoLlDmean}{0.000}
\newcommand{\codiprontoLlDdonor}{+0.003}
\newcommand{\simProntoLlDnoise}{0.000}     
\newcommand{\simProntoLlDmean}{0.000}
\newcommand{\simProntoLlDdonor}{0.000}
\newcommand{\ctrlLlCzero}{0.877}           
\newcommand{\ctrlLlCone}{0.850}            
\newcommand{\ctrlLlHelped}{16}             
\newcommand{\ctrlLlHurt}{8}                
\newcommand{\ctrlLlDelta}{-0.027}          
\newcommand{\ctrlLlMDE}{0.031}            
\newcommand{\ctrlLlMoved}{0}               
\newcommand{\simProsqaLlFitPzero}{0.070}   
\newcommand{\simProsqaLlFitPfive}{0.923}   
\newcommand{\ctrlLlFitPzero}{0.250}
\newcommand{\ctrlLlFitPfive}{0.217}
\newcommand{\simProsqaLlBoth}{0.083}       
\newcommand{\ctrlLlBoth}{0.043}            
\newcommand{\simProsqaLlLive}{23}          
\newcommand{\ctrlLlLive}{12}

\newcommand{\simProsqaLlStatesFive}{3.88}  

\newcommand{\ctrlLlStatesFive}{2.00}       

\newcommand{\simProsqaLlDeltaK}{0.106}     
\newcommand{\ctrlLlDeltaK}{1.48}           

\newcommand{\simProntoLlCzero}{0.993}
\newcommand{\simProntoLlCone}{0.990}
\newcommand{\simProntoLlHelped}{2}          
\newcommand{\simProntoLlHurt}{1}            
\newcommand{\simProntoLlDelta}{-0.003}
\newcommand{\simProntoLlMDE}{0.008}
\newcommand{\simProntoLlMoved}{3}           
\newcommand{\simProntoLlCkpt}{9}
\newcommand{\lTwoSweepConstant}{8}          
\newcommand{\lTwoSweepTotal}{13}            
\newcommand{\lTwoSweepAccNine}{1.000}       
\newcommand{\simProntoLlKtwo}{0.990}        
\newcommand{\simProntoLlKfour}{0.993}       
\newcommand{\simProntoLlStatesFive}{1.000}  

\newcommand{\simProsqaLlFitPzeroR}{0.457}
\newcommand{\simProsqaLlFitPfiveR}{0.777}
\newcommand{\ctrlLlFitPzeroR}{0.587}
\newcommand{\ctrlLlFitPfiveR}{0.497}
\newcommand{\simProsqaLlBothR}{0.103}       
\newcommand{\ctrlLlBothR}{0.077}
\newcommand{\simProsqaLlLiveR}{24}          
\newcommand{\ctrlLlLiveR}{17}               

\newcommand{\llamaCeilPCA}{0.627}           
\newcommand{\llamaCeilRidge}{0.960}
\newcommand{\simProsqaLlKone}{0.987}        
\newcommand{\simProsqaLlKtwo}{0.997}
\newcommand{\simProsqaLlKfour}{0.997}

\newcommand{\ctrlLlKone}{0.860}
\newcommand{\ctrlLlKtwo}{0.863}
\newcommand{\ctrlLlKfour}{0.867}
\newcommand{\simProsqaLlMovedMax}{0}

\newcommand{\gptTwinMovedRandom}{16}        
\newcommand{\gptTwinMovedMean}{24}
\newcommand{\gptTwinMovedDonor}{23}
\newcommand{\simProsqaLlAlong}{0.165}
\newcommand{\simProsqaLlRefitHi}{0.099}     
\newcommand{\simProsqaLlCrossLo}{0.191}     
\newcommand{\ctrlLlAlong}{0.097}
\newcommand{\ctrlLlRefitHi}{0.078}
\newcommand{\ctrlLlCrossLo}{0.181}
\newcommand{\pairSimTwinsLl}{53}            
\newcommand{\pairSimTwinsGpt}{64}
\newcommand{\pairSimTwinsCtrl}{76}

\newcommand{\pairSimFourLl}{1.000}

\newcommand{\pairSimFourGpt}{0.250}

\newcommand{\diagCocoKzeroPzero}{0.035}
\newcommand{\diagCocoKzeroPone}{0.038}
\newcommand{\diagCocoKzeroPtwo}{0.038}
\newcommand{\diagCocoKzeroPthree}{0.031}
\newcommand{\diagCocoKzeroPfour}{0.035}
\newcommand{\diagCocoKzeroPfive}{0.031}

\newcommand{\diagCocoKonePzero}{0.052}
\newcommand{\diagCocoKonePone}{0.048}
\newcommand{\diagCocoKonePtwo}{0.041}
\newcommand{\diagCocoKonePthree}{0.048}
\newcommand{\diagCocoKonePfour}{0.045}
\newcommand{\diagCocoKonePfive}{0.045}

\newcommand{\diagCocoKtwoPzero}{0.179}
\newcommand{\diagCocoKtwoPone}{0.203}
\newcommand{\diagCocoKtwoPtwo}{0.189}
\newcommand{\diagCocoKtwoPthree}{0.199}
\newcommand{\diagCocoKtwoPfour}{0.196}
\newcommand{\diagCocoKtwoPfive}{0.192}

\newcommand{\diagCocoKthreePzero}{0.607}
\newcommand{\diagCocoKthreePone}{0.641}
\newcommand{\diagCocoKthreePtwo}{0.641}
\newcommand{\diagCocoKthreePthree}{0.641}
\newcommand{\diagCocoKthreePfour}{0.641}
\newcommand{\diagCocoKthreePfive}{0.641}

\newcommand{\diagCocoLlKzeroPzero}{0.017}
\newcommand{\diagCocoLlKzeroPone}{0.038}
\newcommand{\diagCocoLlKzeroPtwo}{0.038}
\newcommand{\diagCocoLlKzeroPthree}{0.038}
\newcommand{\diagCocoLlKzeroPfour}{0.042}
\newcommand{\diagCocoLlKzeroPfive}{0.028}

\newcommand{\diagCocoLlKonePzero}{0.038}
\newcommand{\diagCocoLlKonePone}{0.031}
\newcommand{\diagCocoLlKonePtwo}{0.034}
\newcommand{\diagCocoLlKonePthree}{0.031}
\newcommand{\diagCocoLlKonePfour}{0.038}
\newcommand{\diagCocoLlKonePfive}{0.038}

\newcommand{\diagCocoLlKtwoPzero}{0.052}
\newcommand{\diagCocoLlKtwoPone}{0.045}
\newcommand{\diagCocoLlKtwoPtwo}{0.069}
\newcommand{\diagCocoLlKtwoPthree}{0.062}
\newcommand{\diagCocoLlKtwoPfour}{0.065}
\newcommand{\diagCocoLlKtwoPfive}{0.048}

\newcommand{\diagCocoLlKthreePzero}{0.197}
\newcommand{\diagCocoLlKthreePone}{0.179}
\newcommand{\diagCocoLlKthreePtwo}{0.205}
\newcommand{\diagCocoLlKthreePthree}{0.197}
\newcommand{\diagCocoLlKthreePfour}{0.154}
\newcommand{\diagCocoLlKthreePfive}{0.162}

\newcommand{\diagSimLlKzeroPzero}{0.490}
\newcommand{\diagSimLlKzeroPone}{0.135}
\newcommand{\diagSimLlKzeroPtwo}{0.090}
\newcommand{\diagSimLlKzeroPthree}{0.052}
\newcommand{\diagSimLlKzeroPfour}{0.052}
\newcommand{\diagSimLlKzeroPfive}{0.049}

\newcommand{\diagSimLlKonePzero}{0.328}
\newcommand{\diagSimLlKonePone}{0.462}
\newcommand{\diagSimLlKonePtwo}{0.179}
\newcommand{\diagSimLlKonePthree}{0.097}
\newcommand{\diagSimLlKonePfour}{0.069}
\newcommand{\diagSimLlKonePfive}{0.059}

\newcommand{\diagSimLlKtwoPzero}{0.052}
\newcommand{\diagSimLlKtwoPone}{0.344}
\newcommand{\diagSimLlKtwoPtwo}{0.625}
\newcommand{\diagSimLlKtwoPthree}{0.409}
\newcommand{\diagSimLlKtwoPfour}{0.340}
\newcommand{\diagSimLlKtwoPfive}{0.244}

\newcommand{\diagSimLlKthreePzero}{0.043}
\newcommand{\diagSimLlKthreePone}{0.094}
\newcommand{\diagSimLlKthreePtwo}{0.530}
\newcommand{\diagSimLlKthreePthree}{0.812}
\newcommand{\diagSimLlKthreePfour}{0.769}
\newcommand{\diagSimLlKthreePfive}{0.735}

\newcommand{\brdCocoPzero}{0.807}\newcommand{\brdCocoPone}{0.817}\newcommand{\brdCocoPtwo}{0.800}\newcommand{\brdCocoPthree}{0.813}\newcommand{\brdCocoPfour}{0.813}\newcommand{\brdCocoPfive}{0.810}

\newcommand{\brdSimPzero}{0.043}\newcommand{\brdSimPone}{0.220}\newcommand{\brdSimPtwo}{0.700}\newcommand{\brdSimPthree}{0.887}\newcommand{\brdSimPfour}{0.920}\newcommand{\brdSimPfive}{0.937}

\newcommand{\brdCocoLlPzero}{0.167}\newcommand{\brdCocoLlPone}{0.210}\newcommand{\brdCocoLlPtwo}{0.207}\newcommand{\brdCocoLlPthree}{0.203}\newcommand{\brdCocoLlPfour}{0.210}\newcommand{\brdCocoLlPfive}{0.193}
\newcommand{\brdCocoLlRPzero}{0.760}
\newcommand{\brdSimLlPzero}{0.043}\newcommand{\brdSimLlPone}{0.080}\newcommand{\brdSimLlPtwo}{0.580}\newcommand{\brdSimLlPthree}{0.883}\newcommand{\brdSimLlPfour}{0.967}\newcommand{\brdSimLlPfive}{0.963}

\newcommand{\cosSlotCocoGsmTone}{0.497}

\newcommand{\cosSlotCocoGsmTtwo}{0.552}

\newcommand{\cosSlotCocoGsmTthree}{0.619}

\newcommand{\cosSlotCocoGsmTfour}{0.516}

\newcommand{\cosSlotCocoGsmTfive}{0.582}

\newcommand{\cosSlotCodiGsmTone}{0.592}

\newcommand{\cosSlotCodiGsmTtwo}{0.611}

\newcommand{\cosSlotCodiGsmTthree}{0.629}

\newcommand{\cosSlotCodiGsmTfour}{0.639}

\newcommand{\cosSlotCodiGsmTfive}{0.648}

\newcommand{\cosSlotSimSixGsmTone}{0.504}

\newcommand{\cosSlotSimSixGsmTtwo}{0.608}

\newcommand{\cosSlotSimSixGsmTthree}{0.669}

\newcommand{\cosSlotSimSixGsmTfour}{0.554}

\newcommand{\cosSlotSimSixGsmTfive}{0.622}

\newcommand{\cosSlotCocoProsqaTone}{0.981}

\newcommand{\cosSlotCocoProsqaTtwo}{0.997}

\newcommand{\cosSlotCocoProsqaTthree}{0.999}

\newcommand{\cosSlotCocoProsqaTfour}{1.000}

\newcommand{\cosSlotCocoProsqaTfive}{1.000}

\newcommand{\cosSlotCodiProsqaTone}{1.000}

\newcommand{\cosSlotCodiProsqaTtwo}{1.000}

\newcommand{\cosSlotCodiProsqaTthree}{1.000}

\newcommand{\cosSlotCodiProsqaTfour}{1.000}

\newcommand{\cosSlotCodiProsqaTfive}{1.000}

\newcommand{\cosSlotSimProsqaTone}{0.687}

\newcommand{\cosSlotSimProsqaTtwo}{0.694}

\newcommand{\cosSlotSimProsqaTthree}{0.777}

\newcommand{\cosSlotSimProsqaTfour}{0.900}

\newcommand{\cosSlotSimProsqaTfive}{0.971}

\newcommand{\cosSlotCocoProntoTone}{0.823}

\newcommand{\cosSlotCocoProntoTtwo}{0.812}

\newcommand{\cosSlotCocoProntoTthree}{0.799}

\newcommand{\cosSlotCocoProntoTfour}{0.803}

\newcommand{\cosSlotCocoProntoTfive}{0.815}

\newcommand{\cosSlotCodiProntoTone}{1.000}

\newcommand{\cosSlotCodiProntoTtwo}{1.000}

\newcommand{\cosSlotCodiProntoTthree}{1.000}

\newcommand{\cosSlotCodiProntoTfour}{1.000}

\newcommand{\cosSlotCodiProntoTfive}{1.000}

\newcommand{\cosSlotSimProntoTone}{0.660}

\newcommand{\cosSlotSimProntoTtwo}{0.655}

\newcommand{\cosSlotSimProntoTthree}{0.676}

\newcommand{\cosSlotSimProntoTfour}{0.612}

\newcommand{\cosSlotSimProntoTfive}{0.803}

\newcommand{\cosSlotCocoGsmLlTone}{0.972}

\newcommand{\cosSlotCocoGsmLlTtwo}{0.985}

\newcommand{\cosSlotCocoGsmLlTthree}{0.989}

\newcommand{\cosSlotCocoGsmLlTfour}{0.992}

\newcommand{\cosSlotCocoGsmLlTfive}{0.993}

\newcommand{\cosSlotCodiGsmLlTone}{0.634}

\newcommand{\cosSlotCodiGsmLlTtwo}{0.626}

\newcommand{\cosSlotCodiGsmLlTthree}{0.458}

\newcommand{\cosSlotCodiGsmLlTfour}{0.562}

\newcommand{\cosSlotCodiGsmLlTfive}{0.816}

\newcommand{\cosSlotCocoProsqaLlTone}{0.976}

\newcommand{\cosSlotCocoProsqaLlTtwo}{0.993}

\newcommand{\cosSlotCocoProsqaLlTthree}{0.996}

\newcommand{\cosSlotCocoProsqaLlTfour}{0.998}

\newcommand{\cosSlotCocoProsqaLlTfive}{0.998}

\newcommand{\cosSlotCodiProsqaLlTone}{0.864}

\newcommand{\cosSlotCodiProsqaLlTtwo}{0.995}

\newcommand{\cosSlotCodiProsqaLlTthree}{0.999}

\newcommand{\cosSlotCodiProsqaLlTfour}{1.000}

\newcommand{\cosSlotCodiProsqaLlTfive}{1.000}

\newcommand{\cosSlotCocoProntoLlTone}{0.974}

\newcommand{\cosSlotCocoProntoLlTtwo}{0.991}

\newcommand{\cosSlotCocoProntoLlTthree}{0.996}

\newcommand{\cosSlotCocoProntoLlTfour}{0.998}

\newcommand{\cosSlotCocoProntoLlTfive}{0.999}

\newcommand{\cosSlotCodiProntoLlTone}{0.931}

\newcommand{\cosSlotCodiProntoLlTtwo}{0.999}

\newcommand{\cosSlotCodiProntoLlTthree}{1.000}

\newcommand{\cosSlotCodiProntoLlTfour}{1.000}

\newcommand{\cosSlotCodiProntoLlTfive}{1.000}

\newcommand{\cosSlotSimProsqaLlTone}{0.771}

\newcommand{\cosSlotSimProsqaLlTtwo}{0.803}

\newcommand{\cosSlotSimProsqaLlTthree}{0.757}

\newcommand{\cosSlotSimProsqaLlTfour}{0.849}

\newcommand{\cosSlotSimProsqaLlTfive}{0.939}

\newcommand{\cosSlotSimProntoLlTone}{0.990}

\newcommand{\cosSlotSimProntoLlTtwo}{0.996}

\newcommand{\cosSlotSimProntoLlTthree}{0.998}

\newcommand{\cosSlotSimProntoLlTfour}{0.999}

\newcommand{\cosSlotSimProntoLlTfive}{1.000}

\newcommand{\simSixGsmLlCzero}{0.390}        
\newcommand{\simSixGsmLlCone}{0.307}         
\newcommand{\simSixGsmLlHelped}{56}          
\newcommand{\simSixGsmLlHurt}{31}            
\newcommand{\simSixGsmLlKtwo}{0.383}         
\newcommand{\simSixGsmLlKfour}{0.387}        
\newcommand{\simSixGsmLlDnoise}{-0.093}      
\newcommand{\simSixGsmLlDmean}{-0.070}       
\newcommand{\simSixGsmLlDdonor}{-0.087}      
\newcommand{\simSixGsmLlRandNorm}{-0.080}    
\newcommand{\simSixGsmLlMeanEps}{-0.070}     
\newcommand{\simSixGsmLlPoisonPair}{44/43}   
\newcommand{\simSixGsmLlPoisonP}{1.000}      
\newcommand{\simSixGsmLlBoundN}{271000}      
\newcommand{\simSixGsmLlPoolN}{1319}         
\newcommand{\simSixGsmLlTwoByTwoSpread}{0.023} 
\newcommand{\cosSlotSimSixGsmLlTone}{0.930}      
\newcommand{\cosSlotSimSixGsmLlTtwo}{0.965}

\newcommand{\cosSlotSimSixGsmLlTthree}{0.972}

\newcommand{\cosSlotSimSixGsmLlTfour}{0.978}

\newcommand{\cosSlotSimSixGsmLlTfive}{0.985}

\newcommand{\accumFirstSimSixGsmLl}{0.363}       
\newcommand{\accumLastSimSixGsmLl}{0.487}        
\newcommand{\accumDeltaSimSixGsmLl}{0.123}       
\newcommand{\accumMddSimSixGsmLl}{0.112}         
\newcommand{\settleStepSimSixGsmLl}{2.470}       
\newcommand{\settleNullSimSixGsmLl}{4.650}       
\newcommand{\settleDeltaSimSixGsmLl}{2.18}       

\newcommand{\probeLayerGpt}{12}   
\newcommand{\probeLayerLl}{16}    

%% file: sections/0_abstract.tex
\begin{abstract}
\hyphenpenalty=3000 \exhyphenpenalty=3000
Latent chain-of-thought replaces written reasoning steps with continuous
hidden states, giving models additional steps to compute before answering.
It remains unclear what computation these steps actually perform. High task
accuracy and decodable intermediate answers do not establish that a model
uses its latent chain to reach an answer. We study this question across
twenty-one model-task configurations covering three training recipes, three
reasoning tasks, and two backbones. We combine chain deletion and content
replacement with probes and trajectory analysis to test whether the chain
affects the answer, what it encodes, and how its states change. On logical
tasks, \coconut{} often carries an answer that is already readable before
the chain begins. Its states settle early, and deleting the chain leaves
predictions almost unchanged. Per-step supervision produces an ordered
layout of reasoning steps and can make the model depend on the chain. Yet
even an ordered chain can act as a placeholder, with deletion hurting
performance while content replacement leaves answers unchanged. On
arithmetic tasks, all tested recipes show dependence on chain content.
These results show why readable latent states alone are insufficient
evidence of reasoning. We introduce a taxonomy that separates what a chain represents from how the model uses it. Seen this way, earlier reports no longer compete: placeholder tokens, collapsed superposition and step-dependent causal roles are each consistent with a different kind of chain, each obtained on a different task, training recipe or measurement. Once a study says which property it tested, whether the model uses the chain, what the chain holds, or how its state moves, claims about latent reasoning can be compared.
\end{abstract}

%% file: sections/1_intro_cur.tex
\section{Introduction}
\label{sec:intro}

Chain-of-thought gives a language model intermediate steps to work through
before answering \citep{yu2025stateaware,wei2026reasoningenergy}. Latent reasoning methods aim to do this without writing
out every step. In the methods we study, the model feeds continuous hidden
states back as inputs, forming a latent chain between the question and the
answer \citep{hao2024coconut,shen2025codi,wei2026simcot}. Each position in
this chain, which we call a slot, provides an additional computation step.
But what do these steps actually do? A model may use them to build an answer,
or it may already have enough information to answer after reading the
question. Distinguishing these cases matters for understanding the benefit
of latent steps and how to train models to use them. Task accuracy alone
cannot tell us where the answer was formed, and decoding an intermediate
result from a state does not show that the model uses it
\citep{elazar2021amnesic}. Prior studies report several behaviors, including
slots that act as placeholders \citep{zhang2025latenttokens}, candidate
superposition that collapses or goes unused after fine-tuning
\citep{rizvimartel2026illusion}, and different causal roles across steps
\citep{li2026dynamics}. These studies use different tasks, training recipes
and measurements, leaving open how their observations fit together.


\begin{figure}[t]
\centering
\includegraphics[width=\linewidth]{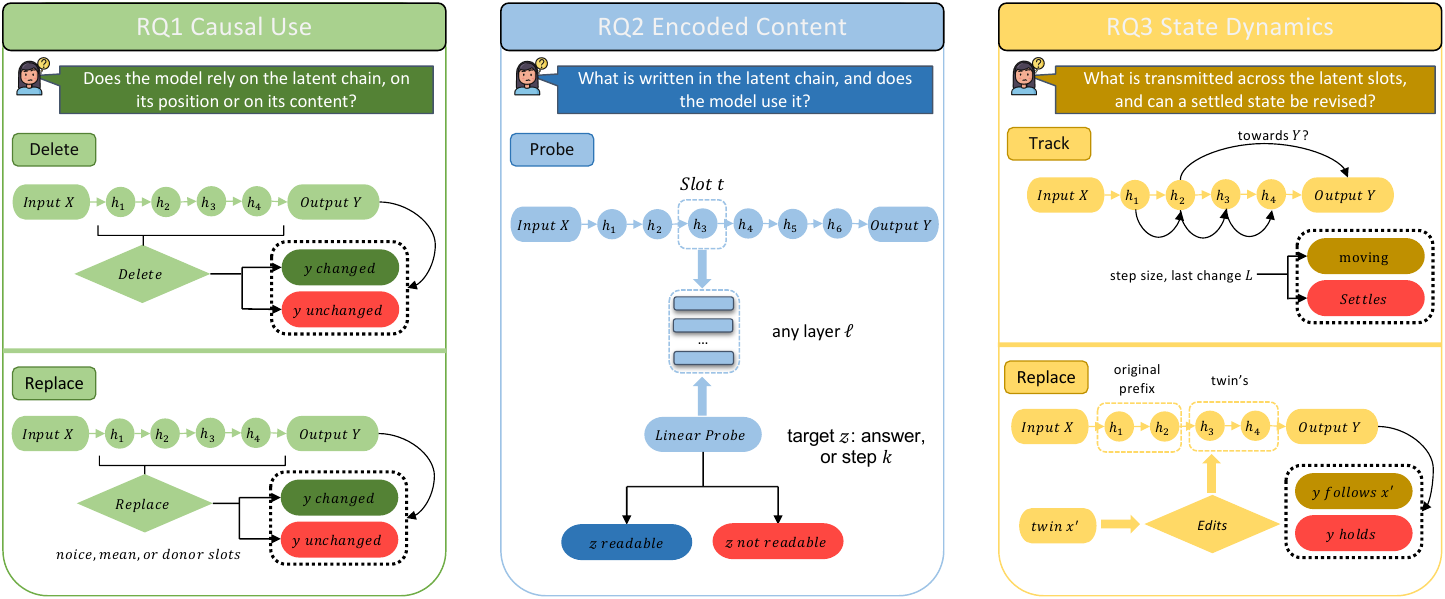}
\caption{\textbf{The three questions and the five tests, drawn on one latent
chain.} RQ1 deletes the slots or replaces their contents and reads the
answer. RQ2 probes one slot at one layer against a permuted-label null.
RQ3 tracks the state across slots and edits the chain mid-way. Answering all three lets us separate what a chain represents
from how the model uses it, and place the resulting behaviors in a single
taxonomy.}
\label{fig:overview}
\end{figure}

We study how latent chains contribute to answering across tasks and training
recipes within a common experimental framework. Our grid covers twenty-one
model-task configurations across \prosqa{}, \prontoqa{} and \gsm{}, using
GPT-2 and \llamaB{}. We compare \coconut{}, which recycles hidden states
without direct supervision of slot content, \codi{}, which aligns latent
and written reasoning at the answer position, and \simcoco{}, which adds
per-step supervision to \coconut{}. The last comparison holds the inference
mechanism and slot budget fixed. Continued-training controls help separate
the effect of per-step supervision from that of extra training. On this
grid, we ask three questions about causal use, encoded content and state
dynamics (Figure~\ref{fig:overview}).

\textbf{RQ1: Does the model rely on the latent chain, on its positions or on
its contents?} We first test whether the model depends on the chain by
deleting its slots. Where prior causal tests apply a single corruption (deletion \citep{dilgren2026interpretable} or
noise \citep{cui2026supervision}) each,
we pair deletion with three replacements (noise, the dataset mean, and
another problem's vectors) plus a same-base training-time control that
switches off the per-step loss (Section~\ref{sec:determination}). This
isolates a placeholder pattern a single corruption cannot: on one backbone's logical-task cell, deletion costs accuracy yet no replacement moves an
answer, while on most content-used cells another problem's slots hurt more
than deleting the chain outright. Deleting the unsupervised recipe's chains changes at most
\inertBound{} answers on the logical tasks but hurts accuracy on arithmetic.
Per-step supervision can establish reliance on a chain where the same base
model had little.

\textbf{RQ2: What is written in the latent chain, and does the model use
it?} A chain that does not affect the answer may still contain information
about it. Prior probing has asked one question at a time, whether the answer
is decodable early \citep{rizvimartel2026illusion,li2026dynamics} or whether a slot superposes candidates \citep{rizvimartel2026illusion,aswal2026observable}; we probe for the
final answer, the candidates and each reasoning step, at every slot and
layer, and read each profile against its RQ1 verdict
(Section~\ref{sec:static}). On the logical tasks, unsupervised chains carry
an answer that is already readable at the question's final hidden state,
before the chain begins, and hold the path spread across slots rather than
in order; no slot holds both candidates. Per-step supervision lays the steps
out one per slot, and that ordered layout appears even in a chain whose
contents can be replaced without changing an answer.
Placed beside RQ1, this gives the paper's central observation: on the
logical tasks the chains with the richest readable content are ones the
model does not rely on, and an ordered layout can sit in a chain whose
contents are never consulted.

\textbf{RQ3: What is transmitted across the latent slots, and can a settled
state be revised?} Knowing what individual slots contain leaves open how the
state evolves between them. State movement has been described by trajectory
geometry and answer commitment \citep{duraipandian2026dynamical,aswal2026observable,li2026dynamics} but not
decided against a null for whether a chain settles; we turn it into a decision across all recipes and
both backbones, and add a mid-chain edit that splices a twin problem's
content into the chain (Section~\ref{sec:dynamics}). On the
logical tasks, chains without per-step supervision settle within the first
slots, while per-step chains keep changing to the last slot, though the
movement is directed toward the answer on only one task. The edit, a test the prior
studies do not include, shows that revising an answer is set by the
backbone, not the mechanism or the supervision: one backbone increasingly ignores
later edits while the other follows them. State movement and the ability to
revise an answer thus describe different aspects of the chain.

These results show why a single measurement gives an incomplete account of
latent reasoning. A chain can carry a readable answer without affecting the
output, or encode ordered steps while serving only as a set of extra
positions under the tested interventions. We combine the three measurements
in a taxonomy that distinguishes six types of latent chain, while leaving
unresolved cases unclassified (Section~\ref{sec:taxonomy}). Its value is that the scattered reports above no longer appear to make conflicting claims about latent chains in general. Placeholder behaviour, collapsed superposition, and step-dependent causal roles are each consistent with a different kind of chain, arising under different tasks, training recipes, or measurements. By stating what each test establishes about a chain, the taxonomy explains why earlier findings seem to disagree and clarifies which property—use, content, or dynamics—a claim about latent reasoning addresses.

%% file: sections/2_related.tex
\section{Related Work}
\label{sec:related}

\textbf{Latent reasoning methods.}
Written chains are costly both to train, as post-training samples and
scores many rollouts per problem
\citep{surana2026rollout,mundada2026wsgrpo,xiong2026rrpo,surana2026massdpo},
and to decode token by token \citep{lu2024multicandidate}.
Several approaches give language models more computation before an answer
without requiring a full written chain. Pause and filler tokens provide
extra positions without explicit reasoning content
\citep{goyal2024pause,pfau2024dotbydot}, and learned soft prompts likewise
feed continuous vectors rather than tokens \citep{wu2023infoprompt}.
Recurrent-depth models apply additional computation within the network
\citep{geiping2025recurrent}.
Other methods train models to internalize written reasoning or compress it
into continuous representations \citep{deng2024implicit,cheng2024ccot}.
We focus on methods that feed hidden states back as inputs to form a chain
of latent slots. \coconut{} gradually replaces written steps with recycled
hidden states during training \citep{hao2024coconut}. \codi{} uses
self-distillation to align the latent pathway with written reasoning at the
answer position \citep{shen2025codi}. \simcot{} adds direct supervision of
individual slots through an auxiliary decoder \citep{wei2026simcot}. These
training signals offer different ways to shape a latent chain. Comparing
\coconut{} with its per-step supervised variant lets us test how slot
supervision changes the chain while keeping its inference mechanism fixed.

\textbf{Interpreting latent states.}
Prior work has looked for recognizable reasoning structure inside continuous
states. \citet{hao2024coconut} interpret vocabulary projections of
\coconut{} states as evidence of breadth-first search, and
\citet{zhu2025superposition} study how continuous thoughts can represent
multiple search paths. Work on \codi{} and \simcot{} examines whether
intermediate reasoning content can be recovered from latent states
\citep{shen2025codi,wei2026simcot}. These interpretations depend on how the
model is trained. For example, \citet{rizvimartel2026illusion} find that
candidate superposition can collapse after fine-tuning. Analyses of state
transitions also provide ways to study how reasoning evolves across steps
\citep{yu2025stateaware,wu2025ctrls}. Such measurements, like information-theoretic analyses of what
representations carry \citep{huang2025traceable}, describe the information and
structure present in a trajectory, but their relevance to the answer needs
separate evidence. This distinction follows a broader lesson from probing:
a target can be predictable from a representation without being used for
the model's task \citep{ravichander2021probing,elazar2021amnesic}. We pair
content and trajectory measurements with interventions on the same model
and task configurations.

\textbf{Causal use of latent chains.}
Studies of written chain-of-thought evaluate reasoning traces offline
\citep{wu2025ocean} and use interventions to test, or correct, how traces
affect model answers \citep{lanham2023faithfulness,wu2024decot}.
Interventions on hidden states can also steer reasoning, for example to
break repetitive self-loops \citep{yu2026selfloops}, elicit latent
skills \citep{jiang2026latentskills} or adjust modality preference
\citep{huang2026amps}; we use them only as diagnostics.
For latent chains, \citet{dilgren2026interpretable} study both trace
readability and necessity, finding that removing latent steps can preserve
performance on logical tasks. \citet{zhang2025latenttokens} report
insensitivity to content interventions and describe a placeholder role,
while \citet{li2026dynamics} find different causal roles across steps and
information flow between nonadjacent positions. Closely related to our
study, \citet{aswal2026observable} combine matched controls, causal
interventions and trajectory analysis. They show that observable reasoning
patterns can also appear in controls, and that behavioral influence can be
concentrated in a small subspace even when the full states appear nearly
static. 

%% file: sections/3_setup.tex
\section{Experimental Setup}
\label{sec:setup}

\subsection{Latent-Chain Formulation}
\label{sec:probing}
\label{sec:notation}

Given a question $x$, a latent-reasoning model with parameters $\theta$
produces $T$ continuous states before decoding an answer. We index these
slots from zero to match the recorded trajectories. Let
$\mathbf h_t^{(\ell)}\in\mathbb R^d$ be the hidden state at layer $\ell$ of
slot $t$, and let $\mathbf h_t$ denote its final-layer state. The latent
chain and answer distribution can be written as
\begin{equation}
\mathbf h_t=F_\theta\!\left(x,\mathbf h_{0:t-1}\right),
\qquad p_\theta\!\left(y\mid x,\mathbf h_{0:T-1}\right).
\label{eq:chain}
\end{equation}
Here $F_\theta$ includes the model-specific mapping from a hidden state to
the next input. This is the identity map for \coconut{} and includes a
projection network for \codi{}. The model can attend to the question as
well as earlier slots, so the notation does not assume that each state
depends only on the preceding one. We write $y^\star$ for the gold answer
and $z_k$ for step $k$ of an annotated reasoning path, where available.
Probe targets $z$ are defined for each task in RQ2.

\subsection{Models, Tasks, and Controls}
\label{sec:models}
\label{sec:grid}
\label{sec:tasks}
\label{sec:paradigms}

We compare three training recipes on GPT-2 and \llamaB{}.
\coconut{} recycles hidden states without a direct loss on slot content
\citep{hao2024coconut}. \codi{} aligns the hidden state at the answer
position with that of its written reasoning pathway \citep{shen2025codi}.
\simcoco{} adds the per-step supervision of \simcot{} to \coconut{}:
an auxiliary decoder reconstructs written step $k$ from slot $k$ during
training and is discarded at inference \citep{wei2026simcot}. All main-grid
models use $T=\slotsCoconut{}$ slots. Table~\ref{tab:models} summarizes
the input mapping, supervision and checkpoint coverage of each recipe.

The tasks cover proof search, syllogisms and arithmetic. \prosqa{} asks
which of two candidate concepts follows from a set of facts and rules
\citep{hao2024coconut}. \prontoqa{} asks yes/no questions about logical
implications \citep{saparov2023prontoqa}. \gsm{} contains arithmetic word
problems with numerical answers \citep{cobbe2021gsm8k}.

For \coconut{} and \codi{}, we use the public checkpoints of
\citet{dilgren2026interpretable} on both backbones. We also include a second
independently trained public \coconut{}/\prosqa{} checkpoint on GPT-2.
We train \simcoco{} on all three tasks and both backbones at the matched
slot budget. On \prosqa{}, a control on each backbone continues the same
\coconut{} base for the same training schedule with the auxiliary loss
weighted to zero. These controls separate the effect of per-step
supervision from extra training. The grid contains twenty-one cells, where
a cell is one checkpoint evaluated on one task. Checkpoint sources and
training details are in App.~\ref{app:protocol}. The \llamaB{}/\prontoqa{}
per-step checkpoint was selected by a saved sweep before interventions,
as described in App.~\ref{app:llamaperstep}.

The \coconut{} and \simcoco{} comparison holds the inference mechanism
and slot budget fixed. Comparisons involving \codi{} change both mechanism
and supervision, so they do not isolate either factor. The two backbones
differ in scale and instruction tuning as well as model family. We report
them separately and interpret their differences as backbone effects.

\subsection{Evaluation Protocol}
\label{sec:rationale}
\label{sec:standards}

Every cell is evaluated on the same frozen set of $N=\Neval{}$ held-out
problems for its task. Public checkpoints are checked against their
reported accuracy before analysis. Our trained checkpoints are anchored to
the base models they continue. Split differences and parity checks are
reported in App.~\ref{app:protocol}.

We keep the model parameters fixed during evaluation. Following the
interventional view of causal analysis in LLMs \citep{liu2025causalsurvey},
an intervention $\mathcal I$ changes the latent computation during the forward pass,
giving the clean and intervened predictions
\begin{equation}
\hat y=\operatorname{Decode}_\theta(x),\qquad
\hat y^{\mathcal I}=\operatorname{Decode}_\theta(x;\mathcal I).
\label{eq:interventions}
\end{equation}
Deletion skips latent slots. Content replacement substitutes the vectors
fed into the latent loop while keeping the slot positions, and subsequent
states are computed under that intervention. Mid-chain edits use the
separate protocol described in RQ3. We compare predictions on the same
problems and report both accuracy and changes in individual answers,
since unchanged average accuracy can hide changes in opposite directions.

Probes are trained and evaluated with folds split by problem, so states
from one problem do not appear in both training and test folds. Probe
readouts are compared with permuted-label nulls, and trajectory statistics
with the nulls specified in RQ3. Each test reports its minimum detectable
effect (MDE). Effects too small to resolve are kept separate from bounded
null results. Confirming and falsifying criteria were recorded before the
first forward pass of each experiment, with later amendments retained
alongside the originals. The following sections define the measurements
for each RQ, and the appendices give their settings and per-cell results.

%% file: sections/4_rq1_determination.tex
\begin{table*}[t]
\centering\footnotesize\setlength{\tabcolsep}{3pt}
\caption{\textbf{Deleting the chain and replacing its contents} on the
$3{\times}3$ grid of both backbones, with the outcome of each test.}
\label{tab:deletion}
\resizebox{\textwidth}{!}{%
\begin{tabular}{@{}ll ccc l ccc l@{}}
\toprule
& & \multicolumn{4}{c}{\textbf{Delete the chain}} & \multicolumn{4}{c}{\textbf{Replace the contents ($\Delta$ accuracy)}} \\
\cmidrule(lr){3-6}\cmidrule(lr){7-10}
\textbf{Task} & \textbf{Recipe} & $C_0$ & $C_1$ & helped/hurt & verdict & noise & mean & donor & verdict \\
\midrule
\rowcolor{gray!10}[0pt][0pt]\multicolumn{10}{@{}l@{}}{\textbf{GPT-2}} \\
\gsm{}      & \coconut{} & \cocogsmCzero & \cocogsmCone & \cocogsmHelped/\cocogsmHurt & \vConsulted & $\cocogsmDnoise$ & $\cocogsmDmean$ & $\cocogsmDdonor$ & \vContent \\
            & \codi{}    & \codigsmCzero & \codigsmCone & \codigsmHelped/\codigsmHurt & \vConsulted & $\codigsmDnoise$ & $\codigsmDmean$ & $\codigsmDdonor$ & \vContent \\
            & \simcoco{} & \simSixGsmCzero & \simSixGsmCone & \simSixGsmHelped/\simSixGsmHurt & \vConsulted & $\simSixGsmDnoise$ & $\simSixGsmDmean$ & $\simSixGsmDdonor$ & \vContent \\
\addlinespace[2pt]
\prosqa{}   & \coconut{} & \cocoprosqadwCzero & \cocoprosqadwCone & \cocoprosqadwHelped/\cocoprosqadwHurt & \vInert & $\cocoprosqadwDnoise$ & $\cocoprosqadwDmean$ & $\cocoprosqadwDdonor$ & \textcolor{lcMuted}{\emph{not reached}} \\
            & \codi{}    & \codiprosqaCzero & \codiprosqaCone & \codiprosqaHelped/\codiprosqaHurt & \vUnresolved & $\codiprosqaDnoise$\textsuperscript{a} & $\codiprosqaDmean$ & $\codiprosqaDdonor$ & \textcolor{lcMuted}{\emph{not reached}} \\
            & \simcoco{} & \simProsqaCzero & \simProsqaCone & \simProsqaHelped/\simProsqaHurt & \vConsulted & $\simProsqaDnoise$ & $\simProsqaDmean$ & $\simProsqaDdonor$ & \vContent \\
\addlinespace[2pt]
\prontoqa{} & \coconut{} & \cocoprontoCzero & \cocoprontoCone & \cocoprontoHelped/\cocoprontoHurt & \vInert & $\cocoprontoDnoise$ & $\cocoprontoDmean$ & $\cocoprontoDdonor$ & \textcolor{lcMuted}{\emph{not reached}} \\
            & \codi{}    & \codiprontoCzero & \codiprontoCone & \codiprontoHelped/\codiprontoHurt & \vInert & $\codiprontoDnoise$ & $\codiprontoDmean$ & $\codiprontoDdonor$ & \textcolor{lcMuted}{\emph{not reached}} \\
            & \simcoco{} & \simProntoCzero & \simProntoCone & \simProntoHelped/\simProntoHurt & \vConsulted & $\simProntoDnoise$ & $\simProntoDmean$ & $\simProntoDdonor$ & \vContent \\
\addlinespace[2pt]
\rowcolor{gray!10}[0pt][0pt]\multicolumn{10}{@{}l@{}}{\textbf{\llamaB{}}} \\
\gsm{}      & \coconut{} & \cocogsmLlCzero & \cocogsmLlCone & \cocogsmLlHelped/\cocogsmLlHurt & \vConsulted & $\cocogsmLlDnoise$ & $\cocogsmLlDmean$ & $\cocogsmLlDdonor$ & \vContent \\
            & \codi{}    & \codigsmLlCzero & \codigsmLlCone & \codigsmLlHelped/\codigsmLlHurt & \vConsulted & $\codigsmLlDnoise$ & $\codigsmLlDmean$ & $\codigsmLlDdonor$ & \vContent \\
            & \simcoco{} & \simSixGsmLlCzero & \simSixGsmLlCone & \simSixGsmLlHelped/\simSixGsmLlHurt & \vConsulted & $\simSixGsmLlDnoise$ & $\simSixGsmLlDmean$ & $\simSixGsmLlDdonor$ & \vContent \\
\addlinespace[2pt]
\prosqa{}   & \coconut{} & \cocoprosqaLlCzero & \cocoprosqaLlCone & \cocoprosqaLlHelped/\cocoprosqaLlHurt & \vInert & $\cocoprosqaLlDnoise$ & $\cocoprosqaLlDmean$ & $\cocoprosqaLlDdonor$ & \textcolor{lcMuted}{\emph{not reached}} \\
            & \codi{}    & \codiprosqaLlCzero & \codiprosqaLlCone & \codiprosqaLlHelped/\codiprosqaLlHurt & \vConsulted & $\codiprosqaLlDnoise$ & $\codiprosqaLlDmean$ & $\codiprosqaLlDdonor$ & \vPlaceholder \\
            & \simcoco{} & \simProsqaLlCzero & \simProsqaLlCone & \simProsqaLlHelped/\simProsqaLlHurt & \vConsulted & $\simProsqaLlDnoise$ & $\simProsqaLlDmean$ & $\simProsqaLlDdonor$ & \vPlaceholder \\
\addlinespace[2pt]
\prontoqa{} & \coconut{} & \cocoprontoLlCzero & \cocoprontoLlCone & \cocoprontoLlHelped/\cocoprontoLlHurt & \vInert & $\cocoprontoLlDnoise$ & $\cocoprontoLlDmean$ & $\cocoprontoLlDdonor$ & \textcolor{lcMuted}{\emph{not reached}} \\
            & \codi{}    & \codiprontoLlCzero & \codiprontoLlCone & \codiprontoLlHelped/\codiprontoLlHurt & \vInert & $\codiprontoLlDnoise$ & $\codiprontoLlDmean$ & $\codiprontoLlDdonor$ & \textcolor{lcMuted}{\emph{not reached}} \\
            & \simcoco{} & \simProntoLlCzero & \simProntoLlCone & \simProntoLlHelped/\simProntoLlHurt & \vInert & $\simProntoLlDnoise$ & $\simProntoLlDmean$ & $\simProntoLlDdonor$ & \textcolor{lcMuted}{\emph{not reached}} \\
\bottomrule
\end{tabular}}\par\vspace{2pt}
\begin{minipage}{\textwidth}\footnotesize
\textsuperscript{a}\,An off-manifold effect: random noise breaks the readout, while the
  dataset mean and donor vectors move no answer ($\Delta{=}0$,
  App.~\ref{app:substitution}).
\end{minipage}
\end{table*}

\section{RQ1: Does the Model Rely on the Latent Chain, on Its Positions or on Its Contents?}
\label{sec:determination}
\label{sec:a1}
\label{sec:a2}
\label{sec:a3}
\label{sec:replication}

A latent chain gives the model extra computation before answering, but its
presence does not establish that the model needs it. Even when deleting the
chain hurts performance, the loss could reflect missing computation steps
rather than missing problem-specific content. We therefore test two forms
of dependence: whether the model needs the latent slots, and whether its
answers are sensitive to the vectors supplied at those slots.

\textbf{Interventions and readouts.} For each cell, we evaluate the same
\Neval{} problems with the model parameters and question held fixed.
Deletion skips every latent slot and decodes the answer. This tests whether
the chain is necessary under that intervention, but removes both its
contents and its computation steps. Substitution keeps the slot positions
and replaces the vectors fed into the loop with norm-matched Gaussian
noise, the dataset-mean vector, or recorded vectors from another problem
at the same positions. Subsequent states are recomputed under the
intervention. These tests measure how replacement inputs affect the
remaining computation and the answer. Noise can move states away from their
natural distribution, so noise sensitivity alone is not evidence that the
model uses problem-specific content. Mean and donor replacements provide
complementary tests, with direction-by-scale controls on selected cells
(App.~\ref{app:twobytwo}). 
Pairing deletion with substitution separates two patterns a single corruption cannot: cells where a donor's contents are more disruptive than removing the chain outright, and cells that resist deletion but are also immune to any replacement (the placeholder signature reported below).

We report intact and deleted accuracy ($C_0$, $C_1$), prediction changes,
and paired counts of problems \emph{helped} by the chain (correct only when
intact) and \emph{hurt} (correct only when deleted). These counts distinguish
unchanged predictions from changes in opposite directions that cancel in
the average. In Table~\ref{tab:deletion}, \emph{consulted} means deletion
reduces accuracy beyond the cell's MDE, and \emph{inert} means at most
\inertBound{} answers change. Other deletion outcomes are unresolved.
Among consulted cells, \emph{content used} denotes sensitivity to
replacement beyond the MDE, while \emph{placeholder} denotes no prediction
changes under any tested replacement. Smaller unresolved effects are not
assigned to either outcome. Full thresholds and per-cell tests are in
App.~\ref{app:rq1}. An inert outcome does not prove that the chain has no
role in the intact forward pass, just as replacement sensitivity does not
by itself establish step-by-step reasoning.

\textbf{Finding 1: Dependence on the chain varies across tasks.}
On GPT-2, deleting \coconut{}'s chain leaves \prosqa{} accuracy unchanged
at \cocoprosqadwCzero{}, but reduces \gsm{} accuracy from
\cocogsmCzero{} to \cocogsmCone{} (Table~\ref{tab:deletion}). The same
logical-versus-arithmetic split appears on \llamaB{}. \codi{} also depends
on the chain on \gsm{} and has little deletion effect on \prontoqa{}.
Its \prosqa{} behavior differs by backbone. On GPT-2, deletion loses
\codiprosqaHelped{} correct answers and gains \codiprosqaHurt{}, leaving
the direction unresolved despite the small average gap. On \llamaB{},
deletion causes a clear loss. These results show that the training recipe
alone does not specify whether the chain is needed.

\textbf{Finding 2: Per-step supervision can establish dependence on the
chain.} On GPT-2, \simcoco{} depends on its chain on both logical tasks,
where the \coconut{} base has little deletion effect. On \prosqa{},
accuracy falls from \simProsqaCzero{} to \simProsqaCone{}, with
\simProsqaHelped{} correct answers lost. The continued-training control
remains inert, supporting an effect of the per-step loss beyond extra
training time (App.~\ref{app:controls}). Truncation provides a further
check: in every consulted GPT-2 cell, accuracy rises monotonically as more
slots are retained (App.~\ref{app:truncation}). The \llamaB{}/\prosqa{}
per-step model also depends on its chain, while its continued-training
control has an unresolved deletion effect below its MDE. This pattern is
not universal. The selected \llamaB{}/\prontoqa{} per-step checkpoint
remains inert, and the \llamaB{}/\gsm{} per-step cell has no matched
training-time control.

\textbf{Finding 3: Dependence on slots does not always imply sensitivity
to their contents.} All tested \gsm{} recipes on both backbones are
sensitive to content replacement. Donor vectors can be more harmful than
removing the chain altogether. For GPT-2 \coconut{}/\gsm{}, the accuracy
change is $\cocogsmDdonor$ under donor replacement, compared with
$-\cocogsmGap$ under deletion. This supports dependence on the supplied
vectors beyond the presence of extra slots. The ordering is not required
for content sensitivity. On GPT-2 \simcoco{}/\prosqa{}, donor replacement
hurts less than deletion, even though the replacement effect is detectable.
Other donor-versus-deletion contrasts remain bounded by their MDE
(App.~\ref{app:substitution}).
Across the eight content-used cells (Table~\ref{tab:corruption}), donor replacement is more harmful than deletion in five, less harmful in one, and undetermined in two.

The \llamaB{}/\prosqa{} \codi{} and per-step cells show a different
pattern. Deletion hurts performance, but no tested replacement changes an
answer, including a different-answer donor at twice the natural norm.
This is placeholder-like behavior, consistent with the benefit of extra
positions studied by \citet{goyal2024pause,pfau2024dotbydot}. It is bounded
by the tested interventions and their sensitivity, since no replacement
moves an answer in any \llamaB{}/\prosqa{} cell. The model also retains
access to the question during replacement, so this result does not imply
that its later computation uses no problem information.

The model's dependence on its latent chain varies with task, supervision
and backbone. Deletion and replacement distinguish chains that can be
skipped, chains whose supplied vectors affect the answer, and chains that
show dependence only when the slots are removed. These tests establish
behavioral effects under intervention. RQ2 next examines what information
the states contain, including in chains that are not necessary to answer.

%% file: sections/5_rq2_static.tex
\section{RQ2: {What Is Written in the Latent Chain, and Does the Model Use It?}}
\label{sec:static}
\label{sec:s1}
\label{sec:s2}
\label{sec:s3acc}
\label{sec:s3}

RQ1 showed which chains the model reads, but not what they contain. An
unused chain may still encode information, and the main accounts make
different predictions: superposition puts multiple candidate answers in the
same slot \citep{hao2024coconut,zhu2025superposition}; stepwise reasoning
lays the path out in order, one step per slot
\citep{wei2026simcot,deng2024implicit}; and computed-at-the-question
reasoning predicts that the answer is already present before the chain
begins \citep{pal2023futurelens,yang2024multihop}.

\begin{figure*}[t]
\centering
\input{figures/fig_layout}
\caption{\textbf{Linear probe accuracy across slots} on \prosqa{}. Upper
rows: path steps $k\in\{0,\dots,3\}$;
bottom row: final answer; shared colour scale. Probes read each slot's
final-layer state, the vector fed back as the next input (layer
\probeLayerGpt{} on GPT-2, \probeLayerLl{} on \llamaBshort{}). Hatched:
projection artefact of the record probe on the \llamaBshort{} \coconut{}
answer row (App.~\ref{app:depth}).}
\label{fig:layout}
\label{fig:static}
\end{figure*}
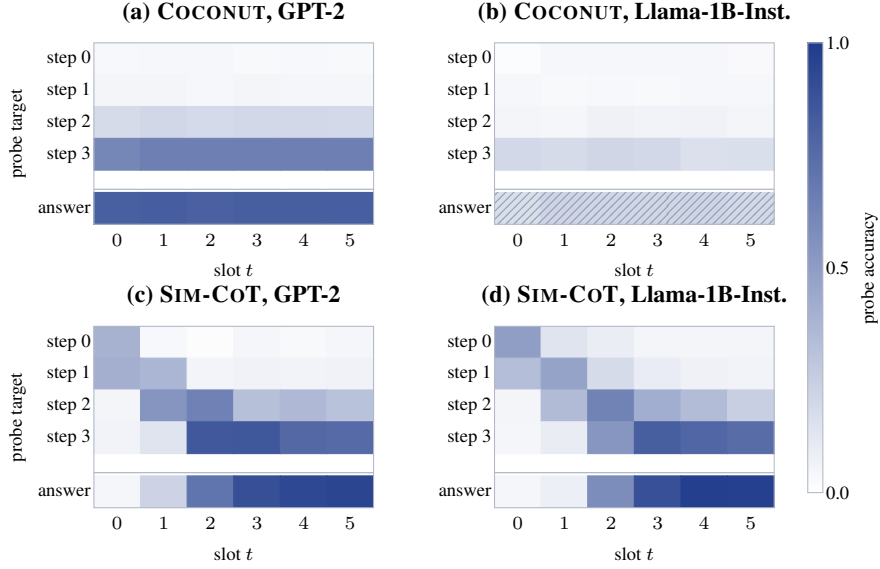

\textbf{Setup.} We fit a linear probe $p_{t,\ell}$ to predict a target
$z$ from $\mathbf h_t^{(\ell)}$. Targets include the final answer concept
and gold path step $z_k$ on \prosqa{}, and a magnitude bin of the numerical
answer on \gsm{}. We report final-layer readouts by slot here and the full
layer sweep in App.~\ref{app:sweep}. Probe accuracy is measured on held-out
problems using the cross-validation protocol of Section~\ref{sec:standards}.
A target is called readable only when its accuracy exceeds the
permuted-label null by more than the cell's MDE. We do not compare amounts
of readable content across backbones. The probe of record uses
\probepca{} principal components followed by logistic regression.
On \prosqa{}, we also fit a full-dimension ridge probe and state an
unqualified finding only where both agree. A nonlinear probe at the same
points provides a further check (largest gain \mlpMaxGain{}).
These readouts test access to information in a state. They do not establish
that the model uses that information to answer.

\textbf{Finding 4: The unsupervised chain carries an answer that is already
readable before the chain begins; per-step supervision distributes
computation across the slots.} The answer rows of Figure~\ref{fig:layout}
show two regimes on both backbones. Inert \coconut{} chains read the answer
from the first slot as well as from the last, while per-step chains start
near chance and build the answer monotonically toward the last slot. The hatched \llamaB{} \coconut{}
row is a projection artefact of the probe of record; the ridge probe reads
\brdCocoLlRPzero{} from slot 0. The answer is
already readable at the question's final hidden state, before any latent
slot is written, on both independent trainings (App.~\ref{app:depth}): the
unsupervised chain is a carrier rather than a workspace, consistent with
RQ1's inert verdict and with prior work showing that answers form during
the question pass \citep{pal2023futurelens,yang2024multihop}. On \gsm{}
the answer's magnitude becomes more readable along the chain in the GPT-2
\coconut{}-mechanism cells and on the \llamaB{} per-step twin
(App.~\ref{app:accum}).

\textbf{Finding 5: Per-step supervision installs an ordered layout, with
step $k$ written into slot $k$.} In the supervised panels, the step rows are
diagonal on both backbones: step $k$ becomes readable at slot $k$; early
steps fade along the chain, while later steps and the answer persist to the
last slot. In both probe arms, no node off the gold path is readable, even
when off-path nodes are matched to on-path nodes in graph degree
(App.~\ref{app:layout}). The training-time control, trained equally
long without the per-step loss, shows no ordered path (App.~\ref{app:layout}).
This ordered layout is not restricted to chains the model consults: the \llamaB{} \simcoco{}/\prosqa{} cell shows the same diagonal structure while RQ1 found its contents replaceable without changing the answer (Section~\ref{sec:determination}).

\textbf{Finding 6: The unsupervised chain stores the path uniformly, not
step by step: the chains with the richest readable content are inert.} The
unsupervised step rows are flat. Early steps are not linearly readable from
any slot, and the readable steps read about equally from every slot, on
both backbones and on an independent training, the opposite of the frontier
expansion a search account predicts. Nor does any \prosqa{} cell hold both
candidates at once: the highest both-readable rate in either arm is
\breadthMaxPeakR{}, against a bar of \superposthresh{}
(App.~\ref{app:candidates}). Answer-aligned chains hold little of the path
at all (App.~\ref{app:layout}).

Together, these results show that content follows the training objective: unsupervised chains carry a
finished answer across the slots, and per-step supervision lays the path out one step per slot. 

%% file: figures/fig_layout.tex
\providecolor{lcNavy}{HTML}{1E3A8A}%
\providecolor{lcSlate}{HTML}{64748B}%
\providecolor{lcCrimson}{HTML}{E11D48}%
\begin{tikzpicture}
\pgfplotsset{
  hmbase/.style={
    scale only axis, width=3.7cm, height=2.4cm, line width=0.4pt,
    tick style={draw=none}, axis line style={lcSlate!50},
    label style={font=\scriptsize}, tick label style={font=\scriptsize}, y tick label style={font=\scriptsize, inner sep=1pt},
    title style={font=\footnotesize, yshift=-2pt},
    xtick={0,...,5}, ytick={0,1,2,3,4.7}, yticklabels={step 0,step 1,step 2,step 3,answer}, xmin=-0.5, xmax=5.5, ymin=-0.5, ymax=5.2,
    y dir=reverse, enlargelimits=false, axis on top,
    colormap={wp}{color(0cm)=(white); color(1cm)=(lcNavy)},
    point meta min=0, point meta max=1,
  },
}
\begin{groupplot}[group style={group size=2 by 2, horizontal sep=1.6cm, vertical sep=1.35cm,
  ylabels at=edge left}, hmbase,
  xlabel={slot $t$}, ylabel={probe target}]
\nextgroupplot[title={\textbf{(a) \coconut{}, GPT-2}}]
\addplot[matrix plot*, mesh/cols=6, point meta=explicit] table[x=t,y=k,meta=v] {
t k v
0 0 \diagCocoKzeroPzero
1 0 \diagCocoKzeroPone
2 0 \diagCocoKzeroPtwo
3 0 \diagCocoKzeroPthree
4 0 \diagCocoKzeroPfour
5 0 \diagCocoKzeroPfive
0 1 \diagCocoKonePzero
1 1 \diagCocoKonePone
2 1 \diagCocoKonePtwo
3 1 \diagCocoKonePthree
4 1 \diagCocoKonePfour
5 1 \diagCocoKonePfive
0 2 \diagCocoKtwoPzero
1 2 \diagCocoKtwoPone
2 2 \diagCocoKtwoPtwo
3 2 \diagCocoKtwoPthree
4 2 \diagCocoKtwoPfour
5 2 \diagCocoKtwoPfive
0 3 \diagCocoKthreePzero
1 3 \diagCocoKthreePone
2 3 \diagCocoKthreePtwo
3 3 \diagCocoKthreePthree
4 3 \diagCocoKthreePfour
5 3 \diagCocoKthreePfive
};
\addplot[matrix plot*, mesh/cols=6, point meta=explicit] table[x=t,y=k,meta=v] {
t k v
0 4.7 \brdCocoPzero
1 4.7 \brdCocoPone
2 4.7 \brdCocoPtwo
3 4.7 \brdCocoPthree
4 4.7 \brdCocoPfour
5 4.7 \brdCocoPfive
0 5.7 \brdCocoPzero
1 5.7 \brdCocoPone
2 5.7 \brdCocoPtwo
3 5.7 \brdCocoPthree
4 5.7 \brdCocoPfour
5 5.7 \brdCocoPfive
};
\draw[lcSlate!60, line width=0.4pt] (axis cs:-0.5,4.1) -- (axis cs:5.5,4.1);
\nextgroupplot[title={\textbf{(b) \coconut{}, \llamaBshort{}}}, colorbar, colorbar style={y dir=normal, width=0.2cm, at={(1.12,1.0)}, anchor=north west, height=5.95cm, ytick={0,0.5,1}, yticklabels={0.0,0.5,1.0}, ylabel={probe accuracy}, ylabel style={font=\scriptsize}, tick label style={font=\scriptsize}, y tick label style={font=\scriptsize, inner sep=1pt}, axis line style={lcSlate!50}, tick style={draw=none}}]
\addplot[matrix plot*, mesh/cols=6, point meta=explicit] table[x=t,y=k,meta=v] {
t k v
0 0 \diagCocoLlKzeroPzero
1 0 \diagCocoLlKzeroPone
2 0 \diagCocoLlKzeroPtwo
3 0 \diagCocoLlKzeroPthree
4 0 \diagCocoLlKzeroPfour
5 0 \diagCocoLlKzeroPfive
0 1 \diagCocoLlKonePzero
1 1 \diagCocoLlKonePone
2 1 \diagCocoLlKonePtwo
3 1 \diagCocoLlKonePthree
4 1 \diagCocoLlKonePfour
5 1 \diagCocoLlKonePfive
0 2 \diagCocoLlKtwoPzero
1 2 \diagCocoLlKtwoPone
2 2 \diagCocoLlKtwoPtwo
3 2 \diagCocoLlKtwoPthree
4 2 \diagCocoLlKtwoPfour
5 2 \diagCocoLlKtwoPfive
0 3 \diagCocoLlKthreePzero
1 3 \diagCocoLlKthreePone
2 3 \diagCocoLlKthreePtwo
3 3 \diagCocoLlKthreePthree
4 3 \diagCocoLlKthreePfour
5 3 \diagCocoLlKthreePfive
};
\addplot[matrix plot*, mesh/cols=6, point meta=explicit] table[x=t,y=k,meta=v] {
t k v
0 4.7 \brdCocoLlPzero
1 4.7 \brdCocoLlPone
2 4.7 \brdCocoLlPtwo
3 4.7 \brdCocoLlPthree
4 4.7 \brdCocoLlPfour
5 4.7 \brdCocoLlPfive
0 5.7 \brdCocoLlPzero
1 5.7 \brdCocoLlPone
2 5.7 \brdCocoLlPtwo
3 5.7 \brdCocoLlPthree
4 5.7 \brdCocoLlPfour
5 5.7 \brdCocoLlPfive
};
\draw[lcSlate!60, line width=0.4pt] (axis cs:-0.5,4.1) -- (axis cs:5.5,4.1);
\fill[pattern=north east lines, pattern color=lcSlate!70] (axis cs:-0.5,4.2) rectangle (axis cs:5.5,5.2);
\nextgroupplot[title={\textbf{(c) \textsc{Sim-CoT}, GPT-2}}]
\addplot[matrix plot*, mesh/cols=6, point meta=explicit] table[x=t,y=k,meta=v] {
t k v
0 0 \diagKzeroPzero
1 0 \diagKzeroPone
2 0 \diagKzeroPtwo
3 0 \diagKzeroPthree
4 0 \diagKzeroPfour
5 0 \diagKzeroPfive
0 1 \diagKonePzero
1 1 \diagKonePone
2 1 \diagKonePtwo
3 1 \diagKonePthree
4 1 \diagKonePfour
5 1 \diagKonePfive
0 2 \diagKtwoPzero
1 2 \diagKtwoPone
2 2 \diagKtwoPtwo
3 2 \diagKtwoPthree
4 2 \diagKtwoPfour
5 2 \diagKtwoPfive
0 3 \diagKthreePzero
1 3 \diagKthreePone
2 3 \diagKthreePtwo
3 3 \diagKthreePthree
4 3 \diagKthreePfour
5 3 \diagKthreePfive
};
\addplot[matrix plot*, mesh/cols=6, point meta=explicit] table[x=t,y=k,meta=v] {
t k v
0 4.7 \brdSimPzero
1 4.7 \brdSimPone
2 4.7 \brdSimPtwo
3 4.7 \brdSimPthree
4 4.7 \brdSimPfour
5 4.7 \brdSimPfive
0 5.7 \brdSimPzero
1 5.7 \brdSimPone
2 5.7 \brdSimPtwo
3 5.7 \brdSimPthree
4 5.7 \brdSimPfour
5 5.7 \brdSimPfive
};
\draw[lcSlate!60, line width=0.4pt] (axis cs:-0.5,4.1) -- (axis cs:5.5,4.1);
\nextgroupplot[title={\textbf{(d) \textsc{Sim-CoT}, \llamaBshort{}}}]
\addplot[matrix plot*, mesh/cols=6, point meta=explicit] table[x=t,y=k,meta=v] {
t k v
0 0 \diagSimLlKzeroPzero
1 0 \diagSimLlKzeroPone
2 0 \diagSimLlKzeroPtwo
3 0 \diagSimLlKzeroPthree
4 0 \diagSimLlKzeroPfour
5 0 \diagSimLlKzeroPfive
0 1 \diagSimLlKonePzero
1 1 \diagSimLlKonePone
2 1 \diagSimLlKonePtwo
3 1 \diagSimLlKonePthree
4 1 \diagSimLlKonePfour
5 1 \diagSimLlKonePfive
0 2 \diagSimLlKtwoPzero
1 2 \diagSimLlKtwoPone
2 2 \diagSimLlKtwoPtwo
3 2 \diagSimLlKtwoPthree
4 2 \diagSimLlKtwoPfour
5 2 \diagSimLlKtwoPfive
0 3 \diagSimLlKthreePzero
1 3 \diagSimLlKthreePone
2 3 \diagSimLlKthreePtwo
3 3 \diagSimLlKthreePthree
4 3 \diagSimLlKthreePfour
5 3 \diagSimLlKthreePfive
};
\addplot[matrix plot*, mesh/cols=6, point meta=explicit] table[x=t,y=k,meta=v] {
t k v
0 4.7 \brdSimLlPzero
1 4.7 \brdSimLlPone
2 4.7 \brdSimLlPtwo
3 4.7 \brdSimLlPthree
4 4.7 \brdSimLlPfour
5 4.7 \brdSimLlPfive
0 5.7 \brdSimLlPzero
1 5.7 \brdSimLlPone
2 5.7 \brdSimLlPtwo
3 5.7 \brdSimLlPthree
4 5.7 \brdSimLlPfour
5 5.7 \brdSimLlPfive
};
\draw[lcSlate!60, line width=0.4pt] (axis cs:-0.5,4.1) -- (axis cs:5.5,4.1);
\end{groupplot}
\end{tikzpicture}

%% file: sections/6_rq3_dynamics.tex
\begin{figure*}[b]
\centering
\input{figures/fig_rq3}
\caption{\textbf{Step size along the chain,} measured as
$1-\cos(\mathbf h_t,\mathbf h_{t-1})$, for every grid cell on
\textbf{(a)}~GPT-2 and \textbf{(b)}~\llamaB{}, on the layer-mean state
$\bar{\mathbf h}_t$ of Eq.~\ref{eq:states}. Shared colour scale; darker is a
larger step size.}
\label{fig:rqthree}
\end{figure*}
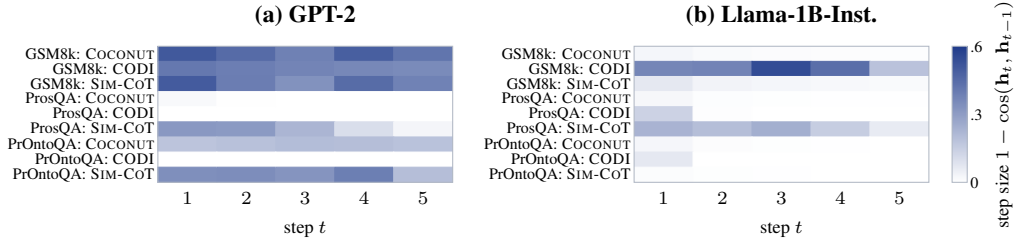

\section{RQ3: What Is Transmitted Across the Latent Slots, and Can a Settled State Be Revised?}
\label{sec:dynamics}
\label{sec:t2}
\label{sec:t4}
\label{sec:a4}

RQ2 showed what each slot contains. RQ3 asks whether that content is merely
carried forward or transformed across slots; only the second uses the
recurrence that gives a latent chain its capacity for serial computation.

\textbf{Setup.} We measure step size as the cosine distance between
consecutive states. To locate the last change in a discrete state sequence,
we follow the state-transition approach of \citet{yu2025stateaware}:
\begin{equation}
s_t=c\!\left(P\bar{\mathbf h}_t\right),\qquad
L=\max\{t:s_t\neq s_{t-1}\}.
\label{eq:states}
\end{equation}
Here $\bar{\mathbf h}_t$ is the mean hidden state across layers, $P$ is a
shared random projection, and $c$ is a $K$-means clustering fitted jointly
to the model's latent and written-step states. This gives all positions a
common set of state labels without fitting them to the answer. Chains with
no transition are counted separately. A cross-problem null replaces slots
with those of other problems at the same positions. A chain settles early
when its last change $L$ precedes its null $L^0$ by at least
\dynFloorSlots{} slots under all \dynSeeds{} clustering seeds. The
reading is unresolved when the null reaches the last slot
(App.~\ref{app:statemachine}). The along-answer ratio $r$ measures the
share of a step along a fitted answer direction. We call movement
answer-directed only when it exceeds a label-permuted refit null
(App.~\ref{app:reconciliation}). Step size alone is descriptive.

For mid-chain edits, we retain the first $i$ slots of the original problem
and use an edited twin with a different gold answer for the remaining
computation. The recycling-model driver splices in the twin's recorded
suffix, while the \codi{} driver recomputes later slots. We therefore use
this battery to measure the answer response, not to infer re-planning from
post-edit trajectories. The adaptive rate $A(i)$ is the fraction of answers
that follow the twin, restricted to problems solved on the twin from
scratch. Backbone comparisons use matched problems and are reported
separately (App.~\ref{app:edits}).

\textbf{Finding 7: On logical tasks, unsupervised and answer-aligned chains
settle early; per-step supervision sustains state movement to the last slot
and, on \prosqa{}, toward the answer.} Both the discrete
reading and the step size show early settling (App.~\ref{app:settling};
Figure~\ref{fig:rqthree}): settled chains halt state changes up to three
slots before their null and remain nearly stationary from the first step.
Conversely, per-step chains continue moving to the end. Directional
analysis reveals that this movement rarely turns toward the answer: in
\dynCellsNotAlignedRefit{} of \dynCellsUsablePCA{} cells with a usable
answer direction, each step's along-answer component stays at or below its
null. Per-step \prosqa{} is the one exception confirmed under both probe
arms; the other three are not confirmed under both arms
(App.~\ref{app:reconciliation}). Even there, the state's identity alone
never predicts the answer (Figure~\ref{fig:layout}). As RQ2 implies, a chain
whose answer is already readable at its first input has no reason to evolve,
whereas per-step supervision forces a new readable step at each slot. On
\gsm{} the \llamaB{} per-step twin settles early like the unsupervised
chains, so the split is a finding about the logical tasks
(App.~\ref{app:cosine}).

\begin{figure}[t]
\centering
\begin{minipage}[c]{0.47\textwidth}
\centering
\input{figures/fig_edit}
\end{minipage}\hfill
\begin{minipage}[c]{0.50\textwidth}
\caption{\textbf{Answers after a mid-chain edit.} Adaptive rate $A(i)$, the
fraction of answers that follow an edit made after $i$ slots, for GPT-2
(solid) and \llamaB{} (dashed) on \twinNenlarged{}-twin sets; every backbone
gap exceeds chance after Holm correction (Table~\ref{tab:backbone}). After
the edit only the prefix belongs to the original problem
(App.~\ref{app:edits}).}
\label{fig:edit}
\end{minipage}
\end{figure}
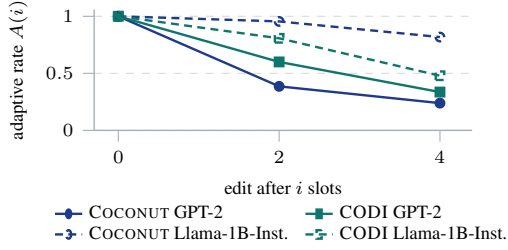

\textbf{Finding 8: Whether a mid-chain edit revises the answer is set by the
backbone, not by the mechanism or the supervision.}
Disrupting settled states mid-chain reveals a split by backbone. On GPT-2,
answers remain locked to the original prefix: the adaptive rate falls at
each later edit point in every consulted cell, directly measuring premature
commitment \citep{wu2025ctrls}. On \llamaB{}, the
answer follows the edit: $A(i)$ stays high under the same edits on matched
problems (Figure~\ref{fig:edit}). A held answer on GPT-2 reflects a readout
anchored in the prefix slots; on \llamaB{}, the same readout follows the
twin.

Together, these results show that the recurrence is largely unused for
serial computation. Answer aligned chains transmit a state fixed at the question; per-step supervision keeps the state moving, though only on \prosqa{} is that movement confirmed as answer-directed.

%% file: figures/fig_rq3.tex
\providecolor{lcNavy}{HTML}{1E3A8A}%
\providecolor{lcSlate}{HTML}{64748B}%
\begin{tikzpicture}
\pgfplotsset{
  hm/.style={
    scale only axis, width=3.9cm, height=1.8cm, line width=0.4pt,
    tick style={draw=none}, axis line style={lcSlate!50},
    label style={font=\scriptsize}, tick label style={font=\scriptsize},
    title style={font=\footnotesize, yshift=-2pt},
    xtick={1,...,5}, xmin=0.5, xmax=5.5, xlabel={step $t$},
    y dir=reverse, enlargelimits=false, axis on top,
    colormap={wp}{color(0cm)=(white); color(1cm)=(lcNavy)},
    point meta min=0, point meta max=0.6,
    y tick label style={font=\tiny, inner sep=1pt},
  },
}
\begin{groupplot}[group style={group size=2 by 1, horizontal sep=2.4cm}, hm]
\nextgroupplot[title={\textbf{(a) GPT-2}}, ymin=-0.5, ymax=8.5, ytick={0,...,8},
  yticklabels={\gsm{}: \coconut{}, \gsm{}: \codi{}, \gsm{}: \textsc{Sim-CoT}, \prosqa{}: \coconut{}, \prosqa{}: \codi{}, \prosqa{}: \textsc{Sim-CoT}, \prontoqa{}: \coconut{}, \prontoqa{}: \codi{}, \prontoqa{}: \textsc{Sim-CoT}}]
\addplot[matrix plot*, mesh/cols=5, point meta=explicit] table[x=t,y=r,meta expr={1-\thisrow{c}}] {
t r c
1 0 \cosSlotCocoGsmTone
2 0 \cosSlotCocoGsmTtwo
3 0 \cosSlotCocoGsmTthree
4 0 \cosSlotCocoGsmTfour
5 0 \cosSlotCocoGsmTfive
1 1 \cosSlotCodiGsmTone
2 1 \cosSlotCodiGsmTtwo
3 1 \cosSlotCodiGsmTthree
4 1 \cosSlotCodiGsmTfour
5 1 \cosSlotCodiGsmTfive
1 2 \cosSlotSimSixGsmTone
2 2 \cosSlotSimSixGsmTtwo
3 2 \cosSlotSimSixGsmTthree
4 2 \cosSlotSimSixGsmTfour
5 2 \cosSlotSimSixGsmTfive
1 3 \cosSlotCocoProsqaTone
2 3 \cosSlotCocoProsqaTtwo
3 3 \cosSlotCocoProsqaTthree
4 3 \cosSlotCocoProsqaTfour
5 3 \cosSlotCocoProsqaTfive
1 4 \cosSlotCodiProsqaTone
2 4 \cosSlotCodiProsqaTtwo
3 4 \cosSlotCodiProsqaTthree
4 4 \cosSlotCodiProsqaTfour
5 4 \cosSlotCodiProsqaTfive
1 5 \cosSlotSimProsqaTone
2 5 \cosSlotSimProsqaTtwo
3 5 \cosSlotSimProsqaTthree
4 5 \cosSlotSimProsqaTfour
5 5 \cosSlotSimProsqaTfive
1 6 \cosSlotCocoProntoTone
2 6 \cosSlotCocoProntoTtwo
3 6 \cosSlotCocoProntoTthree
4 6 \cosSlotCocoProntoTfour
5 6 \cosSlotCocoProntoTfive
1 7 \cosSlotCodiProntoTone
2 7 \cosSlotCodiProntoTtwo
3 7 \cosSlotCodiProntoTthree
4 7 \cosSlotCodiProntoTfour
5 7 \cosSlotCodiProntoTfive
1 8 \cosSlotSimProntoTone
2 8 \cosSlotSimProntoTtwo
3 8 \cosSlotSimProntoTthree
4 8 \cosSlotSimProntoTfour
5 8 \cosSlotSimProntoTfive
};
\nextgroupplot[title={\textbf{(b) \llamaBshort{}}}, ymin=-0.5, ymax=8.5, ytick={0,...,8},
  yticklabels={\gsm{}: \coconut{}, \gsm{}: \codi{}, \gsm{}: \textsc{Sim-CoT}, \prosqa{}: \coconut{}, \prosqa{}: \codi{}, \prosqa{}: \textsc{Sim-CoT}, \prontoqa{}: \coconut{}, \prontoqa{}: \codi{}, \prontoqa{}: \textsc{Sim-CoT}},
  colorbar, colorbar style={y dir=normal, width=0.2cm, height=1.8cm, ytick={0,0.3,0.6}, yticklabels={0,.3,.6}, ylabel={step size $1-\cos(\mathbf h_t,\mathbf h_{t-1})$}, ylabel style={font=\scriptsize}, tick label style={font=\scriptsize}, axis line style={lcSlate!50}, tick style={draw=none}}]
\addplot[matrix plot*, mesh/cols=5, point meta=explicit] table[x=t,y=r,meta expr={1-\thisrow{c}}] {
t r c
1 0 \cosSlotCocoGsmLlTone
2 0 \cosSlotCocoGsmLlTtwo
3 0 \cosSlotCocoGsmLlTthree
4 0 \cosSlotCocoGsmLlTfour
5 0 \cosSlotCocoGsmLlTfive
1 1 \cosSlotCodiGsmLlTone
2 1 \cosSlotCodiGsmLlTtwo
3 1 \cosSlotCodiGsmLlTthree
4 1 \cosSlotCodiGsmLlTfour
5 1 \cosSlotCodiGsmLlTfive
1 2 \cosSlotSimSixGsmLlTone
2 2 \cosSlotSimSixGsmLlTtwo
3 2 \cosSlotSimSixGsmLlTthree
4 2 \cosSlotSimSixGsmLlTfour
5 2 \cosSlotSimSixGsmLlTfive
1 3 \cosSlotCocoProsqaLlTone
2 3 \cosSlotCocoProsqaLlTtwo
3 3 \cosSlotCocoProsqaLlTthree
4 3 \cosSlotCocoProsqaLlTfour
5 3 \cosSlotCocoProsqaLlTfive
1 4 \cosSlotCodiProsqaLlTone
2 4 \cosSlotCodiProsqaLlTtwo
3 4 \cosSlotCodiProsqaLlTthree
4 4 \cosSlotCodiProsqaLlTfour
5 4 \cosSlotCodiProsqaLlTfive
1 5 \cosSlotSimProsqaLlTone
2 5 \cosSlotSimProsqaLlTtwo
3 5 \cosSlotSimProsqaLlTthree
4 5 \cosSlotSimProsqaLlTfour
5 5 \cosSlotSimProsqaLlTfive
1 6 \cosSlotCocoProntoLlTone
2 6 \cosSlotCocoProntoLlTtwo
3 6 \cosSlotCocoProntoLlTthree
4 6 \cosSlotCocoProntoLlTfour
5 6 \cosSlotCocoProntoLlTfive
1 7 \cosSlotCodiProntoLlTone
2 7 \cosSlotCodiProntoLlTtwo
3 7 \cosSlotCodiProntoLlTthree
4 7 \cosSlotCodiProntoLlTfour
5 7 \cosSlotCodiProntoLlTfive
1 8 \cosSlotSimProntoLlTone
2 8 \cosSlotSimProntoLlTtwo
3 8 \cosSlotSimProntoLlTthree
4 8 \cosSlotSimProntoLlTfour
5 8 \cosSlotSimProntoLlTfive
};
\end{groupplot}
\end{tikzpicture}

%% file: figures/fig_edit.tex
\providecolor{lcNavy}{HTML}{1E3A8A}%
\providecolor{lcSlate}{HTML}{64748B}%
\providecolor{lcTeal}{HTML}{0F766E}%
\begin{tikzpicture}
\begin{axis}[
    scale only axis, width=5.0cm, height=1.6cm, line width=0.5pt,
    axis x line*=bottom, axis y line*=left,
    tick align=outside, tick style={lcSlate!60, line width=0.4pt}, axis line style={lcSlate!60},
    label style={font=\scriptsize}, tick label style={font=\scriptsize},
    ymajorgrids, grid style={dashed, lcSlate!28, line width=0.3pt},
    every axis plot/.append style={mark size=1.6pt, line width=0.9pt},
    legend columns=2, legend style={draw=none, fill=none, font=\scriptsize, inner sep=1pt, row sep=-2.5pt, at={(0.5,-0.55)}, anchor=north, /tikz/column 2/.style={column sep=4pt}},
    legend cell align=left, legend image post style={xscale=0.6},
    xlabel={edit after $i$ slots}, ylabel={adaptive rate $A(i)$},
    ymin=0, ymax=1.06, xmin=-0.3, xmax=4.4, xtick={0,2,4},
]
\addplot[lcNavy, mark=*] coordinates {(0,1.0)(2,\pairCocoTwoGpt)(4,\pairCocoFourGpt)};
\addplot[lcTeal, mark=square*] coordinates {(0,1.0)(2,\pairCodiTwoGpt)(4,\pairCodiFourGpt)};
\addplot[lcNavy, densely dashed, mark=o] coordinates {(0,1.0)(2,\pairCocoTwoLl)(4,\pairCocoFourLl)};
\addplot[lcTeal, densely dashed, mark=square] coordinates {(0,1.0)(2,\pairCodiTwoLl)(4,\pairCodiFourLl)};
\legend{\coconut{} GPT-2, \codi{} GPT-2, \coconut{} \llamaBshort{}, \codi{} \llamaBshort{}}
\end{axis}
\end{tikzpicture}

%% file: sections/7_discussion.tex
\section{Discussion}
\label{sec:discussion}

\input{sections/4_taxonomy}

\textbf{Where the cells land.} The taxonomy (Figure~\ref{fig:tree}) places
cells in all six leaves. Every inert chain we profiled is either a carrier
or empty. Both placeholders occur on \llamaB{}/\prosqa{}, while
content-consulted chains are assembling on the logical tasks and
unstructured on \gsm{}. The full census and node thresholds appear in
App.~\ref{app:worked} (Table~\ref{tab:census}).

\textbf{Each leaf predicts an intervention not used to define it.} This
makes the taxonomy predictive rather than descriptive: carriers are
unaffected by corruption, placeholders remain immune even to a
different-answer donor at twice the natural norm, and the only chain helped
by donor thoughts is classified as assembling (App.~\ref{app:released}).

\textbf{Reading a latent trace certifies representation, not computation.}
Our results reconcile prior findings once we distinguish what a chain
represents from what the model uses. The placeholder signature reported by
\citet{zhang2025latenttokens} appears in two cells, whereas \coconut{} is a
carrier on our logical tasks. The collapsed superposition reported by
\citet{rizvimartel2026illusion} is consistent with what carriers hold, while
the staged causal structure and non-local routing of \citet{li2026dynamics}
match our consulted \gsm{} leaf.

\textbf{Implications for agents.} LLM agents chain many reasoning steps to
act, recommend, optimize content and manage memory
\citep{yu2026olivia,huang2025agenticrec,yu2026agent2ucb,zhou2026filesystem}.
If such agents adopt latent chains, a readable trace would not certify the
computation behind an action, which matters for auditing and securing them
\citep{yu2026fraud,wang2025cacheprune}.

\section{Conclusion}
\label{sec:conclusion}

We identify three separate properties of latent chains. First, the model
relies on a chain only when the task requires it or per-step supervision
installs that reliance. Second, the chains that encode the most are the ones
the model does not use: their answers are already readable before the chain
begins, and the slots only carry them. Third, most chains transmit a fixed
state from the third slot onward; on the logical tasks, only per-step supervision keeps the state changing, and on \prosqa{} that movement is directed toward the answer.
The three
properties are thus separate, and a claim about latent reasoning should say
which of them it concerns.

%% file: sections/4_taxonomy.tex
\subsection{A Taxonomy of Latent Chains}
\label{sec:taxonomy}

The three RQs describe different aspects of the same chains. We now combine
them to summarize the behaviors observed in our grid
(Figure~\ref{fig:tree}). Deletion and substitution determine the use
category. Content probes and state dynamics describe the profile within
that category. Readability does not override an intervention result, and
cells with unresolved measurements are left unclassified.

Inert chains are called \emph{carriers} when they carry a readable answer
and settle early, or \emph{empty} when little is readable under our probes.
Among chains with the placeholder signature, \emph{bare placeholders}
settle early with little content, while \emph{laid-out
placeholders} contain ordered steps and continue changing. Content-used
chains are \emph{assembling} when they show an ordered step layout
and sustained state changes, or \emph{unstructured} when no such layout is
recovered. These names summarize the measured profiles. In particular,
``assembling'' does not establish that slots causally correspond decoded step. The full decision rules, thresholds and
per-cell assignments are in App.~\ref{app:worked}.

%% file: sections/9_statements.tex

\subsection*{Use of large language models}
In this work, we used generative AI tools to implement analysis code, to
provide feedback on the experimental design and on the interpretation of
results, and to help draft the paper, its figures and its LaTeX. 

%% file: appendix/a_details.tex
\FloatBarrier
\section{Experimental details}
\label{app:protocol}

\subsection{Models and the grid}
\label{app:gridrationale}

\begin{table*}[htbp]
\centering\footnotesize
\setlength{\tabcolsep}{3pt}
\renewcommand{\arraystretch}{1.05}
\caption{\textbf{The model axis.} Input to each slot, what supervises the slots,
the slot budget and the grid cells per model. ``ours'': trained by us;
$\dagger$: trained at \coconut{}'s slot budget (the released \simcoco{}
checkpoint with \slotsSimReleased{} slots is the parity anchor,
App.~\ref{app:released}).}
\label{tab:models}
\resizebox{\ifdim\width>\textwidth\textwidth\else\width\fi}{!}{%
\begin{tabular}{@{}l p{3.4cm} p{4.1cm} c p{3.7cm}@{}}
\toprule
\textbf{Model} & \textbf{Input to slot $k{+}1$} & \textbf{What supervises the slots} & \textbf{Slots} & \textbf{Cells (GPT-2; \llamaBshort{} twins)} \\
\midrule
\coconut{} & last hidden state at slot $k$, recycled as the next input embedding & nothing beyond the answer loss & \slotsCoconut{} & \gsm{}, \prosqa{} ($\times2$), \prontoqa{}; all three twinned \\
\quad continued (control) & identical to \coconut{} & nothing; base continued for \simcoco{}'s schedule, auxiliary loss $=0$ & \slotsCoconut{} & \prosqa{} (ours; \llamaBshort{} twin ours) \\
\simcoco{} & identical to \coconut{} & per-step content: an auxiliary decoder reconstructs written step $k$ from slot $k$; discarded at inference & \slotsSimOurs{}\textsuperscript{$\dagger$} & \gsm{}, \prosqa{}, \prontoqa{} (all ours); \prosqa{} and \gsm{} twins (ours); \prontoqa{} twin at a swept checkpoint (ours) \\
\codi{} & recycled state through a projection network & answer-position alignment: self-distillation from the model's own written pathway & \slotsCodi{} & \gsm{}, \prosqa{}, \prontoqa{}; all three twinned \\
\bottomrule
\end{tabular}}
\end{table*}

\textbf{Parity.} No analysis runs on a checkpoint before it reproduces its
published accuracy. On \prosqa{} our frozen \Neval{}-problem prefixes score
above the published numbers; our scorer on the original full split reproduces
the published table to within $0.002$ (\parityCodiProsqaFull{} against
\parityCodiProsqaRef{}), so the gap comes from the split, not the scoring, and
no within-paper contrast is affected.

\subsection{Interventions, probes and edits}
\label{app:editprotocol}

\textbf{Interventions.} Each condition replaces the vector fed at a slot, and
downstream slots recompute from it. \emph{Noise} is an isotropic Gaussian draw
rescaled to the mean natural slot norm at that position; it matches magnitude
only and is near-orthogonal to the vector it replaces. \emph{Dataset mean} is
the mean slot vector over the evaluation set, and \emph{donor} is another
problem's recorded vector at the same slot. The mean and donor substitutes stay
nearly parallel to natural slot vectors (cosine $0.98$--$1.00$), so noise
probes the removal of content and mean and donor its replacement by wrong
content.

\textbf{Probes.} Probes are PCA to \probepca{} dimensions followed by logistic
regression, five-fold cross-validated over problems, against a label null
permuted \nperm{} times; on \prosqa{} a full-dimension ridge arm is added and a
verdict is stated alone only where both arms agree. Each cell reports its
minimum detectable effect, and cells that could not resolve an effect are
excluded from verdicts rather than counted as nulls.

\textbf{Mid-chain edits.} An edit is valid only if the pre-edit answer is
unreachable in the edited problem (\twinretention{} of \gsm{} twins survive,
\twinretentionprosqa{} on \prosqa{}). Analysis is restricted to problems the
model solves from scratch, and an answer-preserving twin controls for editing
itself: flip-minus-control displacement is $+\specFlipLo{}$ to
$+\specFlipHi{}$ against null 95th percentiles of
\specNullLo{}--\specNullHi{}. The backbone comparison was enlarged once, to
\twinNenlarged{} twins set by a power calculation under a protocol frozen
before scoring; gaps that had reached
\enlGapShareLo{}--\enlGapShareHi{} of their minimum detectable effect before
the enlargement cleared it after.

\subsection{The taxonomy}

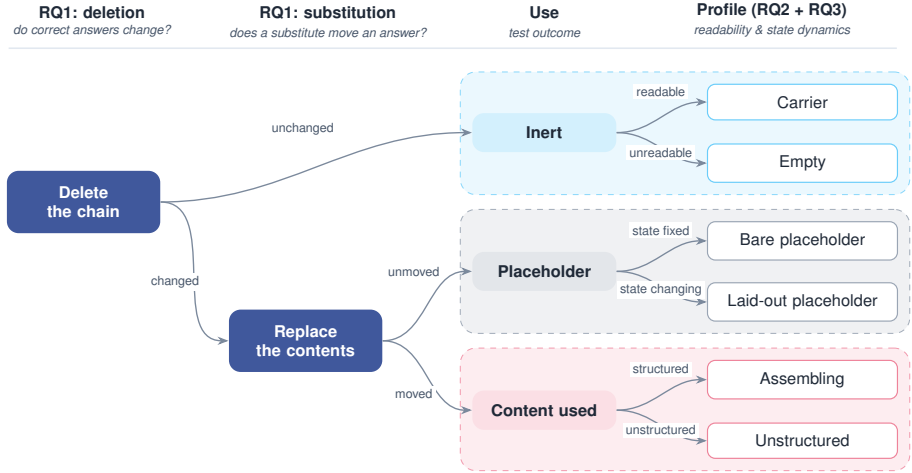
\begin{figure*}[t]
\centering
\resizebox{0.86\textwidth}{!}{\input{figures/fig_taxonomy}}
\caption{\textbf{A taxonomy based on the three RQs.} Deletion and content
replacement distinguish causal-use outcomes. Readability and state
movement describe the profile within each outcome. Unresolved cases are
not assigned to a leaf.}
\label{fig:tree}
\end{figure*}

%% file: figures/fig_taxonomy.tex
\providecolor{lcNavy}{HTML}{1E3A8A}%
\providecolor{lcCyan}{HTML}{38BDF8}%
\providecolor{lcSlate}{HTML}{64748B}%
\providecolor{lcPink}{HTML}{E11D48}%
\providecolor{lcText}{HTML}{1F2937}%
\providecolor{lcMuted}{HTML}{475569}%
\providecolor{lcRule}{HTML}{E2E8F0}%
\begin{tikzpicture}[
  x=1cm, y=1cm, font=\sffamily,
  decision/.style={draw=none, fill=lcNavy!85, text=white, font=\sffamily\bfseries\small,
                   rounded corners=6pt, inner sep=8pt, align=center, text width=2.2cm},
  macro/.style={draw=none, rounded corners=6pt, inner sep=7pt, align=center, text width=2.1cm,
                font=\sffamily\small\bfseries, text=lcText},
  macroI/.style={macro, fill=lcCyan!22},
  macroP/.style={macro, fill=lcSlate!18},
  macroC/.style={macro, fill=lcPink!14},
  leaf/.style={fill=white, line width=0.6pt, rounded corners=4pt, inner sep=6pt, align=center,
               text width=3.0cm, font=\sffamily\small, text=lcText},
  leafI/.style={leaf, draw=lcCyan!75},
  leafP/.style={leaf, draw=lcSlate!65},
  leafC/.style={leaf, draw=lcPink!55},
  groupbox/.style={dashed, line width=0.5pt, rounded corners=8pt, inner sep=6pt},
  groupI/.style={groupbox, draw=lcCyan!60, fill=lcCyan!10},
  groupP/.style={groupbox, draw=lcSlate!60, fill=lcSlate!10},
  groupC/.style={groupbox, draw=lcPink!45, fill=lcPink!8},
  edge/.style={-{Stealth[length=2mm]}, lcSlate!85, line width=0.7pt},
  el/.style={font=\sffamily\scriptsize, text=lcMuted, fill=white, inner sep=1.5pt, anchor=south},
  coltitle/.style={font=\sffamily\small\bfseries, text=lcText, anchor=south, inner sep=0pt},
  colsub/.style={font=\sffamily\scriptsize\itshape, text=lcMuted, anchor=south, inner sep=0pt},
  note/.style={font=\sffamily\footnotesize, text=lcMuted, anchor=west, inner sep=0pt},
]
\node[coltitle] at (1.5,7.75) {RQ1: deletion};
\node[colsub]   at (1.5,7.42) {do correct answers change?};
\node[coltitle] at (5.75,7.75) {RQ1: substitution};
\node[colsub]   at (5.75,7.42) {does a substitute move an answer?};
\node[coltitle] at (9.65,7.75) {Use};
\node[colsub]   at (9.65,7.42) {test outcome};
\node[coltitle] at (13.75,7.75) {Profile (RQ2 + RQ3)};
\node[colsub]   at (13.75,7.42) {readability \& state dynamics};
\draw[lcRule, line width=0.8pt] (0,7.22) -- (16,7.22);

\node[decision] (n1) at (1.35,4.45) {Delete\\ the chain};
\node[decision] (n2) at (5.35,1.95) {Replace\\ the contents};

\node[macroI] (inert) at (9.65,5.7) {Inert};
\node[macroP] (plh)   at (9.65,3.2) {Placeholder};
\node[macroC] (cont)  at (9.65,0.7) {Content used};

\node[leafI] (carrier)    at (14.3,6.25) {Carrier};
\node[leafI] (empty)      at (14.3,5.15) {Empty};
\node[leafP] (bare)       at (14.3,3.75) {Bare placeholder};
\node[leafP] (laid)       at (14.3,2.65) {Laid-out placeholder};
\node[leafC] (assembling) at (14.3,1.25) {Assembling};
\node[leafC] (unstruct)   at (14.3,0.15) {Unstructured};

\begin{scope}[on background layer]
  \node[groupI, fit=(inert) (carrier) (empty)] {};
  \node[groupP, fit=(plh) (bare) (laid)] {};
  \node[groupC, fit=(cont) (assembling) (unstruct)] {};
\end{scope}

\draw[edge] (n1.east) to[out=0, in=180] (inert.west);
\draw[edge] (n1.east) to[out=0, in=180] (n2.west);
\draw[edge] (n2.east) to[out=0, in=180] (plh.west);
\draw[edge] (n2.east) to[out=0, in=180] (cont.west);
\foreach \g/\l in {inert/carrier, inert/empty, plh/bare, plh/laid, cont/assembling, cont/unstruct} {
  \draw[edge] (\g.east) to[out=0, in=180] (\l.west);
}
\node[el] at (5.3,5.6)   {unchanged};
\node[el] at (3.0,2.85)  {changed};
\node[el] at (7.3,3.05) {unmoved};
\node[el] at (7.3,0.85) {moved};
\node[el] at (11.75,6.3) {readable};
\node[el] at (11.75,5.2) {unreadable};
\node[el] at (11.75,3.8) {state fixed};
\node[el] at (11.75,2.7) {state changing};
\node[el] at (11.75,1.3) {structured};
\node[el] at (11.75,0.2) {unstructured};

\end{tikzpicture}

%% file: appendix/b_rq1.tex
\FloatBarrier
\section{RQ1 supplement: deletion and substitution}
\label{app:rq1}
\label{app:percell}

GPT-2 and \llamaB{} are reported apart and never pooled.

\subsection{Replication and training-time controls}
\label{app:controls}

\begin{table}[htbp]
\centering\scriptsize\setlength{\tabcolsep}{2.2pt}
\caption{\textbf{The three control cells under deletion}, in the format of
Table~\ref{tab:deletion}. The second training is an independent
\coconut{}/\prosqa{} run on GPT-2; the loss-off controls continue the
checkpoint of each backbone for the per-step twin's schedule with the
auxiliary loss weighted to zero (Section~\ref{sec:models}).}
\label{tab:controls}
\begin{tabular}{@{}ll cccl@{}}
\toprule
\textbf{Control} & \textbf{Backbone} & $C_0$ & $C_1$ & helped/hurt & verdict \\
\midrule
2nd training & GPT-2 & \cocoprosqaCzero & \cocoprosqaCone & \cocoprosqaHelped/\cocoprosqaHurt & inert \\
loss off     & GPT-2 & \ctrlProsqaCzero & \ctrlProsqaCone & \ctrlProsqaHelped/\ctrlProsqaHurt & inert \\
loss off     & \llamaBshort{} & \ctrlLlCzero & \ctrlLlCone & \ctrlLlHelped/\ctrlLlHurt & unresolved \\
\bottomrule
\end{tabular}
\end{table}

The second training agrees with the primary cell on every condition within
\replmaxdelta{} (Section~\ref{sec:replication}). The GPT-2 loss-off control is
as inert as the cell it continues; the \llamaB{} control moves answers both
ways below its MDE (\ctrlLlMDE{}) and is unresolved.

\subsection{Truncation}
\label{app:truncation}

\begin{table*}[htbp]
\centering\small\setlength{\tabcolsep}{5pt}
\caption{\textbf{Accuracy by slots kept.} $k{=}0$ is full deletion ($C_1$),
\emph{all} is the intact chain ($C_0$); the interior conditions keep two
and four slots.}
\label{tab:truncation}
\begin{tabular}{@{}lccccl@{}}
\toprule
\textbf{Cell} & $k{=}0$ & $k{=}2$ & $k{=}4$ & \textbf{all} & \textbf{shape} \\
\midrule
\multicolumn{6}{@{}l}{\emph{GPT-2 --- the supervision grid}} \\
\coconut{}/\gsm{}      & \cocogsmCone      & \cocogsmKtwo      & \cocogsmKfour      & \cocogsmCzero      & monotone \\
\coconut{}/\prosqa{}   & \cocoprosqadwCone & \cocoprosqadwKtwo & \cocoprosqadwKfour & \cocoprosqadwCzero & flat \\
\coconut{}/\prontoqa{} & \cocoprontoCone   & \cocoprontoKtwo   & \cocoprontoKfour    & \cocoprontoCzero   & flat \\
\simcoco{}/\gsm{} & \simSixGsmCone & \simSixGsmKtwo & \simSixGsmKfour & \simSixGsmCzero & monotone \\
\simcoco{}/\prosqa{}   & \simProsqaCone    & \simProsqaKtwo    & \simProsqaKfour     & \simProsqaCzero    & monotone \\
\simcoco{}/\prontoqa{} & \simProntoCone    & \simProntoKtwo    & \simProntoKfour     & \simProntoCzero    & monotone \\
\codi{}/\gsm{}         & \codigsmCone      & \codigsmKtwo      & \codigsmKfour      & \codigsmCzero      & monotone \\
\codi{}/\prosqa{}      & \codiprosqaCone   & \codiprosqaKtwo   & \codiprosqaKfour    & \codiprosqaCzero   & flat \\
\codi{}/\prontoqa{}    & \codiprontoCone   & \codiprontoKtwo   & \codiprontoKfour    & \codiprontoCzero   & flat \\
\midrule
\multicolumn{6}{@{}l}{\emph{\llamaB{} --- the backbone twins}} \\
\coconut{}/\gsm{}      & \cocogsmLlCone    & \cocogsmLlKtwo    & \cocogsmLlKfour     & \cocogsmLlCzero    & saturates early \\
\codi{}/\gsm{}         & \codigsmLlCone    & \codigsmLlKtwo    & \codigsmLlKfour    & \codigsmLlCzero    & dips at $k{=}4$ \\
\simcoco{}/\gsm{}      & \simSixGsmLlCone  & \simSixGsmLlKtwo  & \simSixGsmLlKfour  & \simSixGsmLlCzero  & saturates early \\
\bottomrule
\end{tabular}
\end{table*}

Truncation keeps the first $k$ slots and inserts the end-of-thought marker,
so $k{=}0$ is deletion and $k{=}T$ the intact chain. On GPT-2 every inert cell
is flat at all depths and every consulted cell rises monotonically, so the
shape tracks reliance rather than architecture; the last two slots of every
consulted \gsm{} cell still carry \tailshareLo{}--\tailshareHi{} of the
deletion gap. On \llamaB{} the \gsm{} curves saturate early: the model has
finished, not that thinking hurts.

\subsection{Substitution: per-cell contrasts}
\label{app:substitution}

\begin{table*}[htbp]
\centering\small\setlength{\tabcolsep}{5pt}
\caption{\textbf{Substitution, per cell.} Change in accuracy against the intact
chain when every slot is replaced by noise, by the dataset-mean vector or by
another problem's recorded thoughts (donor), and the donor-versus-deletion
contrast. Discordant pairs read donor-worse\,/\,deletion-worse; $p$ is
McNemar. GPT-2 and \llamaB{} are never pooled.}
\label{tab:corruption}
\resizebox{\ifdim\width>\textwidth\textwidth\else\width\fi}{!}{%
\begin{tabular}{@{}lcccl@{}}
\toprule
\textbf{Cell} & \textbf{noise} & \textbf{mean} & \textbf{donor} & \textbf{donor vs.\ deletion} \\
\midrule
\multicolumn{5}{@{}l}{\emph{GPT-2 --- the supervision grid}} \\
\coconut{}/\gsm{}  & \cocogsmDnoise  & \cocogsmDmean  & \cocogsmDdonor  & donor worse \cocogsmPoisonPair{}, $p{=}\cocogsmPoisonP$ \\
\simcoco{}/\gsm{} & \simSixGsmDnoise & \simSixGsmDmean & \simSixGsmDdonor & donor worse \simSixGsmPoisonPair{}, $\simSixGsmPoisonGap{>}\simSixGsmPoisonMDE$ \\
\simcoco{}/\prontoqa{} & \simProntoDnoise & \simProntoDmean & \simProntoDdonor & donor worse \simProntoPoisonPair{}, $\simProntoPoisonGap{>}\simProntoPoisonMDE$ \\
\codi{}/\gsm{}     & \codigsmDnoise  & \codigsmDmean  & \codigsmDdonor  & donor worse \codigsmPoisonPair{}, $p{=}\codigsmPoisonP$ \\
\simcoco{}/\prosqa{} & \simProsqaDnoise & \simProsqaDmean & \simProsqaDdonor & \textbf{donor above deletion}: \simProsqaInvPair{}, $\simProsqaInvGap{>}\simProsqaInvMDE$ \\
\codi{}/\prosqa{}  & \codiprosqaDnoise & \codiprosqaDmean & \codiprosqaDdonor & bounded (\codiprosqaBoundPair{}; needs $N{>}\codiprosqaBoundN$) \\
\coconut{}/\prosqa{} & \cocoprosqadwDnoise & \cocoprosqadwDmean & \cocoprosqadwDdonor & not applicable (inert under deletion) \\
\coconut{}/\prontoqa{} & \cocoprontoDnoise & \cocoprontoDmean & \cocoprontoDdonor & not applicable (inert under deletion) \\
\codi{}/\prontoqa{} & \codiprontoDnoise & \codiprontoDmean & \codiprontoDdonor & not applicable (inert under deletion) \\
\midrule
\multicolumn{5}{@{}l}{\emph{\llamaB{} --- the backbone twins}} \\
\codi{}/\gsm{}     & \codigsmLlDnoise & \codigsmLlDmean & \codigsmLlDdonor & donor worse \codigsmLlPoisonPair{}, $p{=}\codigsmLlPoisonP$ \\
\coconut{}/\gsm{}  & \cocogsmLlDnoise & \cocogsmLlDmean & \cocogsmLlDdonor & bounded (needs $N{\approx}\cocogsmLlBoundN$; pool \cocogsmLlPoolN) \\
\simcoco{}/\gsm{} & \simSixGsmLlDnoise & \simSixGsmLlDmean & \simSixGsmLlDdonor & bounded (\simSixGsmLlPoisonPair{}, $p{=}\simSixGsmLlPoisonP$; needs $N{\approx}\simSixGsmLlBoundN$; pool \simSixGsmLlPoolN) \\
\codi{}/\prosqa{}  & \codiprosqaLlDnoise & \codiprosqaLlDmean & \codiprosqaLlDdonor & donor above deletion: \codiprosqaLlInvPair{}, $p{=}\codiprosqaLlInvP$ \\
\coconut{}/\prosqa{} & \cocoprosqaLlDnoise & \cocoprosqaLlDmean & \cocoprosqaLlDdonor & not applicable (inert under deletion) \\
\coconut{}/\prontoqa{} & \cocoprontoLlDnoise & \cocoprontoLlDmean & \cocoprontoLlDdonor & not applicable (inert under deletion) \\
\codi{}/\prontoqa{} & \codiprontoLlDnoise & \codiprontoLlDmean & \codiprontoLlDdonor & not applicable (inert under deletion) \\
\simcoco{}/\prosqa{} & \simProsqaLlDnoise & \simProsqaLlDmean & \simProsqaLlDdonor & donor above deletion: no substitute moves an answer (placeholder) \\
\simcoco{}/\prontoqa{} & \simProntoLlDnoise & \simProntoLlDmean & \simProntoLlDdonor & not applicable (inert under deletion) \\
\bottomrule
\end{tabular}}
\end{table*}

Across the eighteen cells the tally is \poisonCount{} donor worse than
deletion, \naCount{} not applicable, \boundedCount{} bounded and
\invertedCount{} with the donor above deletion. In every content-used \gsm{}
cell and on \prontoqa{}, another problem's thoughts cost more than none, an
ordering a format account of ablation damage cannot produce; on per-step
\prosqa{}, by contrast, a donor's thoughts beat having none.

\subsection{The direction-by-scale control}
\label{app:twobytwo}

\begin{table}[htbp]
\centering\small\setlength{\tabcolsep}{3.5pt}
\caption{\textbf{Direction-by-scale $2{\times}2$} on the pre-registered cell list,
noise seed reused. Columns: random direction at the noise norm $\varepsilon$
and at the norm of the dataset-mean vector; mean direction at its own norm
and at $\varepsilon$.}
\label{tab:twobytwo}
\resizebox{\ifdim\width>\textwidth\textwidth\else\width\fi}{!}{%
\begin{tabular}{@{}lccccl@{}}
\toprule
\textbf{Cell} & rand@$\varepsilon$ & rand@$\lVert\mu\rVert$ & mean@$\lVert\mu\rVert$ & mean@$\varepsilon$ & \textbf{reading} \\
\midrule
\multicolumn{6}{@{}l}{\emph{GPT-2 --- the supervision grid}} \\
\codi{}/\prosqa{} & \codiprosqaDnoise & \codiprosqaRandNorm & \codiprosqaDmean & \codiprosqaMeanEps & direction \\
\codi{}/\gsm{}    & \codigsmDnoise    & \codigsmRandNorm    & \codigsmDmean    & \codigsmMeanEps    & direction \\
\coconut{}/\gsm{} & \cocogsmDnoise    & \cocogsmRandNorm    & \cocogsmDmean    & \cocogsmMeanEps    & mixed (all four within \cocogsmTwoByTwoSpread{}) \\
\simcoco{}/\prontoqa{} & \simProntoDnoise & \simProntoRandNorm & \simProntoDmean & \simProntoMeanEps & direction \\
\midrule
\multicolumn{6}{@{}l}{\emph{\llamaB{} --- the backbone twins}} \\
\codi{}/\gsm{} & \codigsmLlDnoise & \codigsmLlRandNorm & \codigsmLlDmean & \codigsmLlMeanEps & direction \\
\simcoco{}/\gsm{} & \simSixGsmLlDnoise & \simSixGsmLlRandNorm & \simSixGsmLlDmean & \simSixGsmLlMeanEps & mixed (all four within \simSixGsmLlTwoByTwoSpread{}) \\
\bottomrule
\end{tabular}}
\end{table}

\codi{}/\prosqa{} does not read its chain for content (a donor changes zero
answers), yet needs whatever occupies the slots to lie on its training
manifold; its \llamaB{} twin drops even that, a clean placeholder signature.
On \simcoco{}/\prontoqa{} the dataset-mean thought is far more damaging than a
random one at both norms. On the per-step \llamaB{}/\gsm{} twin the four
deltas lie within \simSixGsmLlTwoByTwoSpread{} of each other, so direction and
scale cannot be separated there.

%% file: appendix/c_rq2.tex
\FloatBarrier
\section{RQ2 supplement: probing}
\label{app:rq2}

\subsection{The answer through depth and by slot}
\label{app:depth}

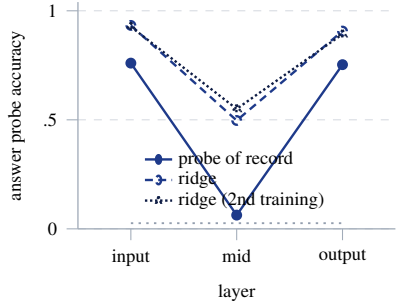
\begin{figure}[htbp]
\centering
\input{figures/fig_depth}
\caption{\textbf{The unsupervised chain's answer through the stack} at the first
slot (\coconut{}/\prosqa{}, GPT-2), under both probe arms and on the second
training.}
\label{fig:depth}
\end{figure}

The answer is present at the slot's input layer, the question's final hidden
state, weaker mid-stack and re-linearised at the output, under both probe arms
and on the second training (Figure~\ref{fig:depth}). Under the probe of record
it reads \portraitUemb{} at the embedding layer, \portraitUlo{} at the
mid-stack trough and \portraitUtop{} at the output (\portraitUembR{},
\portraitUloR{} and \portraitUtopR{} under ridge). By slot, the unsupervised
chain runs from \fitCocoPzero{} to \fitCocoPfive{}, while the supervised chain
is unreadable at its first two slots and reaches \fitSimPfive{} at the last.

\subsection{Position-by-layer sweep}
\label{app:sweep}

\begin{table}[htbp]
\centering\small\setlength{\tabcolsep}{3pt}
\caption{\textbf{Position $\times$ layer sweep} under both probe arms.
\emph{cleared}: cells above their own permutation null; $\geq$5$\times$:
cells at five times chance; \emph{peak}: the strongest cell and its location.
Numeric rows use chance \chanceMag{} and MDE \mdeMag{}; concept rows
\chanceProsqaConcept{} and \mdeProsqaConcept{}. The ridge arm exists on
\prosqa{} only. The last layer column (\probeLayerGpt{} on GPT-2,
\probeLayerLl{} on \llamaBshort{}) is the representation Figure~\ref{fig:layout}
reports.}
\label{tab:depthsweep}
\resizebox{\ifdim\width>\textwidth\textwidth\else\width\fi}{!}{%
\begin{tabular}{@{}llcccl@{}}
\toprule
\textbf{Cell} & \textbf{Task / arm} & \textbf{cleared} & $\geq$5$\times$ & \textbf{peak} & \textbf{at} \\
\midrule
\multicolumn{6}{@{}l}{\emph{GPT-2}} \\
\coconut{}    & \prosqa{}, record & \cocoprosqadwCleared & \cocoprosqadwFiveX & \textbf{\cocoprosqadwPeak} & \cocoprosqadwPeakCell \\
              & \prosqa{}, ridge  & \cocoprosqadwClearedR & \cocoprosqadwFiveXR & \textbf{\cocoprosqadwPeakR} & \cocoprosqadwPeakCellR \\
\coconut{}$'$ & \prosqa{}, record & \cocoprosqaCleared   & \cocoprosqaFiveX & \textbf{\cocoprosqaPeak}   & \cocoprosqaPeakCell \\
              & \prosqa{}, ridge  & \cocoprosqaClearedR  & \cocoprosqaFiveXR & \textbf{\cocoprosqaPeakR}  & \cocoprosqaPeakCellR \\
\simcoco{}    & \prosqa{}, record & \simProsqaCleared    & \simProsqaFiveX & \textbf{\simProsqaPeak}    & \simProsqaPeakCell \\
              & \prosqa{}, ridge  & \simProsqaClearedR   & \simProsqaFiveXR & \textbf{\simProsqaPeakR}   & \simProsqaPeakCellR \\
\codi{}       & \prosqa{}, record & \codiprosqaCleared   & \codiprosqaFiveX & \codiprosqaPeak            & \codiprosqaPeakCell \\
              & \prosqa{}, ridge  & \codiprosqaClearedR  & \codiprosqaFiveXR & \codiprosqaPeakR          & \codiprosqaPeakCellR \\
\codi{}       & \gsm{}    & \codigsmCleared      & --- & \codigsmPeak               & \codigsmPeakCell \\
\simcoco{} (\slotsSimReleased{} sl.) & \gsm{} & \simgsmCleared & --- & \simgsmPeak & \simgsmPeakCell \\
\midrule
\multicolumn{6}{@{}l}{\emph{\llamaB{}}} \\
\coconut{} & \prosqa{}, record & \cocoprosqaLlCleared & \cocoprosqaLlFiveX & \cocoprosqaLlPeak & \cocoprosqaLlPeakCell \\
           & \prosqa{}, ridge  & \cocoprosqaLlClearedR & \cocoprosqaLlFiveXR & \textbf{\cocoprosqaLlPeakR} & \cocoprosqaLlPeakCellR \\
\codi{}    & \prosqa{}, record & \codiprosqaLlCleared & \codiprosqaLlFiveX & \codiprosqaLlPeak & \codiprosqaLlPeakCell \\
           & \prosqa{}, ridge  & \codiprosqaLlClearedR & \codiprosqaLlFiveXR & \codiprosqaLlPeakR & \codiprosqaLlPeakCellR \\
\bottomrule
\end{tabular}}
\end{table}

\subsection{Accumulation on \gsm{}}
\label{app:accum}

\begin{table}[htbp]
\centering\small\setlength{\tabcolsep}{4pt}
\caption{\textbf{Content accumulation across slots on \gsm{},} target magnitude of
the final answer (chance \chanceMag{}, MDE \mdeMag{}): first-slot and
last-slot accuracy and their difference.}
\label{tab:accum}
\resizebox{\ifdim\width>\textwidth\textwidth\else\width\fi}{!}{%
\begin{tabular}{@{}llcccl@{}}
\toprule
\textbf{Cell} & \textbf{Mechanism} & \textbf{first} & \textbf{last} & $\Delta$ & \textbf{Profile} \\
\midrule
\multicolumn{6}{@{}l}{\emph{GPT-2}} \\
\coconut{} & recycling & \cocogsmAccZero & \cocogsmAccLast & \textbf{\cocogsmAccDelta} & builds \\
\simcoco{} (\slotsSimOurs{} sl.) & recycling $+$ per-step & \simSixGsmAccZero & \simSixGsmAccLast & \textbf{\simSixGsmAccDelta} & builds \\
\simcoco{} (\slotsSimReleased{} sl.) & recycling $+$ per-step & \simgsmAccZero & \simgsmAccLast & \textbf{\simgsmAccDelta} & builds \\
\codi{} & distillation & \codigsmAccZero & \codigsmAccLast & \codigsmAccDelta & flat ($p{=}\codigsmAccP$) \\
\midrule
\multicolumn{6}{@{}l}{\emph{\llamaB{}}} \\
\coconut{} & recycling & \cocogsmLlAccZero & \cocogsmLlAccLast & \cocogsmLlAccDelta & flat ($p{=}\cocogsmLlAccP$) \\
\codi{} & distillation & \codigsmLlAccZero & \codigsmLlAccLast & \codigsmLlAccDelta & flat ($p{=}\codigsmLlAccP$) \\
\simcoco{} (\slotsSimOurs{} sl.) & recycling $+$ per-step & \accumFirstSimSixGsmLl & \accumLastSimSixGsmLl & \accumDeltaSimSixGsmLl & builds at the MDD (\accumMddSimSixGsmLl; App.~\ref{app:llamaperstep}) \\
\bottomrule
\end{tabular}}
\end{table}

Accumulation appears in every GPT-2 recycling cell (first-to-last gains
\accDeltaLo{} to \accDeltaHi{}, $p<\accBuildP{}$) and, at the readout's own
resolution, on the \llamaB{} per-step twin, and in no other cell.
Cross-backbone amounts are not compared.

\subsection{Layout and held-out cells}
\label{app:layout}

\begin{table*}[htbp]
\centering\small\setlength{\tabcolsep}{4pt}
\caption{\textbf{Layout on \prosqa{}: cells held out of the body and the chain
funnel} under both probe arms (record\,/\,ridge). \emph{elig.}/\emph{both}:
eligible slots and the pooled both-readable rate at the adjudicating slot;
\emph{present}: path cells present of 24; \emph{local.}: localisation;
\emph{off-path}: live off-path cells. \coconut{}$'$ is the independently
trained replication checkpoint.}
\label{tab:staticheldout}
\resizebox{\ifdim\width>\textwidth\textwidth\else\width\fi}{!}{%
\begin{tabular}{@{}llcccccl@{}}
\toprule
\textbf{Cell} & \textbf{Superv.} & \textbf{elig.} & \textbf{both} & \textbf{present} & \textbf{local.} & \textbf{off-path} & \textbf{Chain verdict} \\
\midrule
\multicolumn{8}{@{}l}{\emph{GPT-2}} \\
\coconut{} (primary) & none & 6/6 & \cocoprosqadwBoth\,/\,\cocoprosqadwBothR & \presentCocoB\,/\,\presentCocoBR & --- & \offpathClean\,/\,\offpathClean & present, not localised (both arms) \\
\coconut{}$'$ (replication) & none & 6/6 & \cocoprosqaBoth\,/\,\cocoprosqaBothR & \presentCoco\,/\,\presentCocoR & --- & \offpathClean\,/\,\offpathClean & present, not localised (both arms) \\
\simcoco{} & per-step & \simProsqaElig & \simProsqaBoth\,/\,\simProsqaBothR & \presentSim\,/\,\presentSimR & \checkmark & \offpathClean\,/\,\offpathClean & localised and ordered (both arms) \\
\codi{} & aligned & 0/6\,/\,0/6 & n/a & \presentCodi\,/\,\presentCodiR & --- & \offpathCodi\,/\,\offpathCodiR & not localised; specificity probe-dependent \\
\midrule
\multicolumn{8}{@{}l}{\emph{\llamaB{}}} \\
\coconut{} & none & 6/6 & \cocoprosqaLlBoth\,/\,\cocoprosqaLlBothR & \presentCocoLl\,/\,\presentCocoLlR & --- & \offpathClean\,/\,\offpathClean & present, not localised (both arms) \\
\codi{} & aligned & 0/6\,/\,\codiprosqaLlEligR & n/a\,/\,\codiprosqaLlBothR & \presentCodiLl\,/\,\presentCodiLlR & --- & \offpathCodiLl\,/\,\offpathCodiLlR & not localised; specificity probe-dependent \\
\bottomrule
\end{tabular}}
\end{table*}

The replication checkpoint agrees with the primary in both arms. The two
answer-aligned cells fail localisation under both arms (on-path accuracy
\codiOnpathLo{}--\codiOnpathHi{} against the supervised cell's
\simOnpathLo{}--\simOnpathHi{}), and their degree-matched specificity control
flips with the probe arm, so specificity there is unresolved. Both arms are
linear, so ``holds little'' is a statement about linear accessibility.

\subsection{One candidate at a time}
\label{app:candidates}

\begin{table}[htbp]
\centering\small\setlength{\tabcolsep}{4pt}
\caption{\textbf{One candidate at a time.} Rate at which both offered \prosqa{}
answers are readable from the same slot, at each cell's most generous
eligible slot, under both probe arms, against a bar of
\superposThreshCount{} problems of \Neval{}; ceilings are the readout's
measured sensitivity per arm.}
\label{tab:candidates}
\resizebox{\ifdim\width>\textwidth\textwidth\else\width\fi}{!}{%
\begin{tabular}{@{}llccccl@{}}
\toprule
& & & & \multicolumn{2}{c}{\textbf{both}} & \\
\cmidrule(lr){5-6}
\textbf{Cell} & \textbf{Superv.} & \textbf{elig.} & \textbf{slot} & \textbf{record} & \textbf{ridge} & \textbf{Verdict} \\
\midrule
\multicolumn{7}{@{}l}{\emph{GPT-2} (ceiling \ceilGpt{} record, \ceilGptR{} ridge)} \\
\coconut{} & none & 6/6 & \cocoprosqadwBothSlot & \textbf{\cocoprosqadwBoth} & \textbf{\cocoprosqadwBothR} & sub-bar, both arms \\
\simcoco{} & per-step & \simProsqaElig & \simProsqaBothSlot & \textbf{\simProsqaBoth} & \textbf{\simProsqaBothR} & sub-bar, both arms \\
\midrule
\multicolumn{7}{@{}l}{\emph{\llamaB{}} (ceiling \ceilLlama{} record, \ceilLlamaR{} ridge)} \\
\coconut{} & none & 6/6 & \cocoprosqaLlBothSlot & \textbf{\cocoprosqaLlBoth} & \textbf{\cocoprosqaLlBothR} & sub-bar, both arms \\
\codi{} & aligned & 0/6, \codiprosqaLlEligR & \codiprosqaLlBothSlotR & n/a & \textbf{\codiprosqaLlBothR} & sub-bar, ridge only \\
\bottomrule
\end{tabular}}
\end{table}

Each \prosqa{} candidate is scored with its own binary probe, since a
multinomial softmax couples the two scores and manufactures anti-correlation
(\probeauditcorr{} on synthetic data). Problems on which both read from the
same slot are counted against a bar fixed before measurement at
\superposthresh{} (co-occurrence null \superposNull{}, MDE \superposMDE{}); no
cell reaches it in either arm. Where the question cannot be asked (\codi{} on
GPT-2 has no readable answer, \prontoqa{} is yes/no, \gsm{} has no candidate
list) the cell is not applicable rather than negative.

%% file: figures/fig_depth.tex
\providecolor{lcNavy}{HTML}{1E3A8A}%
\providecolor{lcSlate}{HTML}{64748B}%
\begin{tikzpicture}
\begin{axis}[
  scale only axis, width=4.2cm, height=3cm, line width=0.5pt,
  axis x line*=bottom, axis y line*=left,
  tick align=outside, tick style={lcSlate!60, line width=0.4pt}, axis line style={lcSlate!60},
  label style={font=\scriptsize}, tick label style={font=\scriptsize},
  xlabel={layer}, ylabel={answer probe accuracy}, ytick={0,0.5,1}, yticklabels={0,.5,1},
  symbolic x coords={input,mid,output}, xtick=data, ymin=0, ymax=1.04, enlarge x limits=0.25,
  ymajorgrids, grid style={dashed, lcSlate!28, line width=0.3pt},
  legend style={at={(0.5,0.05)}, anchor=south, draw=none, fill=none, font=\scriptsize, inner sep=1pt, row sep=-2.5pt},
  legend image post style={xscale=0.6}, legend cell align=left,
  every axis plot/.append style={mark size=1.6pt, line width=0.9pt},
]
\addplot[lcNavy, mark=*] coordinates {(input,\portraitUemb)(mid,\portraitUlo)(output,\portraitUtop)};
\addplot[lcNavy, densely dashed, mark=o] coordinates {(input,\portraitUembR)(mid,\portraitUloR)(output,\portraitUtopR)};
\addplot[lcNavy!50!black, densely dotted, mark=triangle] coordinates {(input,\portraitBUembR)(mid,\portraitBUloR)(output,\portraitBUtopR)};
\addplot[lcSlate!70, dotted, line width=0.7pt, forget plot] coordinates {(input,\chanceProsqaConcept)(output,\chanceProsqaConcept)};
\legend{probe of record, ridge, ridge (2nd training)}
\end{axis}
\end{tikzpicture}

%% file: appendix/d_rq3.tex
\FloatBarrier
\section{RQ3 supplement: dynamics}
\label{app:rq3}
\label{app:dynamics}

\subsection{State-machine settings}
\label{app:statemachine}

Settling needs an event rather than a threshold, so we read the last slot at
which a discrete state label changes, following the state-transition view of
written reasoning \citep{yu2025stateaware}. The projection $P$ of
Eq.~\ref{eq:states} is a shared random Gaussian map to \dynProjDim{}
dimensions rather than a fitted one, so state labels cannot be tuned toward the
answer, and it is shared across slots and across latent and written states, so
a label means the same thing at every position. The clustering $c$ is one
$K$-means fitted jointly over the model's latent and written-step states. Every
$K\in\{\dynKset\}$ is reported and none is privileged, and a verdict is stated
only where all \dynSeeds{} clustering seeds agree (\dynSeedStable{} of
\dynSeedTotal{} verdicts are unanimous). Nulls replace slot $t$ of one problem
with slot $t$ of another (\dynNperm{} draws); the settling floor is
\dynFloorSlots{} slots, and the reading is unresolved when the null itself sits
at the last slot. A mid-chain edit keeps $\mathbf{h}_{0:i-1}$ and continues on
a twin whose gold answer differs by one changed quantity; an output that keeps
the original answer is \emph{rigid}.

\subsection{Settling}
\label{app:settling}

\begin{table}[htbp]
\centering\scriptsize
\setlength{\tabcolsep}{3pt}
\caption{\textbf{When the chain stops changing.} Slot of the last state transition
at $K{=}5$ (mean over chains with at least one transition), the cross-problem
null and their difference in slots; single-state chains counted apart.
\emph{ceiling}: the null itself sits at the last slot. Verdicts obey the
\dynFloorSlots{}-slot floor.}
\label{tab:settling}
\begin{tabular}{@{}l r r r r l@{}}
\toprule
\textbf{Cell} & \textbf{1-state} & \textbf{Last} & \textbf{Null} & $\Delta$ & \textbf{Verdict} \\
\midrule
\multicolumn{6}{@{}l}{\emph{\prosqa{}}} \\
\coconut{} & \tTwoCocoProsqaSingle & \tTwoCocoProsqaLast & \tTwoCocoProsqaNull & \tTwoCocoProsqaDelta & early \\
\coconut{} (2nd) & \tTwoCocoProsqaBSingle & \tTwoCocoProsqaBLast & \tTwoCocoProsqaBNull & \tTwoCocoProsqaBDelta & early \\
\coconut{} (Llama) & \tTwoCocoProsqaLlSingle & \tTwoCocoProsqaLlLast & \tTwoCocoProsqaLlNull & \tTwoCocoProsqaLlDelta & early \\
\codi{} & \tTwoCodiProsqaSingle & \tTwoCodiProsqaLast & \tTwoCodiProsqaNull & \tTwoCodiProsqaDelta & early \\
\codi{} (Llama) & \tTwoCodiProsqaLlSingle & \tTwoCodiProsqaLlLast & \tTwoCodiProsqaLlNull & \tTwoCodiProsqaLlDelta & early \\
\simcot{} & \tTwoSimProsqaSingle & \tTwoSimProsqaLast & \tTwoSimProsqaNull & \tTwoSimProsqaDelta & \textbf{keeps changing} \\
\multicolumn{6}{@{}l}{\emph{\prontoqa{}}} \\
\coconut{} & \tTwoCocoProntoSingle & \tTwoCocoProntoLast & \tTwoCocoProntoNull & \tTwoCocoProntoDelta & early \\
\coconut{} (Llama) & \tTwoCocoProntoLlSingle & \tTwoCocoProntoLlLast & \tTwoCocoProntoLlNull & \tTwoCocoProntoLlDelta & early \\
\codi{} & \tTwoCodiProntoSingle & \tTwoCodiProntoLast & \tTwoCodiProntoNull & \tTwoCodiProntoDelta & early \\
\codi{} (Llama) & \tTwoCodiProntoLlSingle & \tTwoCodiProntoLlLast & \tTwoCodiProntoLlNull & \tTwoCodiProntoLlDelta & unres. \\
\simcot{} & \tTwoSimProntoSingle & \tTwoSimProntoLast & \tTwoSimProntoNull & \tTwoSimProntoDelta & \textbf{keeps changing} \\
\multicolumn{6}{@{}l}{\emph{\gsm{}}} \\
\coconut{} & \tTwoCocoGsmSingle & \tTwoCocoGsmLast & \tTwoCocoGsmNull & \tTwoCocoGsmDelta & unres., ceiling \\
\simcot{} & \tTwoSimSixGsmSingle & \tTwoSimSixGsmLast & \tTwoSimSixGsmNull & \tTwoSimSixGsmDelta & unres., ceiling \\
\codi{} & \tTwoCodiGsmSingle & \tTwoCodiGsmLast & \tTwoCodiGsmNull & \tTwoCodiGsmDelta & unres., ceiling \\
\coconut{} (Llama) & \tTwoCocoGsmLlSingle & \tTwoCocoGsmLlLast & \tTwoCocoGsmLlNull & \tTwoCocoGsmLlDelta & early \\
\codi{} (Llama) & \tTwoCodiGsmLlSingle & \tTwoCodiGsmLlLast & \tTwoCodiGsmLlNull & \tTwoCodiGsmLlDelta & at null, ceiling \\
\simcot{} (Llama) & \dash{} & \settleStepSimSixGsmLl & \settleNullSimSixGsmLl & \settleDeltaSimSixGsmLl & early \\
\bottomrule
\end{tabular}
\end{table}

\subsection{Consecutive-slot similarity}
\label{app:cosine}

$\cos(\mathbf h_t,\mathbf h_{t-1})$ is computed per problem against a null
that pairs slot $t$ of one problem with slot $t{-}1$ of another. On \prosqa{}
it is \cosSlotCocoProsqaTone{} for \coconut{} and \cosSlotCodiProsqaTone{} for
\codi{} at the first step and stays there, whereas the per-step chain climbs
from \cosSlotSimProsqaTone{} to \cosSlotSimProsqaTfive{} on GPT-2 and from
\cosSlotSimProsqaLlTone{} to \cosSlotSimProsqaLlTfive{} on \llamaB{};
\prontoqa{} shows the same split. On \gsm{} every GPT-2 recipe keeps moving,
whereas on \llamaB{} the unsupervised and per-step chains settle from the first
step (\cosSlotCocoGsmLlTone{}, \cosSlotSimSixGsmLlTone{}) and only \codi{}
keeps moving (Figure~\ref{fig:rqthree}). The pre-registered reading for this
instrument (settled when the curve stays inside the null band) misfires,
because a trajectory that stops moving has cosine near one and leaves the band
upward; no settling verdict in the body rests on it.

\subsection{Answer alignment: fitting point and nulls}
\label{app:reconciliation}

The answer direction of Section~\ref{sec:t4} was pre-registered at the first
latent slot, where it falls below its MDE on every cell that assembles its
answer across slots (\simcoco{}/\prosqa{} reads \tFourSimProsqaAcc{} there and
\tFourLSimProsqaAcc{} at the last slot). It is therefore fitted at the
\tFourFitSlot{}, fixed for all \tFourCellsPCA{} cells before any verdict was
read; \tFourMovedTotal{} of \tFourRows{} cell$\times K$ verdicts moved
(\tFourMovedGain{} gained a reading, \tFourMovedLoss{} lost one). The
pre-registered cross-problem null is answer-aligned almost by construction,
since the answer direction separates problems, so the body reads a null that
permutes the answer labels and refits the direction with the same estimator;
\tFourNullDependent{} of \tFourRows{} verdicts differ between the two nulls,
and Table~\ref{tab:reconcileboth} gives every cell under both. On
\simcoco{}/\prosqa{} the along-component is \tFourLSimProsqaRatio{} times its
null under the probe of record and \tFourLRSimProsqaRatio{} under ridge, with a
state lift of \tFourLSimProsqaLift{} against an MDE of \tFourLiftMDEmin{}.

\begin{table*}[htbp]
\centering\scriptsize\setlength{\tabcolsep}{3.5pt}
\caption{\textbf{Reconciliation at the last-slot fit under both nulls,} all $K$.
\emph{probe}: accuracy of the answer direction at the last slot;
\emph{along}: the along-component of consecutive-slot displacement;
\emph{ratio}: along over the refit-null mean; (i)~refit\,/\,cross: whether
the alignment condition fires under the label-permuted refit null and under
the pre-registered cross-problem null; (ii): whether state identity lifts
answer prediction above the majority prior by its MDE at $K{=}5/8/12$.
\emph{dist.} distinct, \emph{mix.} mixed, \emph{re-m.} re-measurement,
\emph{unr.} unresolved; n/a: direction below its MDE at the last slot.}
\label{tab:reconcileboth}
\resizebox{\ifdim\width>\textwidth\textwidth\else\width\fi}{!}{%
\begin{tabular}{@{}l l r r r c c l l l@{}}
\toprule
\textbf{Cell} & \textbf{arm} & \textbf{probe} & \textbf{along} & \textbf{ratio} & \textbf{(i) refit} & \textbf{(i) cross} & \textbf{(ii) 5/8/12} & \textbf{verdict, refit} & \textbf{verdict, cross} \\
\midrule
\coconut{}/\prosqa{} & record & \tFourLCocoProsqaAcc & \tFourLCocoProsqaAlong & \tFourLCocoProsqaRatio & no & no & no/yes/yes & unr./mix./mix. & dist./mix./mix. \\
 & ridge & \tFourLRCocoProsqaAcc & \tFourLRCocoProsqaAlong & \tFourLRCocoProsqaRatio & no & no & no/yes/yes & dist./mix./mix. & dist./mix./mix. \\
\coconut{}$'$/\prosqa{} & record & \tFourLCocoProsqaBAcc & \tFourLCocoProsqaBAlong & \tFourLCocoProsqaBRatio & yes & no & yes/yes/yes & re-m.\ $\times 3$ & mix.\ $\times 3$ \\
 & ridge & \tFourLRCocoProsqaBAcc & \tFourLRCocoProsqaBAlong & \tFourLRCocoProsqaBRatio & no & no & yes/yes/yes & mix.\ $\times 3$ & mix.\ $\times 3$ \\
\coconut{}/\prosqa{}, \llamaB{} & record & \tFourLCocoProsqaLlAcc & \tFourLCocoProsqaLlAlong & \tFourLCocoProsqaLlRatio & yes & no & yes/no/yes & re-m./mix./re-m. & mix./dist./mix. \\
 & ridge & \tFourLRCocoProsqaLlAcc & \tFourLRCocoProsqaLlAlong & \tFourLRCocoProsqaLlRatio & no & no & yes/no/yes & mix./dist./mix. & mix./dist./mix. \\
\simcoco{}/\prosqa{} & record & \tFourLSimProsqaAcc & \tFourLSimProsqaAlong & \tFourLSimProsqaRatio & yes & no & no/no/no & mix.\ $\times 3$ & dist.\ $\times 3$ \\
 & ridge & \tFourLRSimProsqaAcc & \tFourLRSimProsqaAlong & \tFourLRSimProsqaRatio & yes & no & no/no/no & mix.\ $\times 3$ & dist.\ $\times 3$ \\
\codi{}/\prosqa{} & record & \tFourLCodiProsqaAcc & \tFourLCodiProsqaAlong & \tFourLCodiProsqaRatio & no & no & no/yes/no & unr./mix./unr. & dist./mix./dist. \\
 & ridge & \tFourLRCodiProsqaAcc & --- & --- & --- & --- & no/yes/no & n/a & n/a \\
\codi{}/\prosqa{}, \llamaB{} & record & \tFourLCodiProsqaLlAcc & --- & --- & --- & --- & no/no/no & n/a & n/a \\
 & ridge & \tFourLRCodiProsqaLlAcc & --- & --- & --- & --- & no/no/no & n/a & n/a \\
\midrule
\coconut{}/\gsm{} & record & \tFourLCocoGsmAcc & \tFourLCocoGsmAlong & \tFourLCocoGsmRatio & no & no & no/no/no & dist.\ $\times 3$ & dist.\ $\times 3$ \\
\simcoco{}/\gsm{} (\slotsSimOurs{} sl.) & record & \tFourLSimSixGsmAcc & \tFourLSimSixGsmAlong & \tFourLSimSixGsmRatio & no & no & no/no/no & dist.\ $\times 3$ & dist.\ $\times 3$ \\
\simcoco{}/\gsm{} (\slotsSimReleased{} sl.) & record & \tFourLSimTenGsmAcc & \tFourLSimTenGsmAlong & \tFourLSimTenGsmRatio & no & no & no/no/no & unr.\ $\times 3$ & dist.\ $\times 3$ \\
\codi{}/\gsm{} & record & \tFourLCodiGsmAcc & \tFourLCodiGsmAlong & \tFourLCodiGsmRatio & no & no & no/no/yes & unr./unr./mix. & dist./dist./mix. \\
\coconut{}/\gsm{}, \llamaB{} & record & \tFourLCocoGsmLlAcc & \tFourLCocoGsmLlAlong & \tFourLCocoGsmLlRatio & no & no & no/no/yes & dist./dist./mix. & dist./dist./mix. \\
\codi{}/\gsm{}, \llamaB{} & record & \tFourLCodiGsmLlAcc & \tFourLCodiGsmLlAlong & \tFourLCodiGsmLlRatio & no & no & no/no/yes & unr./unr./mix. & dist./dist./mix. \\
\midrule
\coconut{}/\prontoqa{} & record & \tFourLCocoProntoAcc & \tFourLCocoProntoAlong & \tFourLCocoProntoRatio & no & no & no/yes/yes & unr./mix./mix. & dist./mix./mix. \\
\codi{}/\prontoqa{} & record & \tFourLCodiProntoAcc & --- & --- & --- & --- & no/yes/yes & n/a & n/a \\
\coconut{}/\prontoqa{}, \llamaB{} & record & \tFourLCocoProntoLlAcc & \tFourLCocoProntoLlAlong & \tFourLCocoProntoLlRatio & yes & no & yes/yes/yes & re-m.\ $\times 3$ & mix.\ $\times 3$ \\
\codi{}/\prontoqa{}, \llamaB{} & record & \tFourLCodiProntoLlAcc & \tFourLCodiProntoLlAlong & \tFourLCodiProntoLlRatio & no & no & no/no/no & unr.\ $\times 3$ & dist.\ $\times 3$ \\
\simcoco{}/\prontoqa{} & record & \tFourLSimProntoAcc & \tFourLSimProntoAlong & \tFourLSimProntoRatio & no & yes & no/no/no & dist.\ $\times 3$ & mix.\ $\times 3$ \\
\bottomrule
\end{tabular}}
\end{table*}

\subsection{Mid-chain edits: commitment and backbone comparison}
\label{app:edits}

\begin{table*}[htbp]
\centering\small
\caption{\textbf{Commitment on the GPT-2 grid:} adaptive and rigid rates by
injection position $i$ on ceiling-solvable problems. \coconut{} and \codi{}
rows use the enlarged \twinNenlarged{}-twin set, \simcoco{} rows the
original \Neval{}-twin set; $i{=}0$ rates are $1.000$/$0.000$ by
construction. Terminal labels are descriptive only.}
\label{tab:commitment}
\resizebox{\ifdim\width>\textwidth\textwidth\else\width\fi}{!}{%
\begin{tabular}{@{}lcccl@{}}
\toprule
\textbf{Cell} & \textbf{usable $n$} & \textbf{adaptive} $i{=}0/2/4$ & \textbf{rigid} $i{=}0/2/4$ & \textbf{Terminal} \\
\midrule
\coconut{}/\gsm{} & \cocogsmEnlUsable & $1.000$ / \cocogsmEnlAtwo{} / \cocogsmEnlAfour & $0.000$ / \cocogsmEnlRtwo{} / \cocogsmEnlRfour & rigid-dominant \\
\simcoco{}/\gsm{} & \simSixGsmUsable & $1.000$ / \simSixGsmAtwo{} / \simSixGsmAfour & $0.000$ / \simSixGsmRtwo{} / \simSixGsmRfour & rigid-dominant \\
\simcoco{}/\prosqa{} & \simProsqaUsable & $1.000$ / \simProsqaAtwo{} / \simProsqaAfour & $0.000$ / \simProsqaRtwo{} / \simProsqaRfour & rigid-dominant \\
\codi{}/\gsm{} & \codigsmEnlUsable & $1.000$ / \codigsmEnlAtwo{} / \codigsmEnlAfour & $0.000$ / \codigsmEnlRtwo{} / \codigsmEnlRfour & corruption-dominant \\
\bottomrule
\end{tabular}}
\end{table*}

\begin{table*}[htbp]
\centering\small
\caption{\textbf{The paired backbone comparison} on the intersection of the two
models' ceiling subsets after the enlargement to \twinNenlarged{} twins.
\emph{discordant}: problems adaptive on exactly one backbone
(GPT-2-only\,/\,\llamaB{}-only); $p$ is Holm-corrected McNemar.}
\label{tab:backbone}
\begin{tabular}{@{}lcccccc@{}}
\toprule
\textbf{Paired contrast} & $n$ & \textbf{GPT-2} & \textbf{\llamaB{}} & \textbf{discordant} & \textbf{Holm $p$} & \textbf{gap vs MDE} \\
\midrule
\coconut{}, $i{=}2$ & \pairCocoN & \pairCocoTwoGpt  & \pairCocoTwoLl  & \pairCocoTwoDisc  & \pairCocoTwoP  & $\pairCocoTwoGapv > \pairCocoTwoMDE$ \\
\coconut{}, $i{=}4$ & \pairCocoN & \pairCocoFourGpt & \pairCocoFourLl & \pairCocoFourDisc & \pairCocoFourP & $\pairCocoFourGapv > \pairCocoFourMDE$ \\
\codi{}, $i{=}2$    & \pairCodiN & \pairCodiTwoGpt  & \pairCodiTwoLl  & \pairCodiTwoDisc  & \pairCodiTwoP  & $\pairCodiTwoGapv > \pairCodiTwoMDE$ \\
\codi{}, $i{=}4$    & \pairCodiN & \pairCodiFourGpt & \pairCodiFourLl & \pairCodiFourDisc & \pairCodiFourP & $\pairCodiFourGapv > \pairCodiFourMDE$ \\
\bottomrule
\end{tabular}
\end{table*}

Adaptation falls monotonically with injection position in every cell
(Table~\ref{tab:commitment}). Under identical edits on identical problems the
larger instruction-tuned backbone re-computes where GPT-2 holds or corrupts,
with the latent mechanism held fixed; every gap in Table~\ref{tab:backbone}
clears its own minimum detectable effect. A pre-registered drift readout is
degenerate by construction on the recycling family, whose edit driver takes
every post-edit slot from the twin's clean run, so the body reads only the
answer-side rate $A(i)$.

%% file: appendix/e_checkpoints.tex
\section{Additional checkpoints}
\label{app:checkpoints}

\subsection{Per-step supervision on \llamaB{}}
\label{app:llamaperstep}

We trained the per-step recipe on \llamaB{} for \prosqa{}, with a
training-time control that continues the same base for the same schedule with
the auxiliary loss weighted to zero ($N{=}\Neval{}$, seed \seedval{}). The
\prontoqa{} training saved \lTwoSweepTotal{} checkpoints; a sweep on 60
held-out problems, run before any intervention, found \lTwoSweepConstant{}
emitting a single answer string and selected checkpoint \simProntoLlCkpt{},
the earliest with both answer classes present (sweep accuracy
\lTwoSweepAccNine{}). The \gsm{} twin was trained last and has no
training-time control; its battery is in Tables~\ref{tab:truncation},
\ref{tab:corruption} and~\ref{tab:twobytwo}, and its answer magnitude rises
from \accumFirstSimSixGsmLl{} at slot 0 to \accumLastSimSixGsmLl{} at slot 5,
at the readout's resolution (MDD \accumMddSimSixGsmLl{}).
Table~\ref{tab:llamaperstep} collects the \prosqa{} and \prontoqa{} cells.

Two caveats bind these readings. No substitute moves an answer in any
\llamaB{}/\prosqa{} cell, even a donor with a different gold answer at twice
the natural norm (\simProsqaLlMovedMax{} of \Neval{}), while the same code
moves \gptTwinMovedRandom{}, \gptTwinMovedMean{} and \gptTwinMovedDonor{}
answers on the GPT-2 twin; the placeholder verdicts therefore bound the range
the harness can express on this backbone. The paired backbone comparison for
the per-step family is descriptive only: its twin sets (\pairSimTwinsLl{},
\pairSimTwinsGpt{} and \pairSimTwinsCtrl{} usable problems) are too small to
form the test of Table~\ref{tab:backbone}. On them the \llamaB{} twin and its
control follow every edit (\pairSimFourLl{} at $i{=}4$), while the GPT-2 twin
falls to \pairSimFourGpt{}.

\begin{table}[htbp]
\centering\scriptsize
\setlength{\tabcolsep}{2.5pt}
\caption{\textbf{The \llamaB{} per-step cells} (\prosqa{} and \prontoqa{},
$N{=}\Neval{}$). Deletion: $C_0$, $C_1$, helped/hurt. Truncation: accuracy
with one, two and four slots kept. Substitution: answers moved under each
substitute. Content on \prosqa{}: gold readability at slots 0 and 5, probe of
record\,/\,ridge (chance \chanceProsqaConcept{}; ceilings
\llamaCeilPCA{}\,/\,\llamaCeilRidge{}), path cells live of 24. Dynamics:
distinct states at $K{=}5$, slots settled before the null (floor
\dynFloorSlots{}), along-answer component at the last slot against the refit
and cross-problem nulls.}
\label{tab:llamaperstep}
\begin{tabular}{@{}p{2.3cm} p{1.75cm} p{1.65cm} p{1.4cm}@{}}
\toprule
& \textbf{per-step, \prosqa{}} & \textbf{control, \prosqa{}} & \textbf{per-step, \prontoqa{}} \\
\midrule
$C_0$ / $C_1$ & \simProsqaLlCzero{} / \simProsqaLlCone{} & \ctrlLlCzero{} / \ctrlLlCone{} & \simProntoLlCzero{} / \simProntoLlCone{} \\
helped / hurt & \simProsqaLlHelped{} / \simProsqaLlHurt{} & \ctrlLlHelped{} / \ctrlLlHurt{} & \simProntoLlHelped{} / \simProntoLlHurt{} \\
test & Holm $p{=}\simProsqaLlHolmP{}$ & $\Delta{=}\ctrlLlDelta{}$, MDE \ctrlLlMDE{} & $\Delta{=}\simProntoLlDelta{}$, MDE \simProntoLlMDE{} \\
node 1 & consulted & unresolved & inert (\simProntoLlMoved{} of \Neval{} changed, within the rule-of-three bound) \\
kept 1 / 2 / 4 slots & \simProsqaLlKone{} / \simProsqaLlKtwo{} / \simProsqaLlKfour{} & \ctrlLlKone{} / \ctrlLlKtwo{} / \ctrlLlKfour{} & --- / \simProntoLlKtwo{} / \simProntoLlKfour{} \\
moved under substitution & \simProsqaLlMoved{} of \Neval{} & \ctrlLlMoved{} of \Neval{} & --- \\
node 2 & placeholder (caveat) & --- & --- \\
answer at slot 0 $\to$ 5 & \simProsqaLlFitPzero{} $\to$ \simProsqaLlFitPfive{} / \simProsqaLlFitPzeroR{} $\to$ \simProsqaLlFitPfiveR{} & \ctrlLlFitPzero{} $\to$ \ctrlLlFitPfive{} / \ctrlLlFitPzeroR{} $\to$ \ctrlLlFitPfiveR{} & n/a \\
both candidates (bar \superposthresh{}) & \simProsqaLlBoth{} / \simProsqaLlBothR{} & \ctrlLlBoth{} / \ctrlLlBothR{} & n/a \\
path cells live / 24 & \simProsqaLlLive{} / \simProsqaLlLiveR{} & \ctrlLlLive{} / \ctrlLlLiveR{} & n/a \\
layout & localised and ordered & neighbourhood, not chain & n/a \\
distinct states ($K{=}5$) & \simProsqaLlStatesFive{} & \ctrlLlStatesFive{} & \simProntoLlStatesFive{} \\
settled before null & \simProsqaLlDeltaK{} (at null) & \ctrlLlDeltaK{} (early) & undefined \\
along-answer / refit / cross & \simProsqaLlAlong{} / \simProsqaLlRefitHi{} / \simProsqaLlCrossLo{} & \ctrlLlAlong{} / \ctrlLlRefitHi{} / \ctrlLlCrossLo{} & at chance \\
\bottomrule
\end{tabular}
\end{table}

\subsection{The released ten-slot \simcoco{} checkpoint}
\label{app:released}

The publicly released \simcoco{} checkpoint (\gsm{}, \slotsSimReleased{}
slots) differs in slot budget from every other cell and is kept out of the
body. It is the parity anchor for our six-slot training and an independent
replication of the step-supervised regime on \gsm{}: deleting its chain flips
\simgsmHelped{} of \Neval{} answers ($C_0$ \simgsmCzero{}, $C_1$
\simgsmCone{}), truncation rises monotonically, and a donor's thoughts are
worse than deletion at a pre-registered $N{=}\simgsmPoisonN{}$
($p{=}\simgsmPoisonP$). Its commitment window spans the whole chain, with
adaptation falling to \simgsmAeight{} at $i{=}8$.

%% file: appendix/f_census.tex
\section{Classification rules and census}
\label{app:worked}

\subsection{Classification rules}
\label{app:taxonomy_rules}

\textbf{Node 1: delete the chain.} If at most \inertBound{} answers change
(the rule-of-three bound on an absent effect at $N{=}\Neval{}$), the cell is
\textbf{inert}; if it loses correct answers beyond its MDE, it is
\textbf{consulted}; answers that move both ways without a net direction leave
it unresolved.

\textbf{Node 2: replace the contents.} A consulted cell keeps every slot but
has its contents replaced by noise, the dataset mean or another problem's
slots. If no substitute moves an answer, it is a \textbf{placeholder}, the
filler-token regime of \citet{pfau2024dotbydot} and \citet{goyal2024pause}
realised by a trained chain; if substitutes move answers beyond the MDE, it is
\textbf{content used}; otherwise it is unresolved.

\textbf{Profile.} Within each outcome, \emph{readability} (whether a linear
probe decodes the answer or the step-by-step path) and \emph{state dynamics}
(whether the state keeps changing across slots) describe the chain; the
profile never overrides the nodes. The two readings agree in every cell where
both resolve, so Figure~\ref{fig:tree} labels each branch with one of them; a
cell where they disagreed would name no leaf.

\begin{itemize}[leftmargin=*,noitemsep,topsep=2pt]
  \item \textbf{Carrier} (inert, readable): the answer is readable from the
        first slot and the state settles early.
  \item \textbf{Empty} (inert, unreadable): little is readable and the state
        settles early.
  \item \textbf{Bare placeholder} (placeholder, state fixed): the positions
        are needed, little is readable, and the state settles early.
  \item \textbf{Laid-out placeholder} (placeholder, state changing): step $k$
        is readable at slot $k$ and the state changes through the last slot,
        yet no substitute moves the answer.
  \item \textbf{Assembling} (content used, structured): step $k$ is readable
        at slot $k$, the state changes through the last slot, and the
        contents affect the answer.
  \item \textbf{Unstructured} (content used, unstructured): the contents are
        consulted, but no step-by-slot layout is read.
\end{itemize}

\subsection{Cell assignments}

\begin{table*}[htbp]
\centering\small\setlength{\tabcolsep}{4pt}
\caption{\textbf{The twenty-one cells on the tree.} Node 1: the deletion verdict; node 2:
the substitution verdict (consulted cells only); profile: the RQ2/RQ3
reading; the leaf follows from the tree. ``---'': not reached; ``n/a'': not
applicable. The second \coconut{}/\prosqa{} training and the two training-time
controls are listed as cells of their own.}
\label{tab:census}
\resizebox{\ifdim\width>\textwidth\textwidth\else\width\fi}{!}{%
\begin{tabular}{@{}llllll@{}}
\toprule
\textbf{Backbone} & \textbf{Cell} & \textbf{Node 1} & \textbf{Node 2} & \textbf{Profile} & \textbf{Leaf} \\
\midrule
GPT-2 & \coconut{}/\prosqa{}          & inert      & ---       & readable, state fixed     & carrier \\
GPT-2 & \coconut{}/\prosqa{} (2nd)    & inert      & ---       & readable, state fixed     & carrier \\
GPT-2 & \coconut{}/\prontoqa{}        & inert      & ---       & readable, state fixed     & carrier \\
GPT-2 & control/\prosqa{}             & inert      & ---       & state fixed, readability n/a & inert, unprofiled \\
GPT-2 & \codi{}/\prontoqa{}           & inert      & ---       & unreadable, state fixed   & empty \\
GPT-2 & \codi{}/\prosqa{}             & unresolved & ---       & ---                       & --- \\
GPT-2 & \coconut{}/\gsm{}             & consulted  & content   & unstructured              & unstructured \\
GPT-2 & \codi{}/\gsm{}                & consulted  & content   & unstructured              & unstructured \\
GPT-2 & \simcoco{}/\gsm{}             & consulted  & content   & unstructured              & unstructured \\
GPT-2 & \simcoco{}/\prosqa{}          & consulted  & content   & structured, state changing & assembling \\
GPT-2 & \simcoco{}/\prontoqa{}        & consulted  & content   & structured, state changing & assembling \\
\midrule
\llamaBshort{} & \coconut{}/\prosqa{}   & inert      & ---       & readable, state fixed     & carrier \\
\llamaBshort{} & \coconut{}/\prontoqa{} & inert      & ---       & readable, state fixed     & carrier \\
\llamaBshort{} & \codi{}/\prontoqa{}    & inert      & ---       & unreadable, state fixed   & empty \\
\llamaBshort{} & \codi{}/\prosqa{}      & consulted  & placeholder & unreadable, state fixed & bare placeholder \\
\llamaBshort{} & \simcoco{}/\prosqa{}   & consulted  & placeholder & structured, state changing & laid-out placeholder \\
\llamaBshort{} & \codi{}/\gsm{}         & consulted  & content   & unstructured              & unstructured \\
\llamaBshort{} & \coconut{}/\gsm{}      & consulted  & content   & unstructured              & unstructured \\
\llamaBshort{} & \simcoco{}/\gsm{}      & consulted  & content   & unstructured              & unstructured \\
\llamaBshort{} & control/\prosqa{}      & unresolved & ---       & ---                       & --- \\
\llamaBshort{} & \simcoco{}/\prontoqa{} & inert      & ---       & state fixed, readability n/a & inert, unprofiled \\
\bottomrule
\end{tabular}}
\end{table*}

%% file: main.bbl
\begin{thebibliography}{45}
\providecommand{\natexlab}[1]{#1}
\providecommand{\url}[1]{\texttt{#1}}
\expandafter\ifx\csname urlstyle\endcsname\relax
  \providecommand{\doi}[1]{doi: #1}\else
  \providecommand{\doi}{doi: \begingroup \urlstyle{rm}\Url}\fi

\bibitem[Aswal et~al.(2026)Aswal, Ferraz, Zhou, and Peyrard]{aswal2026observable}
Darpan Aswal, Thomas~Palmeira Ferraz, Yongxin Zhou, and Maxime Peyrard.
\newblock Observable patterns are not explanations: A causal-geometric analysis of latent reasoning models.
\newblock \emph{arXiv preprint arXiv:2606.12689}, 2026.

\bibitem[Cheng \& Van~Durme(2024)Cheng and Van~Durme]{cheng2024ccot}
Jeffrey Cheng and Benjamin Van~Durme.
\newblock Compressed chain of thought: Efficient reasoning through dense representations.
\newblock \emph{arXiv preprint arXiv:2412.13171}, 2024.

\bibitem[Cobbe et~al.(2021)]{cobbe2021gsm8k}
Karl Cobbe et~al.
\newblock Training verifiers to solve math word problems.
\newblock \emph{arXiv preprint arXiv:2110.14168}, 2021.

\bibitem[Cui et~al.(2026)Cui, Dai, He, Shi, Liu, Sun, Liu, Xing, Tang, and Dumoulin]{cui2026supervision}
Yingqian Cui, Zhenwei Dai, Bing He, Zhan Shi, Hui Liu, Rui Sun, Zhiji Liu, Yue Xing, Jiliang Tang, and Benoit Dumoulin.
\newblock How do latent reasoning methods perform under weak and strong supervision?
\newblock \emph{arXiv preprint arXiv:2602.22441}, 2026.

\bibitem[Deng et~al.(2024)Deng, Choi, and Shieber]{deng2024implicit}
Yuntian Deng, Yejin Choi, and Stuart Shieber.
\newblock From explicit {CoT} to implicit {CoT}: Learning to internalize {CoT} step by step.
\newblock \emph{arXiv preprint arXiv:2405.14838}, 2024.

\bibitem[Dilgren \& Wiegreffe(2026)Dilgren and Wiegreffe]{dilgren2026interpretable}
Connor Dilgren and Sarah Wiegreffe.
\newblock Are latent reasoning models easily interpretable?
\newblock \emph{arXiv preprint arXiv:2604.04902}, 2026.

\bibitem[Duraipandian et~al.(2026)Duraipandian, Boyane, Nagesh, Francis, Vaidheeswaran, and Zhu]{duraipandian2026dynamical}
Sabari~Iyyappan Duraipandian, Shreya~Sanjay Boyane, Manju Nagesh, Jerome Francis, Archana Vaidheeswaran, and Kevin Zhu.
\newblock Interpreting latent {CoT} reasoning as dynamical systems.
\newblock In \emph{ICML 2026 Workshop on Foundations of Deep Generative Models (FoGen)}, 2026.
\newblock arXiv:2607.09698.

\bibitem[Elazar et~al.(2021)Elazar, Ravfogel, Jacovi, and Goldberg]{elazar2021amnesic}
Yanai Elazar, Shauli Ravfogel, Alon Jacovi, and Yoav Goldberg.
\newblock Amnesic probing: Behavioral explanation with amnesic counterfactuals.
\newblock \emph{Transactions of the Association for Computational Linguistics}, 9:\penalty0 160--175, 2021.

\bibitem[Geiping et~al.(2025)Geiping, McLeish, Jain, Kirchenbauer, Singh, Bartoldson, Kailkhura, Bhatele, and Goldstein]{geiping2025recurrent}
Jonas Geiping, Sean McLeish, Neel Jain, John Kirchenbauer, Siddharth Singh, Brian~R. Bartoldson, Bhavya Kailkhura, Abhinav Bhatele, and Tom Goldstein.
\newblock Scaling up test-time compute with latent reasoning: A recurrent depth approach.
\newblock In \emph{Advances in Neural Information Processing Systems (NeurIPS)}, 2025.

\bibitem[Goyal et~al.(2024)Goyal, Ji, Rawat, Menon, Kumar, and Nagarajan]{goyal2024pause}
Sachin Goyal, Ziwei Ji, Ankit~Singh Rawat, Aditya~Krishna Menon, Sanjiv Kumar, and Vaishnavh Nagarajan.
\newblock Think before you speak: Training language models with pause tokens.
\newblock In \emph{International Conference on Learning Representations (ICLR)}, 2024.

\bibitem[Hao et~al.(2025)Hao, Sukhbaatar, Su, Li, Hu, Weston, and Tian]{hao2024coconut}
Shibo Hao, Sainbayar Sukhbaatar, DiJia Su, Xian Li, Zhiting Hu, Jason Weston, and Yuandong Tian.
\newblock Training large language models to reason in a continuous latent space.
\newblock In \emph{Conference on Language Modeling (COLM)}, 2025.

\bibitem[Huang et~al.(2025{\natexlab{a}})Huang, Wu, Xia, Yu, Wang, Yu, Zhang, Rossi, Kveton, Zhou, McAuley, and Yao]{huang2025agenticrec}
Chengkai Huang, Junda Wu, Yu~Xia, Zixu Yu, Ruhan Wang, Tong Yu, Ruiyi Zhang, Ryan~A. Rossi, Branislav Kveton, Dongruo Zhou, Julian McAuley, and Lina Yao.
\newblock Towards agentic recommender systems in the era of multimodal large language models.
\newblock \emph{arXiv preprint arXiv:2503.16734}, 2025{\natexlab{a}}.

\bibitem[Huang et~al.(2025{\natexlab{b}})Huang, Wu, Surana, Jain, Yu, Addanki, Arbour, Kim, and McAuley]{huang2025traceable}
Zihan Huang, Junda Wu, Rohan Surana, Raghav Jain, Tong Yu, Raghavendra Addanki, David Arbour, Sungchul Kim, and Julian McAuley.
\newblock Traceable and explainable multimodal large language models: An information-theoretic view.
\newblock In \emph{Second Conference on Language Modeling (COLM)}, 2025{\natexlab{b}}.

\bibitem[Huang et~al.(2026)Huang, Li, Surana, Yu, Wang, McAuley, Shang, and Wu]{huang2026amps}
Zihan Huang, Xintong Li, Rohan Surana, Tong Yu, Rui Wang, Julian McAuley, Jingbo Shang, and Junda Wu.
\newblock {AMPS}: Adaptive modality preference steering via functional entropy.
\newblock \emph{arXiv preprint arXiv:2602.12533}, 2026.

\bibitem[Jiang et~al.(2026)Jiang, Wu, Xiong, Yu, Yu, Arbour, Sinha, McAuley, and Wen]{jiang2026latentskills}
Xunyi Jiang, Junda Wu, Yuxin Xiong, Sheldon Yu, Tong Yu, David Arbour, Ritwik Sinha, Julian McAuley, and Hongyi Wen.
\newblock Toward latent language model skills steering and optimization: An empirical study.
\newblock \emph{arXiv preprint arXiv:2608.29459}, 2026.

\bibitem[Lanham et~al.(2023)Lanham, Chen, Radhakrishnan, Steiner, Denison, Hernandez, Li, Durmus, Hubinger, Kernion, et~al.]{lanham2023faithfulness}
Tamera Lanham, Anna Chen, Ansh Radhakrishnan, Benoit Steiner, Carson Denison, Danny Hernandez, Dustin Li, Esin Durmus, Evan Hubinger, Jackson Kernion, et~al.
\newblock Measuring faithfulness in chain-of-thought reasoning.
\newblock \emph{arXiv preprint arXiv:2307.13702}, 2023.

\bibitem[Li et~al.(2026)Li, Bai, Chen, Li, Yang, Lin, and Zhang]{li2026dynamics}
Zirui Li, Xuefeng Bai, Kehai Chen, Yizhi Li, Jian Yang, Chenghua Lin, and Min Zhang.
\newblock Dynamics within latent chain-of-thought: An empirical study of causal structure.
\newblock In \emph{Proceedings of the 43rd International Conference on Machine Learning (ICML)}, 2026.

\bibitem[Liu et~al.(2025)Liu, Xu, Wu, Yuan, Yang, Zhou, Liu, Guan, Wang, Yu, McAuley, Ai, and Huang]{liu2025causalsurvey}
Xiaoyu Liu, Paiheng Xu, Junda Wu, Jiaxin Yuan, Yifan Yang, Yuhang Zhou, Fuxiao Liu, Tianrui Guan, Haoliang Wang, Tong Yu, Julian McAuley, Wei Ai, and Furong Huang.
\newblock Large language models and causal inference in collaboration: A comprehensive survey.
\newblock In \emph{Findings of the Association for Computational Linguistics: NAACL 2025}, pp.\  7668--7684, 2025.

\bibitem[Lu et~al.(2024)Lu, Zeng, Ma, Yu, and Levorato]{lu2024multicandidate}
Xiaofan Lu, Yixiao Zeng, Feiyang Ma, Zixu Yu, and Marco Levorato.
\newblock Improving multi-candidate speculative decoding.
\newblock In \emph{NeurIPS 2024 Workshop on Efficient Natural Language and Speech Processing (ENLSP)}, 2024.
\newblock arXiv:2409.10644.

\bibitem[Mundada et~al.(2026)Mundada, Huang, Surana, Yu, Zhang, Li, Yu, Yao, Shang, McAuley, and Wu]{mundada2026wsgrpo}
Gagan Mundada, Zihan Huang, Rohan Surana, Sheldon Yu, Jennifer~Yuntong Zhang, Xintong Li, Tong Yu, Lina Yao, Jingbo Shang, Julian McAuley, and Junda Wu.
\newblock {WS-GRPO}: Weakly-supervised group-relative policy optimization for rollout-efficient reasoning.
\newblock \emph{arXiv preprint arXiv:2602.17025}, 2026.

\bibitem[Pal et~al.(2023)Pal, Sun, Yuan, Wallace, and Bau]{pal2023futurelens}
Koyena Pal, Jiuding Sun, Andrew Yuan, Byron~C. Wallace, and David Bau.
\newblock Future lens: Anticipating subsequent tokens from a single hidden state.
\newblock In \emph{Proceedings of the 27th Conference on Computational Natural Language Learning (CoNLL)}, 2023.

\bibitem[Pfau et~al.(2024)Pfau, Merrill, and Bowman]{pfau2024dotbydot}
Jacob Pfau, William Merrill, and Samuel~R. Bowman.
\newblock Let's think dot by dot: Hidden computation in transformer language models.
\newblock In \emph{First Conference on Language Modeling (COLM)}, 2024.

\bibitem[Ravichander et~al.(2021)Ravichander, Belinkov, and Hovy]{ravichander2021probing}
Abhilasha Ravichander, Yonatan Belinkov, and Eduard Hovy.
\newblock Probing the probing paradigm: Does probing accuracy entail task relevance?
\newblock In \emph{Proceedings of the 16th Conference of the European Chapter of the Association for Computational Linguistics (EACL)}, pp.\  3363--3377, 2021.

\bibitem[Rizvi-Martel et~al.(2026)Rizvi-Martel, Rabusseau, and Mosbach]{rizvimartel2026illusion}
Michael Rizvi-Martel, Guillaume Rabusseau, and Marius Mosbach.
\newblock The illusion of superposition? a principled analysis of latent thinking in language models.
\newblock \emph{arXiv preprint arXiv:2604.06374}, 2026.

\bibitem[Saparov \& He(2023)Saparov and He]{saparov2023prontoqa}
Abulhair Saparov and He~He.
\newblock Language models are greedy reasoners: A systematic formal analysis of chain-of-thought.
\newblock In \emph{International Conference on Learning Representations (ICLR)}, 2023.

\bibitem[Shen et~al.(2025)Shen, Yan, Zhang, Hu, Du, and He]{shen2025codi}
Zhenyi Shen, Hanqi Yan, Linhai Zhang, Zhanghao Hu, Yali Du, and Yulan He.
\newblock {CODI}: Compressing chain-of-thought into continuous space via self-distillation.
\newblock In \emph{Proceedings of the 2025 Conference on Empirical Methods in Natural Language Processing (EMNLP)}, pp.\  677--693, 2025.

\bibitem[Surana et~al.(2026{\natexlab{a}})Surana, Li, Yu, Shen, Wang, Yu, Ammanabrolu, Shang, McAuley, and Wu]{surana2026massdpo}
Rohan Surana, Xintong Li, Sheldon Yu, Yiran~Jenny Shen, Chuhan Wang, Tong Yu, Prithviraj Ammanabrolu, Jingbo Shang, Julian McAuley, and Junda Wu.
\newblock {MASS-DPO}: Multi-negative active sample selection for direct policy optimization.
\newblock \emph{arXiv preprint arXiv:2605.10784}, 2026{\natexlab{a}}.

\bibitem[Surana et~al.(2026{\natexlab{b}})Surana, Mundada, Jiang, Wang, Tang, Jiao, Huang, Xiong, Wu, Yu, Li, Jain, Kuang, Zhou, Jin, Chu, Yu, Rossi, Huang, Shang, Han, and McAuley]{surana2026rollout}
Rohan Surana, Gagan Mundada, Xunyi Jiang, Chuhan Wang, Zhenwei Tang, Difan Jiao, Zihan Huang, Yuxin Xiong, Junda Wu, Sheldon Yu, Xintong Li, Raghav Jain, Nikki Kuang, Sizhe Zhou, Bowen Jin, Zhendong Chu, Tong Yu, Ryan Rossi, Kuan-Hao Huang, Jingbo Shang, Jiawei Han, and Julian McAuley.
\newblock Generate, filter, control, replay: A comprehensive survey of rollout strategies for {LLM} reinforcement learning.
\newblock \emph{Transactions on Machine Learning Research (TMLR)}, 2026{\natexlab{b}}.
\newblock arXiv:2605.02913.

\bibitem[Wang et~al.(2025)Wang, Wu, Xia, Yu, Zhang, Rossi, Mitra, Yao, and McAuley]{wang2025cacheprune}
Rui Wang, Junda Wu, Yu~Xia, Tong Yu, Ruiyi Zhang, Ryan Rossi, Subrata Mitra, Lina Yao, and Julian McAuley.
\newblock {CachePrune}: Teaching {LLMs} what not to follow via {KV}-cache editing.
\newblock \emph{arXiv preprint arXiv:2504.21228}, 2025.

\bibitem[Wei et~al.(2026{\natexlab{a}})Wei, Wu, Yu, Zhou, Jiao, Zhong, Jin, Yu, Pan, Han, and McAuley]{wei2026reasoningenergy}
Hui Wei, Junda Wu, Sheldon Yu, Sizhe Zhou, Yizhu Jiao, Ming Zhong, Bowen Jin, Tong Yu, Shijia Pan, Jiawei Han, and Julian McAuley.
\newblock How hard does it think? analyzing step-aware reasoning energy in {LLM} chain-of-thought trajectories.
\newblock \emph{arXiv preprint arXiv:2607.28674}, 2026{\natexlab{a}}.

\bibitem[Wei et~al.(2026{\natexlab{b}})Wei, Liu, Zang, Dong, Cao, Wang, Qiu, and Lin]{wei2026simcot}
Xilin Wei, Xiaoran Liu, Yuhang Zang, Xiaoyi Dong, Yuhang Cao, Jiaqi Wang, Xipeng Qiu, and Dahua Lin.
\newblock {SIM-CoT}: Supervised implicit chain-of-thought.
\newblock In \emph{International Conference on Learning Representations (ICLR)}, 2026{\natexlab{b}}.

\bibitem[Wu et~al.(2023)Wu, Yu, Wang, Song, Zhang, Zhao, Lu, Li, and Henao]{wu2023infoprompt}
Junda Wu, Tong Yu, Rui Wang, Zhao Song, Ruiyi Zhang, Handong Zhao, Chaochao Lu, Shuai Li, and Ricardo Henao.
\newblock {InfoPrompt}: Information-theoretic soft prompt tuning for natural language understanding.
\newblock In \emph{Advances in Neural Information Processing Systems (NeurIPS)}, volume~36, pp.\  61060--61084, 2023.

\bibitem[Wu et~al.(2024)Wu, Yu, Chen, Wang, Rossi, Kim, Rao, and McAuley]{wu2024decot}
Junda Wu, Tong Yu, Xiang Chen, Haoliang Wang, Ryan Rossi, Sungchul Kim, Anup Rao, and Julian McAuley.
\newblock {DeCoT}: Debiasing chain-of-thought for knowledge-intensive tasks in large language models via causal intervention.
\newblock In \emph{Proceedings of the 62nd Annual Meeting of the Association for Computational Linguistics (Volume 1: Long Papers)}, pp.\  14073--14087, 2024.

\bibitem[Wu et~al.(2025)Wu, Li, Wang, Xia, Xiong, Wang, Yu, Chen, Kveton, Yao, Shang, and McAuley]{wu2025ocean}
Junda Wu, Xintong Li, Ruoyu Wang, Yu~Xia, Yuxin Xiong, Jianing Wang, Tong Yu, Xiang Chen, Branislav Kveton, Lina Yao, Jingbo Shang, and Julian McAuley.
\newblock {OCEAN}: Offline chain-of-thought evaluation and alignment in large language models.
\newblock In \emph{International Conference on Learning Representations (ICLR)}, 2025.

\bibitem[Wu et~al.(2026)Wu, Xiong, Li, Yu, Hu, Yu, Wang, Chen, Shang, and McAuley]{wu2025ctrls}
Junda Wu, Yuxin Xiong, Xintong Li, Sheldon Yu, Zhengmian Hu, Tong Yu, Rui Wang, Xiang Chen, Jingbo Shang, and Julian McAuley.
\newblock {CTRLS}: Chain-of-thought reasoning via latent state transition.
\newblock In \emph{Proceedings of the 29th International Conference on Artificial Intelligence and Statistics (AISTATS)}, 2026.

\bibitem[Xiong et~al.(2026)Xiong, Jiang, Surana, Li, Yu, Kuang, Rossi, Shang, Yu, McAuley, and Wu]{xiong2026rrpo}
Yuxin Xiong, Xunyi Jiang, Rohan Surana, Xintong Li, Sheldon Yu, Nikki~Lijing Kuang, Ryan~A. Rossi, Jingbo Shang, Tong Yu, Julian McAuley, and Junda Wu.
\newblock {RRPO}: Reference-relative policy optimization with stratified conditional rollouts.
\newblock \emph{arXiv preprint arXiv:2607.18470}, 2026.

\bibitem[Yang et~al.(2024)Yang, Gribovskaya, Kassner, Geva, and Riedel]{yang2024multihop}
Sohee Yang, Elena Gribovskaya, Nora Kassner, Mor Geva, and Sebastian Riedel.
\newblock Do large language models latently perform multi-hop reasoning?
\newblock In \emph{Proceedings of the 62nd Annual Meeting of the Association for Computational Linguistics (Volume 1: Long Papers)}, 2024.

\bibitem[Yu et~al.(2025)Yu, Xiong, Wu, Li, Yu, Chen, Sinha, Shang, and McAuley]{yu2025stateaware}
Sheldon Yu, Yuxin Xiong, Junda Wu, Xintong Li, Tong Yu, Xiang Chen, Ritwik Sinha, Jingbo Shang, and Julian McAuley.
\newblock Explainable chain-of-thought reasoning: An empirical analysis on state-aware reasoning dynamics.
\newblock In \emph{Findings of the Association for Computational Linguistics: EMNLP}, 2025.

\bibitem[Yu et~al.(2026{\natexlab{a}})Yu, Sun, Guo, and Tong]{yu2026fraud}
Sheldon Yu, Yingcheng Sun, Hanqing Guo, and Qianqian Tong.
\newblock A fraud-detection-inspired framework for {LLM} agents security.
\newblock In \emph{KDD 2026 Workshop on Evaluation and Trustworthiness of Agentic AI}, 2026{\natexlab{a}}.

\bibitem[Yu et~al.(2026{\natexlab{b}})Yu, Wang, Yu, Kim, Dogan, Wu, and McAuley]{yu2026agent2ucb}
Sheldon Yu, Rui Wang, Tong Yu, Sungchul Kim, Doga Dogan, Junda Wu, and Julian McAuley.
\newblock {Agent2UCB}: Agentic system for generative engine optimization.
\newblock \emph{arXiv preprint arXiv:2608.29063}, 2026{\natexlab{b}}.

\bibitem[Yu et~al.(2026{\natexlab{c}})Yu, Wu, Li, Kuang, Zhou, Yu, Han, Shang, and McAuley]{yu2026olivia}
Sheldon Yu, Junda Wu, Xintong Li, Nikki~Lijing Kuang, Sizhe Zhou, Tong Yu, Jiawei Han, Jingbo Shang, and Julian McAuley.
\newblock {OLIVIA}: Online learning via inference-time action adaptation for decision making in {LLM} {ReAct} agents.
\newblock In \emph{Proceedings of the 2026 Conference on Empirical Methods in Natural Language Processing (EMNLP)}, 2026{\natexlab{c}}.
\newblock arXiv:2605.11169.

\bibitem[Yu et~al.(2026{\natexlab{d}})Yu, Yu, Jiang, Surana, Mundada, Kim, Yao, McAuley, and Wu]{yu2026selfloops}
Sheldon Yu, Tong Yu, Xunyi Jiang, Rohan Surana, Gagan Mundada, Sungchul Kim, Lina Yao, Julian McAuley, and Junda Wu.
\newblock Can we break {LLMs} out of self-loops? fine-grained reasoning control with activation steering.
\newblock \emph{arXiv preprint arXiv:2607.18100}, 2026{\natexlab{d}}.

\bibitem[Zhang et~al.(2025)Zhang, Tang, Ju, Duan, and Liu]{zhang2025latenttokens}
Yuyi Zhang, Boyu Tang, Tianjie Ju, Sufeng Duan, and Gongshen Liu.
\newblock Do latent tokens think? a causal and adversarial analysis of chain-of-continuous-thought.
\newblock \emph{arXiv preprint arXiv:2512.21711}, 2025.

\bibitem[Zhou et~al.(2026)Zhou, Yu, Wei, Wu, Ouyang, Jiao, Pan, McAuley, Zhang, Yu, and Han]{zhou2026filesystem}
Sizhe Zhou, Sheldon Yu, Hui Wei, Junda Wu, Siru Ouyang, Yizhu Jiao, Shijia Pan, Julian McAuley, Yu~Zhang, Tong Yu, and Jiawei Han.
\newblock Filesystem-based memory for {LLM} agents: Organization, evolution, and sustainability.
\newblock \emph{arXiv preprint arXiv:2607.26637}, 2026.

\bibitem[Zhu et~al.(2025)Zhu, Hao, Hu, Jiao, Russell, and Tian]{zhu2025superposition}
Hanlin Zhu, Shibo Hao, Zhiting Hu, Jiantao Jiao, Stuart Russell, and Yuandong Tian.
\newblock Reasoning by superposition: A theoretical perspective on chain of continuous thought.
\newblock In \emph{Advances in Neural Information Processing Systems (NeurIPS)}, 2025.

\end{thebibliography}
